\documentclass[twoside,11pt]{article}

\usepackage[abbrvbib]{jmlr2e}

\usepackage[utf8]{inputenc}
\usepackage[T1]{fontenc}
\usepackage{lmodern}
\usepackage{url}
\usepackage{xurl}
\usepackage{booktabs}
\usepackage{amsmath}
\usepackage{amsfonts}
\usepackage{nicefrac}
\usepackage{microtype}
\usepackage{xcolor}
\usepackage{multirow}
\usepackage{longtable}
\usepackage{lastpage}
\usepackage{float}

\renewcommand{\floatpagefraction}{0.75}

\providecommand{\qed}{\hfill\BlackBox}
\newcommand{\hd}[2][c]{\begin{tabular}[b]{@{}#1@{}}#2\end{tabular}}

\makeatletter
\newcommand{\blfootnote}[1]{%
  \begingroup\renewcommand{\thefootnote}{}%
  \long\def\@makefntext##1{\noindent##1}%
  \footnotetext{#1}\endgroup}
\makeatother

\ShortHeadings{Certifying Declared Change Envelopes}{Usama and Qadir}
\firstpageno{1}

\makeatletter\def\@starteditor{}\def\@editor{}\def\@endeditor{}\makeatother

\begin{document}

\title{Where Did the Update Go? Certifying Declared Change Envelopes for Model Updates}

\author{\name Muhammad Usama \email usama@braindeck.net \\
       \addr Braindeck Inc.\\
       Seoul, South Korea
       \AND
       \name Muhammad Saad Qadir \email Saadqadir.h11@bahria.edu.pk \\
       \addr Department of Computer Science\\
       Bahria School of Engineering and Applied Sciences\\
       Bahria University, Islamabad, Pakistan}

\maketitle

\begin{abstract}%
Deployed language models are updated continually, and regulators increasingly accept changes that were declared in advance. Showing that an update changed behavior only where it was declared to is a different statistical problem from bounding how much it changed: a useful update is meant to change some decisions, so global disagreement and regression bounds cannot certify it. We formalize the problem with \emph{declared change envelopes}. Before training, the developer commits to a region of inputs, and every decision that changes outside it counts as leakage. Leakage compares two models, so it can be certified from unlabeled traffic with anytime-valid tests, whose sample cost we bound from above and below, and per-update certificates compose into a bound on lifetime drift outside all declared scopes. We measure leakage for LoRA and full fine-tuning, preference optimization, reinforcement learning, and knowledge editing on models from 1.5B to 72B parameters, mainly on multiple-choice traffic from eleven domains. Narrow updates leak widely: domain and policy LoRA updates keep only about 7--36\% of the decisions they change inside the declared scope. Across 256 trained variants, leakage tracks the update's KL divergence to the incumbent on out-of-scope inputs (Spearman 0.94). None of the customary remedies we tested brings the leakage of a rule-like update to a certifiable level without giving up a large part of its intended change. Distilling the incumbent's decisions on unlabeled prompts in the deployment format does: anchored LoRA, GRPO, and (at a small learning rate) full fine-tuning updates certify 5\% leakage from a few hundred to a few thousand unlabeled samples, and the certificates replicate on held-out audit traffic and on written answers. Anchoring keeps large intended changes but on one of two models costs part of a small medical gain. The certificates cover decision-like outputs only: on open-ended chat, re-batching the unchanged model already changes the substance of 11--21\% of responses, and no update certifies. Code: \url{https://github.com/Usama1002/declared-change-envelopes}.
\end{abstract}


\section{Introduction}
\label{sec:intro}

Deployed language models are adapted to new domains, patched with corrected facts, given new policies, and increasingly trained on their own deployment traffic \citep{cursor2025tab,zweiger2025seal,shenfeld2026sdft}. Every update comes with a stated intent, and regulators have started to build on these statements. The EU AI Act exempts learning-driven changes that were predetermined in the initial conformity assessment from re-assessment \citep{euaiact2024}, and the U.S.\ FDA lets device makers pre-specify future modifications in a predetermined change control plan \citep{fda2024pccp}. Both presuppose evidence that an update changed behavior \emph{only where it was declared to}. We are not aware of a method that produces such evidence for language model updates, and it has not been measured whether updates trained by standard recipes would pass such a test.

\paragraph{The problem.} We study the following question. An operator serves an incumbent model $f_0$ on a traffic distribution $P$ and wants to replace it by a candidate $f_1$ that was trained to change behavior on a stated region of inputs. Before $f_1$ is trained, the developer commits to that region $E$. After training, the operator wants a guarantee of the form ``on traffic outside $E$, $f_1$ makes a different decision from $f_0$ on at most a fraction $\varepsilon$ of inputs'', with error probability at most $\alpha$. The guarantee should need no labels, because deployment traffic is unlabeled; it should stay valid while the operator monitors the audit continuously and stops whenever the evidence suffices; and it should extend over a sequence of updates, each with its own declared region. We call the out-of-scope decision-change rate the \emph{leakage} of the update and a guarantee of this form a \emph{scoped certificate}.

\paragraph{Why existing checks do not answer it.} Existing update checks bound a different quantity. Regression testing, negative-flip rates, churn, and recent statistical certificates bound \emph{how much} a model changed or got worse, over all traffic \citep{yan2021positive,jiang2022churn,echterhoff2024muscle,balachandran2026discern}. A global disagreement bound cannot certify an update that is meant to change behavior: the in-scope change alone keeps the global disagreement rate above any small tolerance (\S\ref{sec:scoped}). Tests based on outcome labels are indifferent to where accuracy-neutral changes occur: a medical adaptation that silently rewrites a tenth of the model's legal answers passes them if the rewrites are accuracy-neutral on average, and this is what we observe for the medical and legal updates of our two main models and for ten public fine-tunes (\S\ref{sec:res-leak}). Model equality tests ask whether two models differ at all \citep{gao2025met}, which every useful update does. Knowledge-editing benchmarks measure locality on curated neighbor prompts, not on the traffic a model serves, and come with no statistical guarantee \citep{meng2022locating,cohen2024evaluating}. A scoped certificate needs a different estimand: decision change restricted to the complement of a region that was fixed before the update existed.

\paragraph{Approach.} We formalize this estimand with \emph{declared change envelopes} (\S\ref{sec:framework}). The developer commits, by publishing a hash, to a region of input space specified by a rule, a natural-language scope evaluated by a fixed judge, or a learned score. After training, the candidate is shadowed on traffic, and every decision that changes outside the envelope counts as leakage. Leakage compares two models, so it needs no labels; it can be certified with an anytime-valid confidence sequence or, at a tolerance fixed in advance, a one-sided sequential test, and per-update certificates compose into a lifetime bound on drift outside the union of all declared scopes. Beneficial out-of-scope changes also count as leakage: the certificate states where an update acted, not whether its changes are good, and it complements label-based tests of the in-scope change.

\paragraph{The empirical question.} A certificate is only useful if real updates can earn it, so most of the paper measures leakage. We train scoped updates with LoRA and full fine-tuning, decision-token DPO, reinforcement learning (GRPO on the decision, and GRPO with verifiable rewards on free-form math), and two knowledge editors, on instruction-tuned models from three families between 1.5B and 72B parameters. The main traffic is 34{,}136 multiple-choice and classification queries from eleven domains, where a model's decision is the option letter it prefers; free-form questions, unsafe prompts, open-ended chat, and ten public fine-tunes extend the audit. Every claim is checked against a self-disagreement floor that measures numerical nondeterminism, and the main certificates are re-tested on a confirmatory audit split that no model was evaluated on during development.

\paragraph{Findings.} Five findings emerge, each stated at the width of its evidence. First, declared scopes do not contain update footprints (\S\ref{sec:res-leak}, \S\ref{sec:res-structure}). Domain and policy LoRA updates keep only about 7--36\% of the decisions they change inside their declared scope, and every update type we trained, except RL with verifiable rewards on strong incumbents, changes out-of-scope decisions at several times the numerical floor or more. For the medical and legal updates of our two main models the damage is invisible to aggregate metrics: net out-of-scope accuracy moves by at most 1.3 points while 7--13\% of out-of-scope decisions switch between right and wrong. Second, leakage tracks a single quantity, the update's KL divergence to the incumbent on out-of-scope inputs (\S\ref{sec:res-kl}): across 256 trained variants of all update types and sizes the Spearman correlation is 0.94, and a pointwise Pinsker condition (Lemma~\ref{lem:pinsker}) bounds leakage within a factor of 4.6 for every variant. Third, the customary remedies do not target this quantity (\S\ref{sec:res-remedies}). Learning-rate choice, weight interpolation, an L2 penalty, replay, KL on generic text, focal distillation on the training prompts, and a stronger reference-model KL in RL either leave the leakage of a rule-like update well above a certifiable level or remove much of its intended change.

Fourth, anchoring the update to the incumbent's decisions on unlabeled prompts in the deployment format makes updates certifiable on multiple-choice traffic (\S\ref{sec:res-anchor}--\ref{sec:res-edit}). It cuts leakage three- to sevenfold for LoRA domain and option-shuffled policy updates, and to 1.2--4.3\% for GRPO and 1.6--2.9\% for full fine-tuning at a small learning rate. Anchored candidates certify 5\% leakage from a few hundred to a few thousand unlabeled samples; the certificates replicate on held-out audit traffic and on the models' written answers, and they compose over a sequence of four updates. Large intended changes are kept, while part of a small medical gain is lost on one of two models. Fifth, the certificates apply to decision-like outputs (\S\ref{sec:res-beyond}). On open-ended chat, re-batching the unchanged model already changes the substance of 11--21\% of responses; anchored updates stay near the floor of a tiny random weight perturbation, but no variant certifies.

\paragraph{Contributions.} The first contribution is the problem formulation: an estimand, leakage outside a region committed before training, that can certify updates meant to change behavior, together with a self-disagreement floor that makes numerical nondeterminism explicit (\S\ref{sec:framework}). The second is a certification analysis for this estimand. The statistical tools are standard, and our contribution is to make their consequences for model audits explicit: a confidence sequence and a one-sided test with sample costs within a small factor of a lower bound (within 16\% for the one-sided test on a contained update), a lifetime scope-drift bound with the conditions under which an audit set can be reused when promotion decisions depend on it, and a pointwise flip condition that ties leakage to out-of-scope KL for any update, however it was trained (\S\ref{sec:cert}--\ref{sec:anchor}). The third is a measurement study across LoRA, full fine-tuning, DPO, GRPO, RL with verifiable rewards, and knowledge editing from 1.5B to 72B parameters that establishes the first two findings, with a self-disagreement floor, a held-out confirmatory split, and controls for each claim (\S\ref{sec:res-leak}--\ref{sec:res-kl}). The fourth is evidence that \emph{decision anchoring}, a known label-free distillation term, is what makes scoped certificates reachable across update types, together with what it requires: anchor prompts in the deployment's decision format and an optimizer that can reach a small anchor KL (\S\ref{sec:res-remedies}--\ref{sec:res-edit}). The fifth is a map of where the approach stops working, on free-form and open-ended outputs, with the floors that explain why (\S\ref{sec:res-beyond}). Code to reproduce every experiment, table, and figure, together with the result files, is available at \url{https://github.com/Usama1002/declared-change-envelopes}.

\section{Related work and positioning}
\label{sec:related}

Three lines of work are closest to ours: regression reduction between model versions, locality of edits and side effects of fine-tuning, and statistical auditing of deployed models. Table~\ref{tab:positioning} summarizes what each measures and why none yields a scoped certificate.

\begin{table}[t]
\centering
\caption{What the closest lines of work measure, and why none of them certifies where an update acted. ``Global'' means over all traffic, including the inputs the update is meant to change.}
\label{tab:positioning}
\small
\setlength{\tabcolsep}{4pt}
\begin{tabular}{p{0.24\linewidth}p{0.30\linewidth}p{0.38\linewidth}}
\toprule
Line of work & Quantity & Gap for a scoped certificate \\
\midrule
Regression reduction and positive-congruent training & Negative flips or regressions, global & Needs labels; blind to accuracy-neutral rewrites; a training goal, not a test \\
Churn reduction & Prediction churn, global & A training goal; churn of a useful update cannot be small \\
Model equality testing & Whether two models differ at all & Rejects every useful update \\
DISCERN (concurrent) & Global disagreement, with labels only on disagreements & Certifies only updates whose global disagreement is small \\
Slice-wise certification (concurrent) & Non-regression per data slice & Label-based; tests improvement, not where change occurred \\
Editing locality & Change on curated neighbor prompts & Not a traffic distribution; no guarantee \\
Anytime-valid model auditing & A property of one deployed model & Different estimand; not a comparison of two models \\
\midrule
This work & Decision change outside a region committed before training & Label-free and anytime-valid; certifies updates meant to change behavior \\
\bottomrule
\end{tabular}
\end{table}

\paragraph{Compatible and regression-free model updates.}
Updating a deployed model can break behavior users relied on even when aggregate accuracy improves \citep{bansal2019updates,srivastava2020empirical}. Positive-congruent training and its successors reduce negative flips, the inputs a new model gets wrong that the old one got right \citep{yan2021positive,echterhoff2024muscle,xie2021regression,cai2022measuring}; weight interpolation toward the old model, which our adapter-scaling control implements, reduces regressions \citep{schumann2023backward}; and prediction churn is reduced by distilling toward the previous model \citep{milanifard2016launch,jiang2022churn}, which our anchoring objective applies at the decision position. Concurrent work certifies update verdicts statistically. DISCERN labels only the inputs on which two models disagree, certifies benign updates from unlabeled traffic when their \emph{global} disagreement rate is below a tolerance, uses anytime-valid confidence sequences, and composes across promotions \citep{balachandran2026discern}. Concurrent slice-wise tests with incumbent fallback certify upgrades per data slice \citep{zhang2026slicewise}. A useful update is \emph{supposed} to change decisions, so its global disagreement rate cannot be small and a global zero-label test cannot certify it (\S\ref{sec:scoped}). Restricting the label-free test to the complement of a region committed before training makes it applicable, and yields a lifetime statement about \emph{where} a sequence of updates acted.

\paragraph{Locality of edits and side effects of narrow fine-tuning.}
Knowledge editors are evaluated on locality \citep{meng2022locating,meng2023memit}, which degrades under sequential editing \citep{gupta2024model}, is preserved for chosen keys by null-space projection \citep{fang2025alphaedit}, and is undermined by hard-to-anticipate ripple effects \citep{cohen2024evaluating}. Locality scores are averages over curated neighbor prompts and carry no guarantee for the traffic a model serves; our edit experiments show that an update can look local on average while rewriting a stratum completely (\S\ref{sec:res-edit}). Narrow fine-tuning can erode safety alignment \citep{qi2024finetuning} or induce broad misalignment \citep{betley2025emergent}, continual fine-tuning forgets prior skills \citep{kirkpatrick2017overcoming,luo2023empirical,biderman2024lora}, on-policy updates forget less \citep{shenfeld2025rl}, model-diffing methods describe differences after the fact \citep{zhong2022describing,dunlap2025vibecheck,kassem2025mneme}, and API models drift across versions \citep{chen2023chatgpt}. We instead make locality a property that is declared before training and certified afterwards on traffic.

\paragraph{Anytime-valid inference and model auditing.}
Confidence sequences remain valid under continuous monitoring and optional stopping \citep{ville1939etude,robbins1970statistical,howard2021time,waudbysmith2024estimating}, whereas repeatedly checking fixed-sample tests inflates error \citep{johari2017peeking}. Sequential tests monitor the risk of a deployed model \citep{podkopaev2022tracking}, anytime-valid betting tests audit a deployed model for fairness \citep{chugg2023auditing}, and model equality testing asks whether an API serves a given model \citep{gao2025met}. Prediction-powered inference \citep{angelopoulos2023ppi}, learn-then-test \citep{angelopoulos2021learn}, Seldonian and high-confidence policy improvement \citep{thomas2015high,thomas2019preventing}, and SPIBB \citep{laroche2019safe} gate changes on statistical tests of performance. We apply standard anytime-valid tools to a different estimand: the rate of decision change between two models in the complement of a region committed before training. The new elements are the estimand and its commitment protocol, the self-disagreement floor that limits what can be certified, the correction needed when promotion decisions reuse the audit set, and the measurement of whether trained updates can meet such a certificate.

\section{Declared change envelopes}
\label{sec:framework}

\paragraph{Setting.}
Let $\mathcal{X}$ be the space of inputs a deployed model receives and $P$ the traffic distribution over $\mathcal{X}$. A model $f$ induces a decision map $d_f:\mathcal{X}\to\mathcal{Y}$; in our experiments $d_f(x)$ is the option selected by $f$ on a multiple-choice or classification query, but the framework only needs $d_f$ to be computable without labels (for free-form generation it can be a semantic-equivalence class). An \emph{update} replaces the incumbent $f_0$ by a candidate $f_1$, with change indicator and \emph{footprint}
\begin{equation}
\Delta(x) = \mathbb{1}\left[d_{f_0}(x)\neq d_{f_1}(x)\right], \qquad \mathcal{F} = \{x : \Delta(x)=1\}, \qquad \phi = P(\mathcal{F}).
\end{equation}

\paragraph{Envelopes.}
A \emph{declared change envelope} is a measurable set $E\subseteq\mathcal{X}$, specified before the update is trained by a membership function $c_E:\mathcal{X}\to\{0,1\}$ (a taxonomy rule, a natural-language scope evaluated by a fixed judge, or a learned score with a fixed threshold). The specification is committed by publishing its hash before training, so it cannot be redrawn around the realized footprint. We define the envelope mass, the \emph{leakage}, the \emph{leaked mass}, and the \emph{containment}:
\begin{equation}
\begin{aligned}
m(E)&=P(E), & \lambda(E)&=P\left(\Delta=1 \mid x\notin E\right),\\
\ell(E)&=P\left(\Delta=1,\ x\notin E\right)=(1-m)\lambda, & \kappa(E)&=P\left(x\in E\mid \Delta=1\right)=1-\ell/\phi.
\end{aligned}
\end{equation}
A declaration is informative only if $m(E)$ is small; $E=\mathcal{X}$ has zero leakage and says nothing. We therefore report leakage together with mass and compare declarations by their \emph{footprint curves} $m\mapsto\lambda(E_m)$, where $E_m$ is the top-$m$ fraction of traffic ranked by a declaration score.

\begin{definition}[Footprint certificate]
An $(E,\varepsilon,\alpha)$ footprint certificate is issued by an audit procedure that, whenever $\lambda(E)>\varepsilon$, certifies with probability at most $\alpha$ over the audit sample, under any monitoring policy. Because $\ell\le\lambda$, a certificate on $\lambda$ also bounds the leaked mass $\ell$.
\end{definition}

\paragraph{Self-disagreement and decisive changes.}
Deployed inference is not bitwise reproducible: batching and kernel selection perturb low-precision logits, so when $f_0$ and $f_1$ see different batches part of the observed $\Delta$ is numerical. We measure this \emph{self-disagreement floor} $\lambda_0=P(d_{f_0}(x)\neq d_{f_0^{\rho}}(x))$, where $f_0^{\rho}$ is the incumbent under a benign re-batching $\rho$. No certificate on raw decisions can be issued below it. The next lemma says where both self-disagreement and small updates act.

\begin{lemma}[Margin lemma]
\label{lem:margin}
For an input with $k$ options, let $z_f(x)\in\mathbb{R}^k$ be the option log-probabilities of $f$ and $\mu_f(x)$ the gap between the largest and second-largest entries. If $\|z_{f_1}(x)-z_{f_0}(x)-s\mathbf{1}\|_\infty\le\delta(x)$ for some scalar $s$, then $\Delta(x)=1$ implies $\mu_{f_0}(x)\le 2\delta(x)$.
\end{lemma}
An update that perturbs option scores by $\delta$ can only change decisions whose margin is below $2\delta$, which is why numerical noise and mild updates act on fragile decisions. We therefore also report \emph{decisive} leakage $\lambda_\tau(E)=P(\Delta=1\mid x\notin E,\ \mu_{f_0}(x)>\tau)$, where $\tau$ is the smallest margin at which the incumbent's self-disagreement falls below $10^{-3}$, estimated on held-out traffic before any update is trained. All proofs are in Appendix~\ref{app:proofs}.

\subsection{Why scoped certification differs from global bounds}
\label{sec:scoped}
Write $\lambda_{\mathrm{in}}(E)=P(\Delta=1\mid x\in E)$ for the in-scope change rate. The global disagreement rate decomposes as
\begin{equation}
\label{eq:decomp}
\phi\ =\ m(E)\,\lambda_{\mathrm{in}}(E)\ +\ \big(1-m(E)\big)\,\lambda(E).
\end{equation}
Three consequences separate a scoped certificate from the global and label-based bounds of prior work.

\emph{A global bound cannot certify a useful update.} A valid test of $\phi\le\varepsilon_g$ passes with probability above $\alpha$ only if $m\lambda_{\mathrm{in}}\le\varepsilon_g$, that is, only if the intended change itself is small. For the Qwen policy update of \S\ref{sec:res-leak}, $m=8.9\%$ and $\lambda_{\mathrm{in}}=23.7\%$, so $\phi\ge2.1\%$ even if the update leaked nothing; for the Qwen medical update the in-scope term alone is $4.1\%$. A global tolerance loose enough to admit such updates says little about out-of-scope traffic, since $\phi\le\varepsilon_g$ implies only $\lambda\le\varepsilon_g/(1-m)$. A scoped certificate sets its tolerance on $\lambda$ independently of the size of the intended change.

\emph{Label-based bounds miss accuracy-neutral change.} With outcome labels, the out-of-scope accuracy change is $P(\text{wrong}\to\text{right})-P(\text{right}\to\text{wrong})$, and negative-flip rates bound the second term. Leakage also counts changes between two wrong answers and flips that cancel in the net, so $\lambda$ can be large while accuracy, and even the negative-flip rate, stay small. The plain Qwen medical update raises out-of-scope accuracy by 1.1 points while 7.7\% of out-of-scope decisions switch between right and wrong and 9.9\% change in total (Table~\ref{tab:footprint}).

\emph{The region must be fixed before the footprint exists.} If $E$ could be chosen after training, $E=\mathcal{F}$ would give zero leakage. Committing to $c_E$ by hash before training, and reporting $m(E)$ with every certificate, is what makes a small $\lambda(E)$ informative. The same requirement rules out choosing $E$ from audit data; Appendix~\ref{app:proofs} shows how to select among a grid of committed envelope sizes after auditing at a corrected level.

\subsection{Label-free, anytime-valid certification}
\label{sec:cert}
Certifying $\lambda(E)$ needs no labels: the operator shadows the candidate on traffic and observes $Z_t=\Delta(x_t)$ on the inputs with $c_E(x_t)=0$. We assume the audit inputs are i.i.d.\ draws from $P$ that were not used to train or select the update, so that $Z_t\sim\mathrm{Bernoulli}(\lambda)$ i.i.d. Because operators monitor continuously, and repeatedly checking a fixed-sample bound inflates its error \citep{johari2017peeking}, we use a Beta-mixture martingale \citep{robbins1970statistical,howard2021time}. With $S_t=\sum_{s\le t}Z_s$ and a $\mathrm{Beta}(1,1)$ mixing prior,
\begin{equation}
M_t(p)=\frac{B(1+S_t,\,1+t-S_t)}{p^{S_t}(1-p)^{t-S_t}},\qquad U_t=\sup\{p\in[0,1]: M_t(p)<1/\alpha\}.
\end{equation}

\begin{theorem}[Anytime-valid footprint certificate]
\label{thm:cs}
If $Z_1,Z_2,\dots$ are i.i.d.\ $\mathrm{Bernoulli}(\lambda)$, then $P(\exists t\ge1: \lambda>U_t)\le\alpha$. Hence stopping at any data-dependent time $\sigma$ with $U_\sigma\le\varepsilon$ yields a valid $(E,\varepsilon,\alpha)$ certificate, and an update with $\lambda>\varepsilon$ is certified with probability at most $\alpha$ under any monitoring policy. Moreover, with $\hat\lambda_t=S_t/t$ and $\mathrm{kl}(a\|b)$ the Bernoulli relative entropy,
\begin{equation}
U_t\ \le\ \sup\left\{p\ge\hat\lambda_t:\ t\,\mathrm{kl}(\hat\lambda_t\,\|\,p)\le \log\tfrac{t+1}{\alpha}\right\},
\end{equation}
so, up to the check interval, a certificate at level $\varepsilon$ is issued no later than the first $t$ with $\hat\lambda_t<\varepsilon$ and $t\,\mathrm{kl}(\hat\lambda_t\|\varepsilon)>\log((t+1)/\alpha)$. If no change is observed ($S_t=0$), then $U_t\le \log((t+1)/\alpha)/t$.
\end{theorem}
The bound makes the cost of certification explicit. A perfectly contained update is certified after 451 unlabeled out-of-scope samples for $\varepsilon=0.02$ and $\alpha=0.05$ (457 by the closed-form bound $\log((t+1)/\alpha)/t\le\varepsilon$). An update with $\lambda=\varepsilon/2$ is certified after about 3{,}600 samples by the bound when the empirical rate equals $\lambda$ (a median of 2{,}340 in simulation, 1{,}070 for the one-sided test of Proposition~\ref{prop:onesided}, Appendix~\ref{app:sim}). The $\log(t+1)$ term is the cost of adapting to an unknown leakage rate with a uniform prior, not of anytime validity itself (Proposition~\ref{prop:onesided}). No valid procedure can do much better than either:

\begin{proposition}[Lower bound]
\label{prop:lower}
Let $\sigma$ be the stopping time of any sequential procedure that certifies $\lambda\le\varepsilon$ with probability at most $\alpha$ whenever $\lambda>\varepsilon$, and with probability at least $1-\beta$ when $\lambda=\lambda_0<\varepsilon$. Then $\mathbb{E}_{\lambda_0}[\sigma]\ge \mathrm{kl}(1-\beta\,\|\,\alpha)/\mathrm{kl}(\lambda_0\|\varepsilon)$, which is about $\log(1/\alpha)/\mathrm{kl}(\lambda_0\|\varepsilon)$ for small $\alpha,\beta$.
\end{proposition}
The mixture certificate is within a factor of about three of this bound at our operating points (451 versus 149 samples for a contained update at $\varepsilon=2\%$, $\alpha=5\%$). When the tolerance $\varepsilon$ is fixed in advance, a one-sided test closes most of the gap:

\begin{proposition}[One-sided certificate at a declared tolerance]
\label{prop:onesided}
For $q\in[0,\varepsilon)$ let
\[
R_t(q)=\prod_{s\le t}\left(\frac{q}{\varepsilon}\right)^{Z_s}\left(\frac{1-q}{1-\varepsilon}\right)^{1-Z_s},
\]
and let $W$ be a prior on $[0,\varepsilon)$ fixed before auditing. Then $R_t^W=\int R_t(q)\,dW(q)$ is a nonnegative supermartingale with $R_0^W=1$ whenever $\lambda\ge\varepsilon$, so certifying at the first $t$ with $R_t^W\ge1/\alpha$ certifies an update with $\lambda\ge\varepsilon$ with probability at most $\alpha$ under any monitoring policy. If no change is observed, it certifies by $t=\lceil\log(1/(\alpha W(\{0\})))/(-\log(1-\varepsilon))\rceil$.
\end{proposition}
With $W$ placing mass $1/2$ at $0$ and $1/18$ at each of $\varepsilon/10,\dots,9\varepsilon/10$, a contained update certifies at $\varepsilon=2\%$ after 172 samples, within 16\% of the lower bound for a contained update. Because our commitment hashes did not record $\varepsilon$, we apply the test at level $\alpha/16$, a Bonferroni correction over four candidates and four tolerances, in \S\ref{sec:res-anchor}. The test certifies only the declared $\varepsilon$; the confidence sequence certifies every $\varepsilon$ at once. Certificates transfer under traffic shift: with density ratio $dP/dQ\le w$ outside $E$, $\ell_P\le w\,\ell_Q$ (Appendix~\ref{app:proofs}, which also shows how to choose the envelope size after auditing).

When several candidate updates are audited on the same traffic and the best is promoted, the level must be split among them; our candidates for each update are plain LoRA and anchoring with $\beta\in\{0.1,1,10\}$, so we certify at $\alpha/4$ (other variants are diagnostics that are never promoted). In simulation (Appendix~\ref{app:sim}), recomputing a one-sided Clopper--Pearson bound every 5 samples falsely certifies 33\% of updates whose leakage equals $\varepsilon=2\%$, more than six times the nominal 5\%, while the mixture sequence falsely certifies 0.05\% and the one-sided test 3.3\%.

\subsection{Lifetime composition}
\label{sec:compose}
A continually improving system applies updates $f_0\to f_1\to\cdots\to f_T$ with committed envelopes $E_1,\dots,E_T$, and small out-of-scope changes could accumulate.

\begin{theorem}[Scope-drift bound]
\label{thm:compose}
Let $\Delta_t(x)=\mathbb{1}[d_{f_{t-1}}(x)\neq d_{f_t}(x)]$, $\ell_t=P(\Delta_t=1,\ x\notin E_t)$, and $\mathcal{U}_T=\bigcup_{t\le T}E_t$. Then for every $T$,
\begin{equation}
P\left(d_{f_0}(x)\neq d_{f_T}(x),\ x\notin\mathcal{U}_T\right)\ \le\ \sum_{t=1}^{T}\ell_t .
\end{equation}
If each $\ell_t$ is certified by the confidence sequence of Theorem~\ref{thm:cs} at level $\alpha_t=6\alpha/(\pi^2t^2)$, with upper bound $U^{(t)}$, on audit traffic that is independent of $(f_{t-1},f_t)$, then with probability at least $1-\alpha$, simultaneously for all $T$, the drift of $f_T$ from $f_0$ outside $\mathcal{U}_T$ is at most $\sum_{t\le T}U^{(t)}$.
\end{theorem}
The bound grows linearly in $T$ (vacuous after roughly $1/\varepsilon$ updates of leakage $\varepsilon$) and is tight when per-step leak sets are disjoint, lie outside $\mathcal{U}_T$, and are never reverted. The independence condition holds with fresh audit traffic at every step. A shared audit set suffices only if no training or promotion decision depended on it: if promotion at a step is gated on a certificate computed on that set, the next model depends on the audit data, and the guarantee then requires fresh traffic or certifying every candidate at step $t$ at level $\alpha_t/K^{t}$ (with $K$ candidates per step, or $(K+1)^t$ if a step may keep the incumbent), which covers all promotion histories.

\subsection{Anchored updates}
\label{sec:anchor}
Theorem~\ref{thm:cs} rewards updates with small leakage, which suggests training for it directly. Given an unlabeled sample $\mathcal{A}$ of prompts in the deployment format, in our main experiments the traffic that the declaration places outside the envelope ($c_E(x)=0$), we add a consistency term to the update objective:
\begin{equation}
\label{eq:anchor}
\mathcal{L}(\theta)=\mathcal{L}_{\text{task}}(\theta)\ +\ \beta\ \mathbb{E}_{x\sim\mathcal{A}}\ \mathrm{KL}\!\left(p_{f_0}(\cdot\mid x)\,\|\,p_{f_\theta}(\cdot\mid x)\right),
\end{equation}
where the KL is taken over the next-token distribution at the decision position. The term is not new as a regularizer. It is churn-reduction distillation \citep{jiang2022churn,li2017lwf}, the locality KL of model editors \citep{mitchell2022mend} (a relative of the weight-change constraints of \citealp{gu2024editing}), the retain-set KL of unlearning \citep{maini2024tofu}, function-space regularization in continual learning \citep{titsias2020functional}, and a relative of self-distillation fine-tuning \citep{yang2024sdft}, here applied at the decision position. We use it to make a scoped certificate reachable and test whether it does so across update types. Our experiments show that it acts as \emph{decision-format distillation}: excluding the envelope from the anchor set does not matter, and anchors from outside the traffic pool work nearly as well as long as they share its decision format (\S\ref{sec:res-anchor}). The envelope defines what is certified, not what is anchored. Anchoring needs no labels and uses only traffic that exists before the update is trained; certificates are computed on disjoint audit traffic. The following bound links the anchoring term to the certified quantity.

\begin{remark}[Anchoring and leakage]
\label{prop:anchor}
Let $g_0(x)$ be the gap between the incumbent's next-token probabilities of its chosen option and of its runner-up option, and $K(x)=\mathrm{KL}(p_{f_0}(\cdot\mid x)\|p_{f_1}(\cdot\mid x))$. For every $\gamma>0$,
\begin{equation}
\lambda(E)\ \le\ P\left(g_0(x)\le\gamma \mid x\notin E\right)\ +\ \frac{2\,\mathbb{E}\left[K(x)\mid x\notin E\right]}{\gamma^2}.
\end{equation}
\end{remark}
Remark~\ref{prop:anchor} is the averaged form of a pointwise condition, which is much tighter and holds for any update, however it was trained.

\begin{lemma}[Pointwise flip condition]
\label{lem:pinsker}
Let $p_0(\cdot\mid x)$ and $p_1(\cdot\mid x)$ be distributions over a set containing the options (the next-token distributions, or their renormalizations to the valid option letters), with decisions $d_{f_i}(x)=\arg\max$ over the options of $p_i$, and let $g_0(x)$ be the gap between the two largest option probabilities under $p_0$. If $d_{f_1}(x)\ne d_{f_0}(x)$, then $\mathrm{KL}(p_0\|p_1)\ge g_0(x)^2/2$. Hence
\begin{equation}
\lambda(E)\ \le\ P\!\left(\mathrm{KL}\big(p_0(\cdot\mid x)\,\|\,p_1(\cdot\mid x)\big)\ \ge\ g_0(x)^2/2\ \Big|\ x\notin E\right).
\end{equation}
\end{lemma}
Remark~\ref{prop:anchor} follows by splitting on $g_0\le\gamma$ and applying Markov's inequality, which is where it loses most of its tightness. \S\ref{sec:res-kl} measures how much of the observed leakage Lemma~\ref{lem:pinsker} explains across update types.

\section{Experimental setup}
\label{sec:setup}

\paragraph{Traffic.}
Traffic is a pool of 34{,}136 multiple-choice and classification queries from eleven domains: MMLU \citep{hendrycks2021mmlu} split into medical, legal, STEM, and humanities/social subjects; ARC \citep{clark2018arc}; CommonsenseQA \citep{talmor2019commonsenseqa}; MedMCQA \citep{pal2022medmcqa}; AG News topic classification \citep{zhang2015agnews}; BoolQ \citep{clark2019boolq}; SST-2 \citep{socher2013sst}; CaseHOLD \citep{zheng2021casehold,chalkidis2022lexglue}; and CounterFact paraphrase and neighborhood prompts \citep{meng2022locating}. The decision $d_f(x)$ is the valid option letter with the highest next-token log-probability after the chat template and a fixed assistant prefix; it is deterministic and label-free. We split the pool once at random into an \emph{anchor} set (30\%), which stands for traffic observed before an update is trained and is the only data used for declarations, thresholds, and anchoring, and a disjoint \emph{audit} set (70\%) on which every footprint quantity, effect, and certificate is computed. Base models are evaluated on the full pool; to fit the experimental grid on one GPU, every update is evaluated on a fixed random half of it (17{,}068 queries), so update statistics use its 11{,}945 audit queries (the development audit set) and 5{,}123 anchor queries, while anchoring draws on the whole anchor set. The other 11{,}951 audit queries form a \emph{confirmatory} set that no model was evaluated on during development; we use it once, at the end, to test whether the certificates replicate (\S\ref{sec:res-anchor}). In the text, Qwen and Llama denote the 7B and 8B incumbents unless a size is given.

\paragraph{Models.}
The main incumbents are Qwen2.5-7B-Instruct \citep{qwen2025qwen25} and Llama-3.1-8B-Instruct \citep{grattafiori2024llama3}. The medical and policy updates are repeated on Mistral-7B-Instruct-v0.3 \citep{jiang2023mistral}, Qwen2.5-32B-Instruct, and Qwen2.5-72B-Instruct, and the medical, legal, policy, and RLVR updates on Qwen2.5-1.5B-Instruct, a weaker incumbent with room for large gains.

\paragraph{Scoped updates.}
Four updates mirror common enterprise requests. Each is trained with LoRA \citep{hu2022lora} (rank 16, all attention and MLP projections, learning rate $10^{-4}$, batch 16, cross-entropy on the answer token):
\emph{med}, domain adaptation on 2{,}000 MedMCQA training questions (declared scope: medical questions);
\emph{legal}, adaptation on 2{,}000 CaseHOLD training items (scope: legal questions);
\emph{policy}, an organizational relabeling rule that files Science/Technology news under Business, trained on 2{,}000 AG News articles (scope: news-topic classification);
\emph{edit}, 400 counterfactual knowledge edits from CounterFact (scope: statements about the edited entities), trained alone or mixed with 400 unedited true facts (\S\ref{sec:res-edit}).
Training prompts never appear in the pool, which contains only the paraphrase and neighborhood prompts of the edited records. The \emph{taxonomy} envelope of each update is the corresponding task family in the pool (for \emph{edit}, the paraphrases of edited facts; neighborhood prompts about other entities are out of scope, as in standard locality evaluation). Updates use two seeds on Qwen-7B, Llama, 32B, and 72B, one on Mistral and 1.5B, and three for key anchored variants.

\paragraph{Other update types.}
The medical and policy updates are also trained (i) with decision-token DPO \citep{rafailov2023dpo}, preferring the target option over the incumbent's strongest wrong option; (ii) by \emph{full fine-tuning} of every weight (fp32 master weights, 8-bit AdamW, learning rates $10^{-6}$, $3\cdot10^{-6}$, and $10^{-5}$); and (iii) by \emph{decision-token GRPO} \citep{shao2024deepseekmath}, a reinforcement-learning update that samples eight answer letters per prompt from the policy, rewards the target letter, uses group-normalized advantages, and penalizes KL to the incumbent on the training prompts with the customary weight $0.04$ or a strong weight $1$. As a realistic reinforcement-learning update we also run \emph{RLVR}: GRPO with free generation on GSM8K training problems \citep{cobbe2021gsm8k}, rewarded when the stated final number is correct (declared scope: the MMLU mathematics and statistics subjects, 3.1\% of the pool). Anchoring (Eq.~\ref{eq:anchor}) is added to each update type unchanged; for RLVR on the 1.5B model it is also applied at the token level on out-of-scope free-form prompts (NQ-open training questions with the incumbent's answers). Details are in Appendix~\ref{app:more}.

\paragraph{Declarations.}
Scores that rank traffic by how likely it is to change, all computed before training: an \emph{NL judge} (the probability that Qwen2.5-7B-Instruct says an input falls within the update's natural-language scope); \emph{data kNN} (embedding similarity to the five nearest training examples); \emph{fragility} (the negative incumbent margin); and a \emph{pilot forecast}, a logistic regression fit on the anchor set to the flips of a cheap pilot update (another seed, 25\% of the steps) from incumbent features. References: random, a \emph{post-hoc} predictor fit to the real update's anchor-set flips, and the oracle. Thresholds are fixed on the anchor set.

\paragraph{Controls and statistics.}
The self-disagreement floor is measured by re-running the incumbent with a different batch size and composition; $\tau$ is the smallest margin above which it falls below $10^{-3}$ on the anchor set. Certificates use Theorem~\ref{thm:cs} at $\alpha=0.05$, checked every 25 audit samples in random order, and the one-sided test of Proposition~\ref{prop:onesided} checked every 5. Our audit streams are random orderings of a finite audit set, that is, sampling without replacement rather than the i.i.d.\ stream of Theorem~\ref{thm:cs}; in 20{,}000 random orderings of pools of 2{,}000 and 8{,}000 inputs with pool leakage 1--10\%, the confidence sequence misses the pool leakage with probability at most 0.029 at $\alpha=0.05$ and 0.007 at $\alpha/4$, so it remains valid for the pool (sequences designed for this setting are tighter; \citealp{waudbysmith2020wor}). The operative certificate is the one computed on a single audit order fixed in advance; the median over 20 random orders is reported only to show how the sample cost varies. Implementation and additional-experiment details are in Appendices~\ref{app:impl} and~\ref{app:more}.

\section{Results}
\label{sec:results}

The results answer four questions in turn. Do declared scopes contain the footprints of real updates (\S\ref{sec:res-leak}, \S\ref{sec:res-structure})? If not, what determines how far an update leaks (\S\ref{sec:res-kl})? Can the remedies a practitioner would try first bring leakage to a certifiable level (\S\ref{sec:res-remedies})? And does anchoring, which targets the quantity identified in \S\ref{sec:res-kl}, make updates certifiable for LoRA (\S\ref{sec:res-anchor}), for other update types and model sizes (\S\ref{sec:res-types}), and over a sequence of updates (\S\ref{sec:res-compose})? The last two subsections mark the limits of the approach: averaged certificates can hide a stratum that an update rewrites (\S\ref{sec:res-edit}), and certificates on decision-like outputs do not extend to open-ended generation (\S\ref{sec:res-beyond}). Unless stated otherwise, numbers are percentages on the development audit set, and $\lambda$ is leakage outside the taxonomy envelope.

\subsection{Declared scopes do not contain update footprints}
\label{sec:res-leak}\label{sec:res-floor}\label{sec:res-public}

Every update we trained changes decisions far outside its declared scope, and for domain updates aggregate accuracy does not reveal it.

\begin{table}[t]
\centering
\caption{Footprints of scoped LoRA updates (percent, development audit set). $m(E)$: envelope mass; effect: intended change inside the envelope; in-scope $\Delta$: in-scope change rate; $\lambda$, $\lambda_\tau$: leakage and decisive leakage; $\kappa$: containment; last column: net out-of-scope accuracy change and out-of-scope correctness churn. Effect is the accuracy gain for med and legal, the gain in Science/Technology articles filed as Business for policy, and the gain in edit efficacy on paraphrases for edit. \emph{null}: incumbent re-batched.}
\label{tab:footprint}
\small
\setlength{\tabcolsep}{4pt}
\begin{tabular}{llrrrrrrr}
\toprule
Model & Update & $m(E)$ & effect & in-scope $\Delta$ & $\lambda$ (s0/s1) & $\lambda_\tau$ & $\kappa$ & net acc.\ / churn out \\
\midrule
Qwen2.5-7B & med & 18.4 & +2.2 & 22.1 & 9.9/8.4 & 8.6 & 33.5 & +1.1 / 7.7 \\
 & legal & 11.1 & +8.1 & 24.2 & 9.5/9.3 & 8.1 & 24.0 & +0.2 / 7.4 \\
 & policy & 8.9 & +82.7 & 23.7 & 21.8/16.3 & 20.5 & 9.7 & $-$10.5 / 18.0 \\
 & edit & 2.2 & +71.4 & 87.6 & 98.1/97.2 & 98.3 & 1.9 & $-$68.2 / 81.8 \\
 & \emph{null} & -- & -- & -- & 0.7 & 0.0 & -- & -- \\
\midrule
Llama-3.1-8B & med & 18.4 & +1.3 & 23.8 & 15.9/14.2 & 13.0 & 25.2 & +0.8 / 12.6 \\
 & legal & 11.1 & +8.8 & 35.9 & 15.6/14.5 & 12.7 & 22.3 & $-$0.5 / 11.8 \\
 & policy & 8.9 & +83.1 & 25.4 & 30.5/32.6 & 28.1 & 7.6 & $-$11.0 / 23.7 \\
 & edit & 2.2 & +85.7 & 91.9 & 81.3/86.7 & 81.7 & 2.4 & $-$52.0 / 65.8 \\
 & \emph{null} & -- & -- & -- & 0.9 & 0.0 & -- & -- \\
\bottomrule
\end{tabular}

\end{table}

\begin{figure}[t]
\centering
\includegraphics[width=\linewidth]{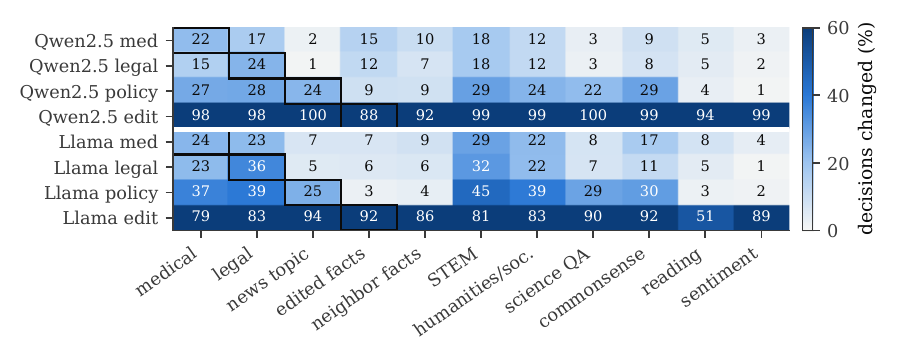}
\caption{Percentage of decisions changed per traffic domain by the controlled LoRA updates (seed 0). Boxes mark the declared scope. Every update changes decisions across most domains. Public fine-tunes: Figure~\ref{fig:heatmap-public}.}
\label{fig:heatmap}
\end{figure}

\paragraph{Floor.} Re-batching the incumbent changes 0.71\% (Qwen) and 0.95\% (Llama) of audit decisions, almost all of them fragile: above the decisive threshold ($\tau=0.40$ and $0.25$ nats, chosen on the anchor set; 96.5\% and 92.4\% of traffic) self-disagreement is 0.04\% and 0.01\%. This floor is the smallest leakage a certificate on raw decisions can resolve.

\paragraph{Leakage.} Table~\ref{tab:footprint} and Figure~\ref{fig:heatmap} give the footprints. The Qwen medical update changes 22.1\% of in-scope and 9.9\% of out-of-scope decisions, fourteen times the floor, and only a third of its footprint lies inside its envelope ($\kappa=33.5\%$); legal behaves alike, and both leak more on Llama (15.9\% and 15.6\%). Preference optimization leaks comparably or more: decision-token DPO leaks 9.0\% (Qwen medical), 17.4\% (Llama medical), and 34.1\% (Qwen policy). For the medical and legal updates on Qwen and Llama, net out-of-scope accuracy moves by at most 1.3 points while 7--13\% of out-of-scope decisions switch correctness, so aggregate metrics do not reveal the footprint. The policy update reaches its goal (83\% of Science/Technology articles refiled as Business), but 90\% of its footprint lies outside news classification and out-of-scope accuracy drops by 7--11 points. The edit update is analyzed in \S\ref{sec:res-edit}. Ten public fine-tunes of the same base models change 6--14\% of audit decisions (a Russian-language adaptation changes 14.1\% of English decisions) while net accuracy moves by at most 1.7 points (Appendix Table~\ref{tab:public}).

\paragraph{Letter decisions as a proxy.} Letter decisions can disagree with written answers \citep{wang2024myanswer}. Under free generation on 2{,}000 out-of-scope queries they match the parsed written answer on 90--96\% of queries, and leakage \emph{rates} measured through generation agree with the letter measurement, although for anchored models the flipped items largely differ (Appendix~\ref{app:more}); \S\ref{sec:res-beyond} certifies written answers directly.

\subsection{Footprints are structured and can be forecast, but not from the scope}
\label{sec:res-structure}\label{sec:res-declare}

Leakage is not spread at random. It concentrates on fragile incumbent decisions that many updates share, and a cheap pilot run forecasts it much better than a description of the update's scope.

\paragraph{Fragility and a shared core.} The probability that an out-of-scope decision changes falls by two orders of magnitude with the incumbent's margin for the domain updates, from 50--60\% at near-ties to 0.1--0.4\% above 8 nats (Appendix Figure~\ref{fig:margin}), as Lemma~\ref{lem:margin} predicts for mild updates. Two seeds of the med, legal, and policy updates share out-of-scope flips with Jaccard overlap 0.61--0.78 (0.05--0.19 under independence), and different updates overlap at 0.15--0.35 (0.04--0.12). Of a domain update's out-of-scope flips, 74--85\% are also made by at least two other models (other updates or public fine-tunes), whereas the policy update adds a large update-specific part (38--53\% unique).

\paragraph{Direction.} The policy rule relabels Science/Technology (option D in AG News) as Business (C), and 82--84\% of its out-of-scope flips move to C and none to D: the model learned a rule about answer letters, which AG News's fixed option order confounds with the intended rule. Shuffling the options per training example keeps the intended change, removes the accuracy damage, and lowers leakage from 21.8\% to 7.9\% (Qwen) and from 30.5\% to 11.4\% (Llama), still an order of magnitude above the floor. We therefore treat this \emph{option-shuffled} policy update as the clean test of a rule-like change in \S\ref{sec:res-anchor}.

\paragraph{Declaring the footprint before training.} Scores that describe what an update is \emph{about} do not predict where it acts (Appendix Table~\ref{tab:decl}): an NL judge of the update's scope and similarity to its training data predict which out-of-scope decisions change with AUROC 0.42--0.58. Incumbent fragility reaches 0.86--0.91 for domain updates (0.64 for policy), and a forecast fit on the flips of a cheap pilot run reaches 0.85--0.96 on seven of eight update and model pairs. Forecasting localizes leakage but does not make it small: holding leakage to 5\% requires declaring 19--40\% of traffic. Shrinking the footprint therefore has to happen in training, which motivates the next three subsections.

\subsection{Leakage tracks out-of-scope KL}
\label{sec:res-kl}

Across all update types, leakage tracks one label-free quantity, the update's KL divergence to the incumbent on out-of-scope inputs. This is the quantity a remedy has to control.

\begin{figure}[t]
\centering
\includegraphics[width=\linewidth]{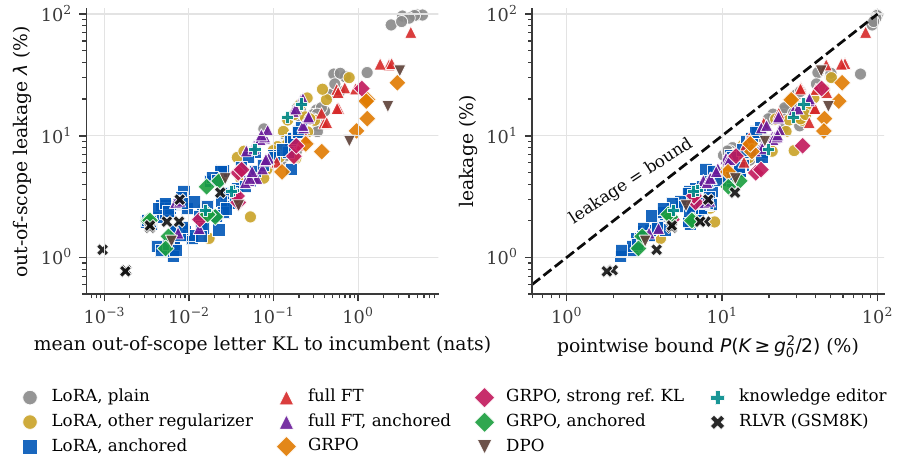}
\caption{Every trained variant (256 variants, six models, all update types, plain and anchored, and all controls; ``other regularizer'' covers KL on generic text, focal distillation, the L2 penalty, replay, and adapter scaling). Left: out-of-scope leakage against the mean out-of-scope KL of the option-letter distribution to the incumbent. Right: leakage against the pointwise bound of Lemma~\ref{lem:pinsker}, computed from the same distributions; points lie below the diagonal.}
\label{fig:kl}
\end{figure}

By Lemma~\ref{lem:pinsker}, a decision can change only where the update moves the letter distribution by at least $g_0^2/2$ in KL. Figure~\ref{fig:kl} shows how much of the observed leakage this explains across the 256 variants we trained, spanning LoRA, full fine-tuning, DPO, GRPO, RLVR, and knowledge editors on six models.

\paragraph{Mean KL and the pointwise bound.} Leakage and mean out-of-scope letter KL have Spearman correlation 0.94, and a log-log fit predicts leakage to within a factor of about 1.4 (residual standard deviation 0.34 in log units). The pointwise bound also uses \emph{where} the KL falls. It is within a factor of 4.6 of the observed leakage (median ratio 0.49) and stays predictive in the range that decides certification: below 6\% leakage it ranks variants with Spearman 0.89, against 0.74 for mean KL, and variants with the same mean KL of 0.01--0.03 nats leak between 1.4\% and 5.2\%. The lemma guarantees that the bound holds; that it is this tight is an empirical finding. The averaged form of Remark~\ref{prop:anchor} is a median 8.0 times the observed leakage and uninformative (at least 100\%) for 53 of the 256 variants.

\paragraph{Robustness of the relation.} The relation holds within each update type (Spearman 0.71--0.96 for mean KL and 0.76--1.00 for the bound, for types with at least eight variants) and within each model (0.96--1.00), so it is not an artifact of pooling. About half of the bound's mass (median 0.51) comes from inputs whose decision did not change. Mean KL is not a sufficient statistic: at matched mean KL, DPO and plain GRPO flip about 0.6 times as many decisions as supervised updates. Still, every method that lowered leakage in our experiments (on-policy RL, a strong in-scope KL, a small learning rate, anchoring) did so by lowering out-of-scope KL, and anchoring is the only one that targets it directly.

\paragraph{A training-time check.} Measured on the anchor split, which the developer holds before any audit, the same quantities predict audit leakage almost as well: Spearman 0.92 for mean KL and 0.95 for the pointwise bound (0.59 and 0.75 below 6\% leakage). Of the 99 variants whose anchor-split bound was below 10\%, 93 leaked less than 5\% on the audit set. In all 256 variants the anchor-split bound was at least the audit leakage, although nothing guarantees this. The anchor-split bound is therefore a label-free, training-time check of whether a candidate is worth auditing, not a substitute for the audit.

\subsection{Customary remedies do not control out-of-scope KL}
\label{sec:res-remedies}

\begin{figure}[t]
\centering
\includegraphics[width=0.92\linewidth]{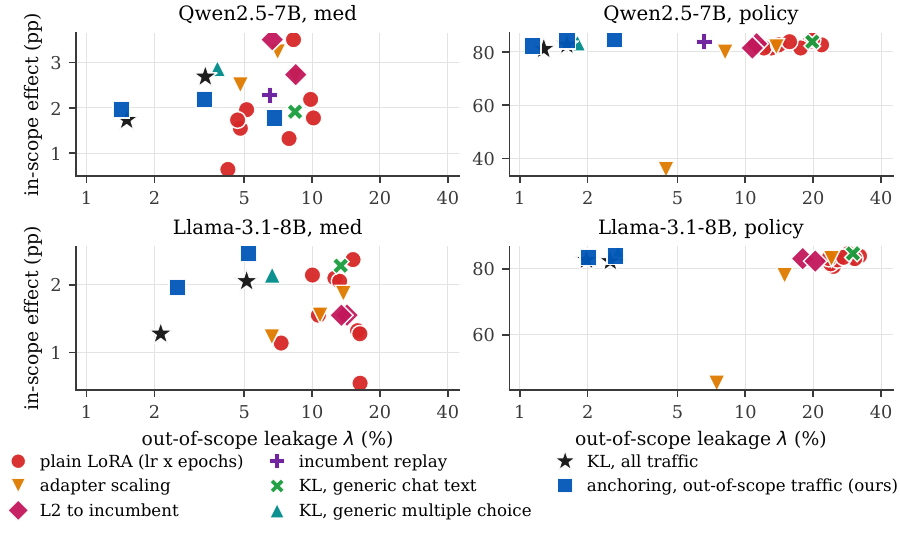}
\caption{In-scope effect versus out-of-scope leakage for every way of trading them off that we tried (development audit set; symmetric-log leakage axis). Only anchoring at the decision position on multiple-choice prompts reaches low leakage without giving up the policy change. Numbers: Appendix Table~\ref{tab:controls}.}
\label{fig:pareto}
\end{figure}

If leakage follows out-of-scope KL, a remedy must keep that KL small on inputs the update is not trained on. The remedies a practitioner would try first constrain something else: the size of the update, the KL on the training prompts, or the KL on inputs in a different format. We compare them with anchoring on the medical and policy updates for Qwen and Llama (Figure~\ref{fig:pareto}; Appendix Tables~\ref{tab:controls} and~\ref{tab:sweep}).

\paragraph{Training less.} Across nine learning-rate $\times$ epoch configurations, plain LoRA never goes below 4.2\% leakage for Qwen medical, 12.1\% for Qwen policy, and 7.3\% and 23.5\% for Llama (the lowest medical values at a medical gain of only $+0.6$ and $+1.1$). Scaling the adapter down (weight interpolation toward the incumbent) reaches its lowest leakage (4.4--7.5\%) at a quarter of the update, where the policy change collapses from 83 to 36--45 points. An L2 penalty toward the incumbent and incumbent replay stay at 6.5--20.4\%.

\paragraph{Constraining the wrong inputs.} Token-level KL on generic chat text barely helps (8.4--30.0\%). Focal distillation \citep{yan2021positive} puts a KL to the incumbent on the \emph{training} prompts, weighted toward inputs the incumbent answers correctly. At weight 1 it leaks 4.2--7.5\% and never certifies, although for the medical update it comes close while keeping the gain (Qwen 5.3\% at $+2.1$; Llama 4.2\% at $+1.5$ with an upper bound of 5.15\%). At weight 10 it certifies, but only by giving up the update: the policy change falls from $+83$ to $+2.4$ and $-0.4$ points, and the medical gain from $+2.2$ to $+1.1$ (Qwen) and from $+1.3$ to $+0.3$ (Llama). The stronger reference-model KL of RL behaves the same way (\S\ref{sec:res-types}): it constrains the prompts where the update is supposed to act and leaves out-of-scope traffic to however the network generalizes.

\paragraph{Where alternatives come close.} For the small medical update, focal distillation at weight 1, a quarter-scaled adapter, plain full fine-tuning at $10^{-6}$, and strong-KL GRPO reach 2.8--6.1\% on Qwen, against 1.4--3.3\% for anchoring. For the policy update, the best alternative that keeps the change (focal distillation at weight 1) leaks about 4.5\% (Qwen) and 7.5\% (Llama), against 1.1--1.6\% and 2.0--2.7\% with anchoring.

\subsection{Anchoring makes LoRA updates certifiable}
\label{sec:res-anchor}

Anchoring (Eq.~\ref{eq:anchor}) is the only intervention we tested that reaches low leakage while keeping the rule-like policy change (Figure~\ref{fig:pareto}), and it makes every anchored LoRA candidate of the domain and policy updates on our two main models certifiable.

\begin{table}[t]
\centering
\caption{Anchoring versus plain LoRA (percent, development audit set; brackets: bootstrap 95\% intervals). Last columns: smallest $\varepsilon\in\{0.5,1,2,5\}\%$ certified by the confidence sequence of Theorem~\ref{thm:cs} at level $\alpha/4$ (Bonferroni over the four candidates) for an anchored candidate in at least half of 20 audit orders, with median samples, and the median samples of the one-sided test of Proposition~\ref{prop:onesided} at that $\varepsilon$ and level $\alpha/16$. $^\dagger$Edit with 400 unedited true facts mixed in (\S\ref{sec:res-edit}). All variants: Appendix Tables~\ref{tab:variants}, \ref{tab:variants-llama}, and~\ref{tab:variants-other}.}
\label{tab:anchor}
\small
\setlength{\tabcolsep}{4pt}
\begin{tabular}{lrlrlll}%
\toprule
& \multicolumn{2}{c}{LoRA} & \multicolumn{2}{c}{anchored ($\beta{=}1$)} & \multicolumn{2}{c}{best certificate at $\alpha/4$} \\
\cmidrule(lr){2-3}\cmidrule(lr){4-5}\cmidrule(lr){6-7}
Update & effect & $\lambda$ [95\% CI] & effect & $\lambda$ [95\% CI] & $\varepsilon$ ($\beta$) & samples: CS / one-sided \\
\midrule
\multicolumn{7}{l}{\emph{Qwen2.5-7B}} \\
med & +2.2 & 9.9 [9.3, 10.4] & +2.2 & 3.3 [3.0, 3.7] & 5\% (10) & 475 / 380 \\
legal & +8.1 & 9.5 [9.0, 10.1] & +6.9 & 2.2 [1.9, 2.5] & 5\% (1) & 575 / 517 \\
policy & +82.7 & 21.8 [21.0, 22.6] & +84.3 & 1.6 [1.4, 1.9] & 2\% (10) & 4{,}437 / 3{,}045 \\
edit$^\dagger$ & +29.3 & 12.1 [11.5, 12.6] & +25.1 & 2.8 [2.5, 3.1] & 5\% (1) & 1{,}575 / 1{,}437 \\
\addlinespace[2pt]
\multicolumn{7}{l}{\emph{Llama-3.1-8B}} \\
med & +1.3 & 15.9 [15.2, 16.6] & +2.5 & 5.2 [4.8, 5.6] & 5\% (10) & 950 / 720 \\
legal & +8.8 & 15.6 [14.9, 16.2] & +11.7 & 3.1 [2.8, 3.4] & 5\% (1) & 1{,}900 / 1{,}620 \\
policy & +83.1 & 30.5 [29.6, 31.4] & +83.9 & 2.7 [2.3, 3.0] & 5\% (10) & 775 / 660 \\
edit$^\dagger$ & +45.9 & 32.1 [31.3, 33.0] & +50.2 & 5.7 [5.3, 6.1] & no & -- \\
\bottomrule
\end{tabular}

\end{table}

\paragraph{Leakage.} Anchoring removes most leakage (Table~\ref{tab:anchor}). On Qwen, leakage falls from 21.8\% to 1.6\% ($\beta=1$) and 1.1\% ($\beta=10$) for policy, from 9.9\% to 3.3\% and 1.4\% for medical, and from 9.5\% to 2.2\% for legal; on Llama, from 30.5\% to 2.7\%, from 15.9\% to 5.2\%, and from 15.6\% to 3.1\%. Every reduction is significant (paired McNemar $p<10^{-100}$). Part of the policy reduction removes the letter confound of \S\ref{sec:res-structure}. Against the option-shuffled policy update, which is not confounded with answer letters, anchoring still cuts leakage about fourfold, from 7.9\% to 1.8\% (Qwen) and from 11.4\% to 2.8\% (Llama), with the change intact ($+83.1$ and $+83.5$), and both anchored versions certify $\varepsilon=5\%$ on the fixed audit order (after 775 and 2{,}550 samples). Anchoring works equally for DPO (2.7\%, 4.4\%, and 1.4\% for the three DPO updates above).

\paragraph{What anchoring needs.} Excluding the envelope from the anchor set does not matter: anchoring on all traffic, in-scope queries included, gives the same leakage at $\beta=1$ and 10. The decision format does most of the work: anchoring on 3{,}000 multiple-choice questions from outside the pool reaches 1.8--6.7\%, against 1.6--5.2\% for in-pool anchors, although in-pool anchors are still significantly better for the medical updates (McNemar $p=0.003$ and $3\cdot10^{-11}$). Anchoring also partly transfers to domains left out of the anchor set (medical: from 10.7\% to 4.8\% and from 19.5\% to 10.1\%, against 4.1\% and 6.4\% when they are included; policy: fully). The envelope's role is to define what is certified.

\paragraph{Cost to the intended change.} Anchored legal and policy updates are non-inferior to plain LoRA at a margin of half the plain effect (TOST at 5\%, both models), and anchored policy DPO gives up 4.7 points of an 86-point change. The small medical effect is not fully preserved on Qwen. On all 6{,}277 medical queries, plain Qwen LoRA gains $+2.8$ points and anchored models $+1.5$ to $+2.2$ (anchored minus plain: $-0.6$ to $-1.3$, significant for four of five seeds); on MedQA, which no model saw, they gain $+1.2$ to $+2.6$ against $+3.1$ to $+3.8$. For Llama, whose plain medical gain is only $+0.9$, anchored models gain as much or more ($+1.2$ to $+1.5$; Appendix Table~\ref{tab:effect}). Anchoring thus trades part of a small in-scope gain for a three- to sevenfold reduction in leakage, and keeps large gains.

\paragraph{Certificates.} No plain LoRA update certifies even $\varepsilon=5\%$. On the single audit order fixed in advance, which is the operative certificate, the six anchored medical, legal, and policy candidates of Table~\ref{tab:anchor} certify $\varepsilon=5\%$ after 200--1{,}750 unlabeled out-of-scope samples, and Qwen policy certifies 2\% after 2{,}325. Over 20 random orders, all six pairs certify $\varepsilon=5\%$ at level $\alpha/4$ after a median of 362--1{,}900 samples with the confidence sequence and 285--1{,}620 with the one-sided test at level $\alpha/16$, and Qwen policy certifies 2\% (4{,}437 and 3{,}045 samples). A router that sends only judge-positive inputs to the update has zero leakage by construction but forfeits 10--25\% of the intended change, because the judge recognizes only 63--77\% of in-scope inputs.

\paragraph{Confirmatory audit.} All choices above were made on the development audit set. On the 11{,}951 confirmatory queries, plain leakage replicates (Qwen medical 10.6\%, policy 20.9\%; Llama 16.0\%, 31.0\%), and all six pairs re-certify at 5\% when retrained with two new seeds (Llama policy with $\beta=1$ rather than the development-best $\beta=10$), with a seed standard deviation of leakage of at most 0.35 points (Appendix Table~\ref{tab:confirm}).

\paragraph{Anchors from other domains.} The anchor set is a random share of the same pool as the audit set, the most favorable case for anchoring. To test a shift, we split the out-of-scope domains into two groups (the MMLU-type knowledge domains with ARC and CommonsenseQA, and all others), anchor on one group, and certify only on the other. With $\beta=10$, seven of the eight shifted runs (two models, two updates, two directions) still certify $\varepsilon=5\%$ on the unseen domains, with 0.9--2.2\% leakage there; the exception, Llama medical anchored without the knowledge domains, leaks 5.5\% on them, against 2.9\% when they are among the anchors. With $\beta=1$ the policy update is unaffected, but medical leakage on the unseen knowledge domains rises to 4.8\% (Qwen) and 10.1\% (Llama). A domain shift between anchors and audit thus costs some certification margin for domain updates, which a stronger anchor weight absorbs; this test keeps the prompt format fixed.

\subsection{Other update types and model sizes}
\label{sec:res-types}

The anchoring result is not specific to LoRA supervised fine-tuning. It carries over to full fine-tuning at a small learning rate, to reinforcement learning on the decision, and to Qwen models from 1.5B to 72B parameters; the one model family where it falls short is Mistral. RL with verifiable rewards on strong incumbents is the one update type that needs no anchoring for a multiple-choice certificate.

\begin{table}[t]
\centering
\caption{Full fine-tuning and reinforcement learning, selected variants (percent, development audit set). Effect and leakage $\lambda$ as in Table~\ref{tab:anchor}; certified: smallest $\varepsilon$ certified by the confidence sequence at level $\alpha/4$ on the audit order fixed in advance, with the samples used. Full FT: all weights; GRPO: decision-token GRPO with the stated weight on KL to the incumbent on training prompts. Seed 0 unless stated. All variants, including the learning rate $3\cdot10^{-6}$ and the 128-anchor runs: Appendix Table~\ref{tab:updtypes-full}.}
\label{tab:updtypes}
\footnotesize
\setlength{\tabcolsep}{3pt}
\begin{tabular}{llrrlrrl}%
\toprule
& & \multicolumn{3}{c}{Qwen2.5-7B} & \multicolumn{3}{c}{Llama-3.1-8B} \\
\cmidrule(lr){3-5}\cmidrule(lr){6-8}
Type & Variant & effect & $\lambda$ & certified & effect & $\lambda$ & certified \\
\midrule
\multicolumn{8}{l}{\emph{med update}} \\
LoRA & plain & +2.2 & 9.9 & -- & +1.3 & 15.9 & -- \\
LoRA & anchored $\beta{=}10$ & +2.0 & 1.4 & 5\% (650) & +2.0 & 2.5 & 5\% (1{,}375) \\
full FT & plain, lr $10^{-6}$ & +3.0 & 6.1 & -- & +1.3 & 12.8 & -- \\
full FT & plain, lr $10^{-5}$ & $-$2.2 & 12.7 & -- & $-$5.3 & 24.1 & -- \\
full FT & anchored $\beta{=}10$, lr $10^{-5}$ & +1.4 & 12.2 & -- & $-$2.4 & 16.7 & -- \\
full FT & anchored $\beta{=}10$, 64 anchors, lr $10^{-6}$ & +1.0 & 1.7 & 5\% (850) & +2.1 & 2.8 & 5\% (2{,}225) \\
GRPO & reference KL 0.04 & +1.9 & 5.0 & -- & +1.6 & 13.8 & -- \\
GRPO & reference KL 1 & +1.2 & 2.8 & 5\% (1{,}850) & +1.0 & 5.0 & -- \\
GRPO & ref.\ KL 0.04 + anchored $\beta{=}1$ & +2.4 & 2.2 & 5\% (800) & +1.2 & 4.3 & -- \\
GRPO & reference KL 0.04, seed 1 & +1.3 & 7.4 & -- & +2.4 & 10.9 & -- \\
GRPO & reference KL 1, seed 1 & +0.9 & 3.2 & 5\% (2{,}625) & +1.0 & 5.3 & -- \\
GRPO & anchored, seed 1 & +1.6 & 2.3 & 5\% (1{,}050) & +2.0 & 3.8 & 5\% (5{,}375) \\
\midrule
\multicolumn{8}{l}{\emph{policy update}} \\
LoRA & plain & +82.7 & 21.8 & -- & +83.1 & 30.5 & -- \\
LoRA & anchored $\beta{=}10$ & +82.3 & 1.1 & 2\% (2{,}325) & +83.5 & 2.0 & 5\% (1{,}675) \\
full FT & plain, lr $10^{-6}$ & +82.7 & 15.4 & -- & +81.9 & 22.7 & -- \\
full FT & plain, lr $10^{-5}$ & +83.9 & 38.3 & -- & +83.9 & 71.0 & -- \\
full FT & anchored $\beta{=}10$, lr $10^{-5}$ & +82.7 & 14.4 & -- & +83.9 & 20.2 & -- \\
full FT & anchored $\beta{=}10$, 64 anchors, lr $10^{-6}$ & +81.9 & 1.6 & 5\% (650) & +83.5 & 2.9 & 5\% (2{,}150) \\
GRPO & reference KL 0.04 & +80.3 & 19.7 & -- & +82.7 & 27.3 & -- \\
GRPO & reference KL 1 & +41.7 & 2.1 & 5\% (1{,}100) & +49.6 & 8.3 & -- \\
GRPO & ref.\ KL 0.04 + anchored $\beta{=}1$ & +84.6 & 1.5 & 5\% (450) & +83.5 & 2.0 & 5\% (1{,}025) \\
GRPO & reference KL 0.04, seed 1 & +78.0 & 8.6 & -- & +82.7 & 19.2 & -- \\
GRPO & reference KL 1, seed 1 & +80.3 & 6.8 & -- & +83.1 & 24.4 & -- \\
GRPO & anchored, seed 1 & +82.7 & 1.2 & 2\% (1{,}625) & +81.9 & 2.0 & 5\% (425) \\
\bottomrule
\end{tabular}

\end{table}

\paragraph{Full fine-tuning.} Plain full-fine-tuning leakage grows with the learning rate (Table~\ref{tab:updtypes}, Appendix Table~\ref{tab:updtypes-full}): for Qwen medical it is 6.1, 9.6, and 12.7\% at $10^{-6}$, $3\cdot10^{-6}$, and $10^{-5}$, and 15.4, 24.8, and 38.3\% for Qwen policy; on Llama it is 12.8--24.1\% and 22.7--71.0\%. At the smallest rate it leaks less than LoRA (9.9\% and 21.8\% on Qwen); at the larger rates it leaks more and overfits the medical update, whose effect turns negative at $10^{-5}$ ($-2.2$ and $-5.3$ points), in line with the observation that full fine-tuning forgets more than LoRA \citep{biderman2024lora}. At $10^{-5}$ anchoring helps only partly (11.7--20.2\% with $\beta\in\{1,10\}$): the anchor KL at the end of training (mean over the last five logged steps) is 0.12--0.26 nats whatever $\beta$, against 0.008--0.05 for anchored LoRA, and leakage on anchor-split traffic is only about 1--3 points below the audit set, so the update does not memorize the anchors but never reaches a low anchor KL. Smaller steps fix this. At $10^{-6}$ with $\beta=10$ and 64 anchors per step, anchored leakage is 1.7\% and 1.6\% on Qwen and 2.8\% and 2.9\% on Llama, against 6.1--22.7\% for plain full fine-tuning at the same rate; all four certify $\varepsilon=5\%$ on the fixed audit order (after 650--2{,}225 samples) and keep the policy change, and the medical gain is kept on Llama ($+2.1$ against $+1.3$ plain) but reduced on Qwen ($+1.0$ against $+3.0$). At $3\cdot10^{-6}$ anchored leakage is 4.1--9.8\% whether 64 or 128 anchors are used per step, and the out-of-scope letter KL stays at 0.05--0.08 nats, against 0.007--0.013 at $10^{-6}$ (Appendix~\ref{app:more}): more anchors do not substitute for a small enough learning rate.

\paragraph{Reinforcement learning.} Decision-token GRPO with the customary reference KL leaks less than LoRA supervised training in all eight matched runs (5.0--27.3\% against 8.4--32.6\% over two seeds, both updates, and both models), in line with the observation that on-policy RL forgets less \citep{shenfeld2025rl}. It is still 7 to 29 times the floor and varies strongly across seeds (Qwen policy: 19.7\% and 8.6\%). A stronger KL to the incumbent on the training prompts is not a reliable substitute for anchoring: on seed 0 a 25 times larger weight cuts leakage to 2.1--8.3\% but reduces the policy change to 52--60\% of the plain one ($+41.7$ and $+49.6$ against $+80.3$ and $+82.7$), and on seed 1 it keeps the policy change but leaks 6.8\% and 24.4\% on policy (3.2\% and 5.3\% on medical). With the customary reference weight, anchored runs keep the policy change ($+81.9$ to $+84.6$) with 1.2--2.0\% leakage on policy and 2.2--4.3\% on medical, across both seeds and models, and their medical gains are within 0.5 points of plain GRPO on Llama ($+1.2$ and $+2.0$ against $+1.6$ and $+2.4$) and at least as large on Qwen. Seven of eight certify $\varepsilon=5\%$: Llama medical seed 0, at 4.3\%, does not, and Qwen policy seed 1 certifies 2\%.

\begin{table}[t]
\centering
\caption{RLVR on GSM8K (percent). Train reward: mean over the first and last ten steps. GSM8K: accuracy change and fraction of changed final answers on 800 test problems (in scope). MC $\lambda$: leakage on multiple-choice traffic outside the declared mathematics scope. NQ: out-of-scope free-form answers whose substance changed (semantic judge; floor 4.6--5.9\%). Certified: smallest $\varepsilon$ certified on MC traffic.}
\label{tab:rlvr}
\footnotesize
\setlength{\tabcolsep}{4pt}
\begin{tabular}{lrrrrrl}%
\toprule
Variant & \hd{train\\reward} & \hd{GSM8K\\acc.\ $\Delta$} & \hd{GSM8K\\changed} & \hd{MC\\$\lambda$} & \hd{NQ changed\\(semantic)} & \hd{MC\\certified} \\
\midrule
\multicolumn{7}{l}{\emph{Qwen2.5-1.5B}} \\
RLVR & 0.32$\to$0.66 & +8.1 & 38.5 & 3.4 & 36.3 & 5\% \\
RLVR + anch. (MC) & 0.30$\to$0.62 & +8.8 & 39.4 & 1.8 & 29.4 & 5\% \\
RLVR + anch. (MC + free-form) & 0.29$\to$0.61 & +9.1 & 38.0 & 1.8 & 12.3 & 5\% \\
\addlinespace[2pt]
\multicolumn{7}{l}{\emph{Qwen2.5-7B}} \\
RLVR & 0.76$\to$0.91 & $-$0.9 & 6.9 & 0.8 & 8.9 & 2\% \\
RLVR + anch. (MC) & 0.74$\to$0.87 & $-$0.1 & 6.0 & 0.8 & 6.4 & 2\% \\
\addlinespace[2pt]
\multicolumn{7}{l}{\emph{Llama-3.1-8B}} \\
RLVR & 0.79$\to$0.93 & +1.7 & 11.6 & 2.0 & 12.1 & 5\% \\
RLVR + anch. (MC) & 0.79$\to$0.91 & $-$0.2 & 11.2 & 1.2 & 5.8 & 2\% \\
RLVR strong & 0.80$\to$0.90 & +1.2 & 12.5 & 2.0 & 11.7 & 5\% \\
RLVR strong + anchored & 0.79$\to$0.89 & +1.9 & 11.8 & 3.0 & 8.7 & 5\% \\
\bottomrule
\end{tabular}

\end{table}

\paragraph{RL with verifiable rewards.} On Qwen2.5-1.5B-Instruct, which solves 63.8\% of our GSM8K problems, 300 steps of GRPO with free generation and a verifiable reward raise GSM8K accuracy by 8.1 points (Table~\ref{tab:rlvr}). This update leaks 3.4\% on multiple-choice traffic and changes the substance of 36.3\% of out-of-scope NQ answers, against a floor of 5.9\% (re-batching alone changes 13.1\% of this model's final math answers). Anchoring on multiple-choice prompts lowers multiple-choice leakage to 1.8\% but NQ changes only to 29.4\%; adding token-level anchoring on out-of-scope free-form prompts (1{,}500 NQ-open training questions with the incumbent's answers) keeps the full GSM8K gain ($+9.1$) and cuts NQ changes to 12.3\%. Anchors must therefore match the format of the traffic one wants to protect. The 7B and 8B incumbents already solve 86.5--91.4\% of the problems, so RLVR raises their training reward (from 0.74--0.80 over the first ten steps to 0.87--0.93 over the last ten) but GSM8K accuracy by at most 1.9 points. Their multiple-choice footprint is small, 0.8--2.0\% for plain runs, with a small out-of-scope letter KL (Figure~\ref{fig:kl}), and certifies $\varepsilon=2$--$5\%$ without anchoring; this is the one update type that needs no anchoring for a multiple-choice certificate. Anchoring gives no consistent multiple-choice benefit here (Llama: 2.0\% to 1.2\% for the default run, 2.0\% to 3.0\% for the strong run) and a mixed GSM8K effect on Llama ($-2.0$ and $+0.6$ points against the plain default and strong runs), but it reduces out-of-scope free-form changes from 8.9--12.1\% to 5.8--8.7\% of NQ answers against a floor of about 5\%. On free-form traffic, where RLVR operates, even this mild update changes the substance of about one in ten out-of-scope answers.

\paragraph{Model size and a third family.} Qwen2.5-1.5B-Instruct leaks most among the Qwen models (15.4\%, 17.0\%, and 26.7\% for the medical, legal, and policy updates, floor 1.2\%; one seed) and gains $+9.7$ points on legal. Anchoring with $\beta=10$ cuts leakage to 2.9\%, 2.5\%, and 2.4\%, keeps the legal gain ($+9.1$) and the policy change, and all three certify $\varepsilon=5\%$. Plain medical leakage falls with size (8.4--9.9\%, 7.5--8.0\%, and 6.9--7.5\% at 7B, 32B, and 72B, two seeds each); policy leakage does not (16.3--21.8\%, 6.9--9.9\%, and 13.8--15.1\%). At 72B leakage is 16 to 35 times the floor of 0.44\%. Anchoring cuts it to 2.5--2.8\% (medical) and 1.5--1.7\% (policy) with the effect kept ($+3.1$ to $+3.2$ against $+3.5$ to $+3.7$, and $+69.3$ to $+70.5$ against $+70.1$ to $+70.9$), and all four anchored 72B runs certify $\varepsilon=5\%$ on the fixed audit order after 375--1{,}200 samples; at 32B it cuts leakage to 2.2--2.3\% and 1.0--1.1\% (Appendix Table~\ref{tab:scale}). Plain LoRA on Mistral-7B leaks more than on any other incumbent (17.0\% medical, 33.0\% policy) and responds less to anchoring at the default learning rate (10.9\% and 9.5\%). At learning rate $3\cdot10^{-5}$ anchored Mistral reaches 6.6\% and 4.2\% while plain LoRA at that rate still leaks 14.9\% and 23.1\%, but no Mistral variant certifies $\varepsilon=5\%$ in most audit orders (Appendix~\ref{app:more}).

\begin{table}[t]
\centering
\caption{Held-out certificates for the other update types and model sizes (percent; confirmatory audit set, never used during development; candidates rerun with the development configuration and seed). Certified: smallest $\varepsilon$ certified by the confidence sequence at level $\alpha/4$ on one fixed audit order, with samples used; ``no'' gives the final upper bound. Effect: in-scope effect with a bootstrap 95\% interval.}
\label{tab:confirm-types}
\small
\setlength{\tabcolsep}{4pt}
\begin{tabular}{llrlr}%
\toprule
Update & Candidate & $\lambda$ & certified (samples) & effect [95\% CI] \\
\midrule
\multicolumn{5}{l}{\emph{Qwen2.5-7B}} \\
med & full FT, anchored ($10^{-6}$) & 2.0 & 5\% (600) & +0.1 [$-$0.8, +1.1] \\
policy & full FT, anchored ($10^{-6}$) & 1.8 & 5\% (975) & +76.4 [+71.3, +81.5] \\
med & GRPO, anchored & 2.7 & 5\% (600) & $-$0.1 [$-$1.1, +0.9] \\
policy & GRPO, anchored & 1.6 & 5\% (350) & +75.3 [+70.2, +80.0] \\
\addlinespace[2pt]
\multicolumn{5}{l}{\emph{Llama-3.1-8B}} \\
med & full FT, anchored ($10^{-6}$) & 2.6 & 5\% (1{,}200) & +1.5 [+0.4, +2.7] \\
policy & full FT, anchored ($10^{-6}$) & 2.7 & 5\% (950) & +80.7 [+76.0, +85.1] \\
med & GRPO, anchored & 4.3 & no (5.3) & +1.8 [+0.5, +3.0] \\
policy & GRPO, anchored & 2.1 & 5\% (1{,}100) & +81.1 [+76.4, +85.5] \\
\addlinespace[2pt]
\multicolumn{5}{l}{\emph{Qwen2.5-1.5B}} \\
med & LoRA, anchored ($\beta{=}10$) & 2.9 & 5\% (1{,}800) & +0.9 [$-$0.2, +2.2] \\
legal & LoRA, anchored ($\beta{=}10$) & 2.9 & 5\% (1{,}850) & +6.2 [+4.1, +8.3] \\
policy & LoRA, anchored ($\beta{=}10$) & 2.5 & 5\% (1{,}625) & +77.5 [+72.7, +82.2] \\
\addlinespace[2pt]
\multicolumn{5}{l}{\emph{Qwen2.5-72B}} \\
med & LoRA, anchored ($\beta{=}1$) & 2.1 & 5\% (550) & +4.1 [+2.8, +5.5] \\
policy & LoRA, anchored ($\beta{=}1$) & 1.8 & 5\% (350) & +67.6 [+61.5, +73.5] \\
\bottomrule
\end{tabular}

\end{table}

\paragraph{Held-out certificates.} We rerun the anchored candidates of full fine-tuning, GRPO, and the 1.5B and 72B models with their development configuration and seed and evaluate them only on the confirmatory set (Table~\ref{tab:confirm-types}). Anchored full fine-tuning at $10^{-6}$ leaks 1.8--2.7\% there, anchored GRPO 1.6--4.3\%, the anchored 1.5B updates 2.5--2.9\%, and the anchored 72B updates 1.8--2.1\%. Twelve of the thirteen candidates certify $\varepsilon=5\%$ on a fixed audit order after 350--1{,}850 samples, which is every candidate that certified during development. The thirteenth, Llama medical GRPO, did not certify during development either (4.3\%) and ends at an upper bound of 5.3\%. The policy change replicates ($+67.6$ to $+81.1$), and the 1.5B legal gain is smaller on this set ($+6.2$ against $+9.1$). The small medical gains are within noise on this set for the Qwen full-fine-tuning and GRPO candidates and for the 1.5B model (intervals include zero), so those three certificates cover updates whose intended effect we cannot confirm.

\subsection{Certificates compose over a sequence of updates}
\label{sec:res-compose}

Per-update certificates matter for continual deployment only if they compose, and Theorem~\ref{thm:compose} says they do. Applying the medical, legal, policy, and edit (with true facts) updates in sequence, each on the merged previous model, drift from the original model outside the union of declared scopes reaches 19.0\% (Qwen) and 41.1\% (Llama) after four plain updates, but only 4.3\% and 5.8\% after four anchored ones, below the certified bounds of Theorem~\ref{thm:compose} (12.0\% and 17.6\%; 5.2\% and 6.8\% when drift is certified directly). The anchored edit step is weaker ($+32.4$ and $+51.7$ points against $+44.8$ and $+60.2$). In a second order on Qwen (policy, edit, legal, medical), anchored drift stays at or below 4.1\% while plain drift peaks at 56.3\%. All steps share one audit set that no chain model was trained or selected on, as Theorem~\ref{thm:compose} requires. The chains were not rerun on the confirmatory set.

\subsection{Edits, and what averaged certificates hide}
\label{sec:res-edit}

Trained only on counterfactual targets, every target is the option the incumbent believes false, and the update learns to avoid the incumbent's answer everywhere: out-of-scope accuracy falls to 7.6\% on Qwen, below the 28.6\% chance rate (Appendix Table~\ref{tab:edit}). Mixing in 400 true facts removes this rule on Qwen and weakens it on Llama (leakage 12.1\% and 32.1\%) at the cost of a smaller edit, and anchoring brings leakage to 2.8\% and 5.7\%. MEMIT \citep{meng2023memit} and AlphaEdit \citep{fang2025alphaedit}, re-implemented for these models (Appendix~\ref{app:more}), edit with 56--73 points of efficacy and leak 2.4--3.6\% on Qwen (both certify $\varepsilon=5\%$) but 7.7--18.2\% on Llama, depending on the covariance weight. The edit also exposes a limit of averaged certificates: the anchored counterfactual-only update leaks only 5.3\% (Qwen) and 7.3\% (Llama) overall yet rewrites 89--90\% of the neighborhood stratum, which MEMIT changes at 4\% (Qwen) and 13--16\% (Llama). Strata that matter should be declared and certified separately.

\subsection{Beyond letter decisions}
\label{sec:res-beyond}

All certificates so far audit letter decisions. We now test how far they extend: to the answers the models write, to short free-form answers, to refusals, and to open-ended chat. The certificates largely carry over to written answers to the same queries. Beyond that, the self-disagreement floor rises with the length of the output, and on open-ended chat neither an update nor even a floor control can be certified.

\begin{table}[t]
\centering
\caption{Certificates on written answers (percent; confirmatory audit set, 3{,}000 out-of-scope queries per update, one audit order fixed in advance). Models answer by free greedy generation; an answer that changes, appears, or disappears counts as leakage. Floor: the incumbent re-generated with another batch size, measured once per model on the queries of the medical row. Anchored (letter): the letter-decision leakage of the same anchored model on the same queries. Last columns: samples needed to certify $\varepsilon=5\%$ with the confidence sequence (level $\alpha/4$) and the one-sided test (level $\alpha/16$); ``no'' gives the final upper bound.}
\label{tab:gencert}
\small
\setlength{\tabcolsep}{4pt}
\begin{tabular}{lrrrrrl}%
\toprule
Update & floor & plain & anchored & anchored (letter) & CS: $n$ to certify 5\% & one-sided: $n$ \\
\midrule
\multicolumn{7}{l}{\emph{Qwen2.5-7B}} \\
med & 0.2 & 10.4 & 1.8 & 1.4 & 700 & 415 \\
legal & -- & 9.1 & 2.3 & 2.3 & 1100 & 995 \\
policy & -- & 35.2 & 1.6 & 1.1 & 625 & 415 \\
\addlinespace[2pt]
\multicolumn{7}{l}{\emph{Llama-3.1-8B}} \\
med & 0.5 & 27.7 & 3.2 & 2.6 & 1575 & 765 \\
legal & -- & 21.0 & 3.4 & 2.9 & no (5.1) & 2580 \\
policy & -- & 28.6 & 2.2 & 1.8 & 350 & 305 \\
\bottomrule
\end{tabular}

\end{table}

\paragraph{Written answers.} Anchoring trains the letter decision that the certificate audits, so we also certify what the models write. We retrain the six anchored LoRA candidates, let every model answer 3{,}000 out-of-scope confirmatory queries by free generation, and count any change in the parsed answer, including one that appears or disappears, as leakage (Table~\ref{tab:gencert}). Plain updates change 9.1--35.2\% of written answers; in three cases this exceeds their letter leakage because the plain update stops stating an answer within 24 tokens (7.6--18.4\% of answers), a format leak, and where both answers parse, written and letter leakage agree. Anchored updates change 1.6--3.4\%, against a generation floor of 0.2--0.5\%, and these are substantive changes: at most 0.1\% of anchored answers lose or gain a parseable answer. On one audit order fixed in advance, five of six certify $\varepsilon=5\%$ with the confidence sequence after 350--1{,}575 samples; the sixth, Llama legal, ends at an upper bound of 5.1\%, and all six certify with the one-sided test after 305--2{,}580 samples. The anchored models' letter leakage on the same queries (1.1--2.9\%) is at or slightly below their written-answer leakage, so letter certificates certify the letter decision; on this traffic the written-answer certificates hold as well.

\paragraph{Free-form answers.} We audit the medical and policy updates on open-ended questions outside every scope: 800 GSM8K problems solved step by step \citep{cobbe2021gsm8k} (decision: the stated final number) and 2{,}000 NQ-open questions \citep{lee2019latent} answered with a short phrase (decision: the answer; changed answers are judged for semantic equivalence by Qwen2.5-32B-Instruct). Free-form decisions have a much higher floor than letter decisions. Re-generating the incumbent with another batch size changes 4.0--6.8\% of final math answers and 4.6--5.2\% of factual answers; a zero-effect adapter under identical batching changes none, and random adapters with tiny weights change about as many as re-batching (2.9--7.4\% of math answers). Free-form certificates must be read relative to this floor. Plain updates change 8.0--16.1\% of math answers and the substance of 31--61\% of factual answers, while accuracy moves by at most about 2 and 5 points; the plain Qwen policy model even stops giving short answers (78 instead of 22 characters; Appendix Table~\ref{tab:freeform}). Anchoring on multiple-choice traffic transfers to both (4.1--12.5\% and 9.8--26.4\%). Adding token-level KL on free-form training prompts brings math answers within about 2 points of the floor and factual answers to 8.7--21.1\%, still above it.

\paragraph{Refusals.} On 720 unsafe prompts with LLM-judged refusals, the medical update changes 1.4--2.1\% of refusal decisions (floor 0.3--1.0\%), and anchoring does not measurably change this (Appendix Table~\ref{tab:safety}).

\begin{table}[t]
\centering
\caption{Open-ended chat traffic (percent of 2{,}000 OpenAssistant prompts outside each update's scope). String: the response text differs; substance: Qwen2.5-32B-Instruct judges that the response no longer gives substantively the same content. Adapters are the saved candidates of Table~\ref{tab:gencert} and the free-form audit.}
\label{tab:chat}
\small
\setlength{\tabcolsep}{4pt}
\begin{tabular}{lrrrr}
\toprule
 & \multicolumn{2}{c}{Qwen2.5-7B} & \multicolumn{2}{c}{Llama-3.1-8B} \\
\cmidrule(lr){2-3}\cmidrule(lr){4-5}
Variant & string & substance & string & substance \\
\midrule
incumbent, other batch size (floor) & 82.8 & 21.4 & 32.0 & 11.4 \\
tiny random adapter ($\sigma_B{=}10^{-3}$) & 86.3 & 23.2 & 77.3 & 27.6 \\
medical, plain & 99.4 & 42.2 & 98.7 & 51.3 \\
medical, anchored ($\beta{=}10$) & 87.9 & 24.3 & 85.8 & 31.9 \\
medical, anchored ($\beta{=}1$) + free-form KL & 93.8 & 28.5 & 93.3 & 39.5 \\
legal, plain & 99.6 & 40.3 & 97.9 & 44.6 \\
legal, anchored ($\beta{=}1$) & 91.1 & 25.2 & 89.2 & 35.5 \\
policy, plain & 99.8 & 45.1 & 97.6 & 45.1 \\
policy, anchored ($\beta{=}10$) & 86.3 & 23.3 & 78.1 & 27.6 \\
policy, anchored ($\beta{=}1$) + free-form KL & 87.1 & 24.3 & 82.8 & 29.6 \\
\bottomrule
\end{tabular}

\end{table}

\paragraph{Open-ended chat.} Finally, we audit the domain and policy updates on 2{,}000 first-turn English prompts from OpenAssistant \citep{kopf2023openassistant}, answered greedily with up to 256 tokens. A response counts as changed when Qwen2.5-32B-Instruct judges that it no longer gives substantively the same content as the incumbent's response; prompts the same judge places in an update's scope (at most 2.8\%) are excluded. Here the floor dominates (Table~\ref{tab:chat}). Re-generating the incumbent with another batch size changes the substance of 21.4\% (Qwen) and 11.4\% (Llama) of responses, and a tiny random adapter changes 23.2\% and 27.6\%, because greedy decoding of long responses amplifies small numerical differences. Plain updates change 40.3--51.3\%, 1.6 to 1.9 times the tiny-adapter floor. Anchored updates change 23.3--25.2\% on Qwen, at the tiny-adapter floor, and 27.6--35.5\% on Llama, at or somewhat above it; free-form anchoring at $\beta=1$ (24.3--39.5\%) does no better than decision anchoring at $\beta=10$. Anchoring thus removes most of an update's excess change over the tiny-adapter floor for five of the six updates (about half for Llama legal), but no variant, not even the floor controls, certifies $\varepsilon=5\%$ or $10\%$. Our certificates are therefore certificates for decision-like outputs.

\section{Discussion}
\label{sec:discussion}

\paragraph{What a certificate does and does not say.} A footprint certificate bounds how often decisions change outside a committed region on a stated traffic distribution. Whether in-scope changes are improvements remains the job of label-based tests \citep{balachandran2026discern,zhang2026slicewise}, and beneficial out-of-scope changes count as leakage by design: plain medical LoRA raises out-of-scope accuracy on Qwen by 1.1 points (Table~\ref{tab:footprint}), and anchoring removes such transfer along with harmful change. An average over broad traffic can hide a stratum that an update rewrites completely (\S\ref{sec:res-edit}), so strata that matter should be declared and certified separately, splitting $\alpha$. The main risk is reading ``certified'' as ``safe'': a certificate says nothing about the correctness of in-scope changes, unsampled traffic, or open-ended generation, so envelope mass, audit distribution, strata, the self-disagreement floor, and the type of output covered should accompany any certified $\varepsilon$.

\paragraph{Why anchoring works, and what it requires.} By Lemma~\ref{lem:pinsker}, an update can change an out-of-scope decision only where it moves the letter distribution far enough relative to the incumbent's gap, and observed leakage tracks that KL across every update type (Figure~\ref{fig:kl}). Remedies that constrain the size of the update or the KL on the training prompts leave the out-of-scope KL to however the network generalizes; anchoring constrains it directly. Two conditions follow. The anchor prompts must share the deployment's decision format (their topic matters less, and generic chat text does not work), and the optimizer must actually reach a small anchor KL: full fine-tuning at $10^{-5}$ or $3\cdot10^{-6}$ and Mistral at the default LoRA rate do not, and they do not certify. Anchors must also match the format of the traffic to protect (\S\ref{sec:res-types}).

\section{Limitations}
\label{sec:limitations}

\paragraph{Decision-like outputs only.} Our main traffic is multiple-choice and classification benchmarks with a label-free letter decision, whereas enterprise traffic is mostly free-form. For short free-form answers the numerical floor is about five to seven times higher than for letter decisions, and anchored updates stay above it. For open-ended chat, re-batching alone changes the substance of 11--21\% of responses and a tiny weight perturbation 23--28\%, so no update can certify leakage at a few percent (\S\ref{sec:res-beyond}). Certifying open-ended behavior will need batch-invariant inference \citep{he2025nondeterminism}, decision-level summaries of responses, or certificates of the excess over a perturbation floor. The free-form and chat results also depend on an LLM judge (Qwen2.5-32B-Instruct) whose error rate we did not measure against human labels.

\paragraph{The floor limits what can be certified.} Letter decisions have a self-disagreement floor of 0.4--1.2\% across our six models, so certificates much below about 1\% per update are out of reach without batch-invariant inference. Because the lifetime bound of Theorem~\ref{thm:compose} grows linearly in the number of updates, it becomes vacuous after roughly 20 updates at $\varepsilon=5\%$.

\paragraph{One benchmark pool, one prompt format.} Anchor and audit sets are random splits of one benchmark pool in one prompt format; we tested only a shift between domain groups (\S\ref{sec:res-anchor}), not temporal drift of real traffic or a change of prompt format. The certificate assumes i.i.d.\ audit traffic independent of every model it compares; our audit streams are random orders of a finite set, checked by simulation (\S\ref{sec:setup}).

\paragraph{Updates and effects.} Our updates are single-task and use at most a few thousand examples. Large multi-task post-training, RL from multi-turn deployment feedback, and strong reasoning-focused RL on large models are not covered (our largest RLVR gain is 9 points on a 1.5B model). The medical effects are small (a few points), so three held-out certificates (\S\ref{sec:res-types}) cover updates whose intended effect we cannot confirm, and on Qwen anchoring costs part of the medical gain. The large policy effect comes from a rule-like relabeling, and its headline reductions partly remove a confound with answer letters; the option-shuffled variant gives the cleaner estimate (\S\ref{sec:res-anchor}).

\paragraph{Seeds and coverage.} Most results use two seeds and key anchored variants three; Mistral, all 1.5B results, and RLVR use one, and the strong-reference-KL GRPO result rests on two seeds that behave differently. Anchoring does not reach a certifiable level on Mistral in most audit orders, and the edit update on Llama does not certify. We did not compare against parameter-isolation methods such as orthogonal-subspace LoRA, or against a router learned on the anchor split.

\section{Conclusion}
\label{sec:conclusion}

Declaring where an update will act, and certifying that it stayed there, is feasible for decision-like outputs. Narrow updates of nearly every kind we trained move decisions far outside their declared scope, often invisibly to aggregate metrics, and leakage follows the update's KL to the incumbent on out-of-scope inputs, which the customary remedies do not target. Decision anchoring on unlabeled prompts in the deployment format does, and makes 5\% certificates reachable from a few hundred to a few thousand unlabeled samples for LoRA, full fine-tuning, GRPO, and models from 1.5B to 72B. This suggests a recipe: commit the envelope and the candidate set before training, anchor at a learning rate where the anchor KL becomes small, screen candidates with the anchor-split pointwise bound, certify on audit traffic that no candidate was trained or selected on at a level split over the candidates, and report the certificate with its envelope mass, audit distribution, strata, and floor. Certifying open-ended generation relative to a perturbation floor remains open.

\clearpage
\renewcommand{\floatpagefraction}{0.5}
\appendix
\section{Proofs}
\label{app:proofs}

\paragraph{Lemma~\ref{lem:margin}.}
Let $i=d_{f_0}(x)$ and $j=d_{f_1}(x)\neq i$. Then $z_{f_1,j}\ge z_{f_1,i}$ and, writing $e=z_{f_1}-z_{f_0}-s\mathbf{1}$, $z_{f_0,i}-z_{f_0,j}= (z_{f_1,i}-z_{f_1,j}) - (e_i-e_j)\le |e_i|+|e_j|\le 2\delta(x)$. Since $\mu_{f_0}(x)\le z_{f_0,i}-z_{f_0,j}$, the claim follows. \qed

\paragraph{Theorem~\ref{thm:cs}.}
For fixed $p$, $M_t(p)=\int_0^1 \prod_{s\le t} \frac{q^{Z_s}(1-q)^{1-Z_s}}{p^{Z_s}(1-p)^{1-Z_s}}\,dq$ is a mixture of likelihood-ratio martingales. It is therefore a nonnegative martingale with $M_0(p)=1$ under $\mathrm{Bernoulli}(p)$. Ville's inequality \citep{ville1939etude} gives $P(\exists t: M_t(\lambda)\ge1/\alpha)\le\alpha$, and $\lambda>U_t$ implies $M_t(\lambda)\ge 1/\alpha$. This proves validity. Since $M_t(\hat\lambda_t)\le 1$ (the mixture is at most the maximized likelihood ratio), $U_t\ge\hat\lambda_t$ and the confidence set is a nonempty interval. For the bound, $B(1+S,1+t-S)=\left[(t+1)\binom{t}{S}\right]^{-1}$ and $\binom{t}{S}\le e^{tH(\hat\lambda_t)}$, where $H$ is the binary entropy in nats. Since $-S\log p-(t-S)\log(1-p)=t\left[H(\hat\lambda_t)+\mathrm{kl}(\hat\lambda_t\|p)\right]$, we get $\log M_t(p)\ge t\,\mathrm{kl}(\hat\lambda_t\|p)-\log(t+1)$. Every $p\ge\hat\lambda_t$ with $t\,\mathrm{kl}(\hat\lambda_t\|p)>\log((t+1)/\alpha)$ therefore lies outside the confidence set. For $S_t=0$, $\mathrm{kl}(0\|p)=-\log(1-p)\ge p$. \qed

\paragraph{Theorem~\ref{thm:compose}.}
If $d_{f_0}(x)\neq d_{f_T}(x)$, some consecutive pair differs at $x$, that is, $\Delta_t(x)=1$ for some $t\le T$. If moreover $x\notin\mathcal{U}_T$, then $x\notin E_t$ for that $t$. The event is therefore contained in $\bigcup_t\{\Delta_t=1,x\notin E_t\}$, and the union bound gives the first claim; it requires no independence between steps. For the second claim, condition on the pair $(f_{t-1},f_t)$. Because the audit traffic is independent of the pair, the step-$t$ indicators are i.i.d.\ Bernoulli given the pair, and Theorem~\ref{thm:cs} shows that the step-$t$ sequence fails with probability at most $\alpha_t$. The union bound over steps needs no independence between the steps, and $\sum_{t\ge1}\alpha_t=\alpha$. The same audit set may therefore serve several steps as long as no model in the chain was trained or selected using it. If a step promotes one of $K$ candidates by a certificate on the shared set, the promoted model is a function of the audit data and the conditional i.i.d.\ property fails. Every model that can be certified at step $t$ is then determined by a promotion history of length $t-1$ together with the step-$t$ candidate, at most $K^{t}$ fixed pairs $(f_{t-1},f_t)$ (or $(K+1)^t$ if a step may keep the incumbent); certifying each at level $\alpha_t/K^{t}$ and taking a union bound over them restores validity. On the complement of the union of failure events, $\ell_t\le U^{(t)}$ for all $t$, whether $U^{(t)}$ certifies $\ell_t$ directly on the unconditional stream (as in our experiments) or certifies $\lambda_t\ge\ell_t$ on the out-of-scope substream. \qed

\paragraph{Natural-language envelopes.} If the intended scope $E^\star$ is stated in words and a committed judge realizes $E$, then $\ell(E^\star)\le\ell(E)+P(c_E=1,\,x\notin E^\star)$. (The false-inclusion mass $P(c_E=1,x\notin E^\star)$ needs scope labels, not outcome labels, and can be bounded once per scope, e.g.\ by prediction-powered inference \citep{angelopoulos2023ppi}, splitting $\alpha$ between the two bounds; a judge that makes $E$ too narrow only makes the certificate conservative.)
A change outside $E^\star$ either lies outside $E$, contributing to $\ell(E)$, or inside $E\setminus E^\star$, whose mass is $P(c_E=1,x\notin E^\star)$. \qed

\paragraph{Remark~\ref{prop:anchor}.}
The decision is the argmax of the option-token probabilities $q_f$, because renormalizing over the options does not change the argmax. If $f_1$ switches from option $i$ to option $j$, then $q_{1,j}\ge q_{1,i}$ and $q_{0,i}-q_{0,j}\le |q_{1,i}-q_{0,i}|+|q_{1,j}-q_{0,j}|\le 2\,\mathrm{TV}(p_{f_0},p_{f_1})\le\sqrt{2K(x)}$ by Pinsker's inequality. Also $g_0(x)\le q_{0,i}-q_{0,j}$, because $j$ is at most the runner-up. A change therefore requires $g_0\le\gamma$ or $K\ge\gamma^2/2$. The first event has the stated conditional probability, and Markov's inequality bounds the second by $2\,\mathbb{E}[K\mid x\notin E]/\gamma^2$. \qed

\paragraph{Lemma~\ref{lem:pinsker}.} Let $i=d_{f_0}(x)$ and $j=d_{f_1}(x)\ne i$, both options, so $p_0(i)-p_0(j)\ge g_0$ and $p_1(j)\ge p_1(i)$. Then $(p_0(i)-p_1(i))+(p_1(j)-p_0(j))\ge p_0(i)-p_0(j)\ge g_0$, so $\mathrm{TV}(p_0,p_1)=\tfrac12\sum_y|p_0(y)-p_1(y)|\ge\tfrac12(|p_0(i)-p_1(i)|+|p_0(j)-p_1(j)|)\ge g_0/2$. Pinsker's inequality $\mathrm{TV}\le\sqrt{\mathrm{KL}/2}$ gives $\mathrm{KL}(p_0\|p_1)\ge g_0^2/2$. The argument uses only the two options involved, so it applies to the full next-token distribution with unnormalized option probabilities and to the renormalized option distribution with its own gap. The leakage bound follows because $\{\Delta=1\}\subseteq\{\mathrm{KL}\ge g_0^2/2\}$.

\paragraph{Proposition~\ref{prop:lower}.}
Let $\sigma$ be the (possibly infinite) time at which the procedure stops, certifying or not, and let $A$ be the event that it certifies. If $\mathbb{E}_{\lambda_0}[\sigma]=\infty$ there is nothing to prove. Otherwise fix $n$ and apply the argument below to the bounded stopping time $\sigma\wedge n$ and the event $A_n=A\cap\{\sigma\le n\}$, which is measurable at time $\sigma\wedge n$; a bounded stopping time is needed because under $\lambda_1>\varepsilon$ a valid procedure may never stop. We use the transportation lemma of \citet{kaufmann2016complexity} (Lemma 1), a consequence of the change-of-measure argument behind Wald's sequential analysis \citep{wald1945sequential}: for i.i.d.\ Bernoulli observations, any almost surely finite stopping time $\sigma$ with $\mathbb{E}_{\lambda_0}[\sigma]<\infty$, and any event $A$ measurable at time $\sigma$, $\mathbb{E}_{\lambda_0}[\sigma]\,\mathrm{kl}(\lambda_0\|\lambda_1)\ge \mathrm{kl}\big(P_{\lambda_0}(A)\,\|\,P_{\lambda_1}(A)\big)$, where on the right $\mathrm{kl}$ is the binary relative entropy. Take $\lambda_1=\varepsilon+\delta$. Validity gives $P_{\lambda_1}(A_n)\le P_{\lambda_1}(A)\le\alpha$, and $P_{\lambda_0}(A_n)\to P_{\lambda_0}(A)\ge1-\beta$ as $n\to\infty$. Binary relative entropy $\mathrm{kl}(a\|b)$ is nondecreasing in $a$ and nonincreasing in $b$ for $a\ge b$, so $\mathbb{E}_{\lambda_0}[\sigma]\,\mathrm{kl}(\lambda_0\|\lambda_1)\ge\mathbb{E}_{\lambda_0}[\sigma\wedge n]\,\mathrm{kl}(\lambda_0\|\lambda_1)\ge\mathrm{kl}\big(P_{\lambda_0}(A_n)\|\alpha\big)$ once $P_{\lambda_0}(A_n)\ge\alpha$. Letting $n\to\infty$, then $\delta\to0$, and using continuity of $\mathrm{kl}$ proves the claim. \qed

\paragraph{Proposition~\ref{prop:onesided}.}
Given the past, the step-$s$ factor of $R_t(q)$ has conditional expectation $\lambda q/\varepsilon+(1-\lambda)(1-q)/(1-\varepsilon)$. This equals $1$ at $\lambda=\varepsilon$ and has derivative $q/\varepsilon-(1-q)/(1-\varepsilon)<0$ in $\lambda$ when $q<\varepsilon$, so it is at most $1$ for every $\lambda\ge\varepsilon$. Each $R_t(q)$ is therefore a nonnegative supermartingale with $R_0(q)=1$, and so is the mixture $R_t^W$ (by Tonelli's theorem, since $W$ is fixed in advance). Ville's inequality for nonnegative supermartingales gives $P(\exists t:R_t^W\ge1/\alpha)\le\alpha$. If $S_t=0$, then $R_t^W\ge W(\{0\})\,(1-\varepsilon)^{-t}$, which reaches $1/\alpha$ at the stated $t$. With $\lambda_0=0$ and $\beta\to0$, Proposition~\ref{prop:lower} requires $\log(1/\alpha)/(-\log(1-\varepsilon))$ samples, so the excess is at most $\log(1/W(\{0\}))/(-\log(1-\varepsilon))$. \qed

\paragraph{Choosing the envelope size.}
The $G$ envelopes are fixed before auditing because the score and the grid are committed. The event that any of the $G$ certificates is wrong has probability at most $G\cdot\alpha/G=\alpha$ by the union bound, and on its complement every reported certificate, including the smallest certified one chosen after seeing the data, is correct. \qed

\paragraph{Traffic shift.}
$\ell_P(E)=\int\Delta(x)\mathbb{1}[x\notin E]\,dP=\int\Delta(x)\mathbb{1}[x\notin E]\,\tfrac{dP}{dQ}\,dQ\le w\,\ell_Q(E)$, and dividing by $P(x\notin E)$ gives $\lambda_P(E)\le w\,\lambda_Q(E)\,Q(x\notin E)/P(x\notin E)$. \qed

\section{Implementation details}
\label{app:impl}\label{app:more}

The code release at \url{https://github.com/Usama1002/declared-change-envelopes} contains scripts that rebuild the traffic pool and training sets, run every update and audit reported here with the hyperparameters below, and regenerate all tables and figures from the saved result files.

\subsection{Decisions, training, and audits}

\paragraph{Prompt and decision.}
Each query is a single user turn: the question (truncated to its last 1{,}800 characters), the options labeled \texttt{A.}, \texttt{B.}, \dots, and the instruction ``Answer with the letter of the correct option only.'' We apply the model's chat template with the generation prompt and append the fixed assistant prefix ``\texttt{The correct option is}''. The decision is the argmax of the next-token log-probabilities restricted to the tokens `` A'', `` B'', \dots\ of the valid options, and the margin $\mu_f(x)$ is the gap between the two largest restricted log-probabilities. All models run in bfloat16 with PyTorch scaled-dot-product attention. We disable the cuDNN attention backend: in our stack (PyTorch 2.10 NGC build, Transformers 4.57) it returned NaN gradients on padded batches, and bfloat16 eager attention for Qwen2.5 disagreed with a float32 reference by more than 2 nats of answer-token loss, whereas flash and math SDPA matched it to within 0.15 nats.

\paragraph{Training.}
LoRA adapters (rank 16, scaling factor 32, no dropout) on all attention and MLP projections are trained with AdamW (learning rate $10^{-4}$, no weight decay, 5\% linear warmup, linear decay, gradient clipping at 1.0), batch size 16, right padding, gradient checkpointing, and cross-entropy on the full-vocabulary next-token distribution at the answer position. \emph{med}, \emph{legal}, and \emph{policy} train for two epochs over 2{,}000 examples (250 steps), \emph{edit} for five epochs over 400 edits (125 steps). Anchored updates add $\beta\,\mathrm{KL}(p_{f_0}\|p_{f_\theta})$ on a batch of 16 anchor-set queries outside the taxonomy envelope per step, with $p_{f_0}$ computed with the adapter disabled. Pilot updates use seed 2 and 25\% of the steps. Sequential chains merge each adapter into the weights before training the next, and the anchored chain anchors each step to its immediate incumbent.

\paragraph{Envelope commitments.}
Before training, each run logs a SHA-256 hash of its envelope specification (update type, envelope rule, natural-language scope, judge identity). The NL-judge scores are computed once, for all scopes, from Qwen2.5-7B-Instruct with the prompt ``Scope: $\langle$scope$\rangle$. Does the following input fall within this scope? Answer Yes or No.'' followed by the input and its options; the score is $P(\text{Yes})/(P(\text{Yes})+P(\text{No}))$.

\paragraph{Confirmatory audit.} The audit split has 23{,}896 queries: the development audit set (11{,}945) and the confirmatory set (11{,}951), on which no model was evaluated during development. Plain adapters are re-evaluated on it; anchored variants are retrained with seeds 1 and 2 and evaluated on the full pool; certificates use one pre-registered random order at level $\alpha/4$ (Table~\ref{tab:confirm}).

\paragraph{Checkpoint provenance.} The anchored Qwen medical seed-0 model used in the effect, free-form, refusal, and generation checks (and as the first step of the anchored chain) is an independent training run of the $\beta=1$ configuration, distinct from the \emph{med\_kl1} run in Table~\ref{tab:variants}; its metadata file records only a later re-evaluation of the saved adapter. Likewise, the Llama ``anchored seed 0'' model in the refusal slice is the first step of the anchored Llama chain, a separate run of the same configuration.

\subsection{Other update types and models}

\paragraph{DPO.} Decision-token DPO \citep{rafailov2023dpo} uses the same 2{,}000 training prompts, the target option as the preferred and the incumbent's highest-probability wrong option as the rejected response, the incumbent (adapter disabled) as reference, and $\beta_{\text{DPO}}=0.1$; other settings match the supervised updates.

\paragraph{Full fine-tuning.} All weights are trained with the decision-token cross-entropy of the LoRA updates (batch 16, same epochs, 5\% linear warmup and linear decay), fp32 master weights with bf16 autocast, 8-bit AdamW \citep{dettmers2022optimizers} without weight decay, gradient clipping at 1, and gradient checkpointing; the anchor KL uses a frozen bf16 copy of the incumbent, and trained weights are cast to bf16 and evaluated like every other update. Plain runs use learning rates $10^{-6}$, $3\cdot10^{-6}$, and $10^{-5}$. Anchored runs use $\beta\in\{1,10\}$ at $10^{-5}$ and $3\cdot10^{-6}$, $\beta=10$ with 64 or 128 anchor prompts per step at $3\cdot10^{-6}$, and $\beta=10$ with 64 anchors at $10^{-6}$. At $3\cdot10^{-6}$ (Table~\ref{tab:updtypes-full}), anchored leakage is 4.4--9.3\% with 64 anchors per step (second seed 4.5--9.8\%), and 128 anchors help little (4.1--4.3\% on Qwen, 7.5--9.7\% on Llama). The out-of-scope letter KL on the audit split stays at 0.05--0.08 nats at $3\cdot10^{-6}$, against 0.007--0.013 at $10^{-6}$ and 0.003--0.028 for anchored LoRA; for Llama with 128 anchors it is only half as large on the anchor split, so the update fits the anchors it sees without carrying the constraint to other traffic. Training takes 2.3--2.7 minutes at 86--92\,GB peak memory for plain updates, 6.4--7.1 minutes at 105--112\,GB with 16 anchors per step, and 21--23 minutes at 122--131\,GB with 64 anchors, on one H200.

\paragraph{Decision-token GRPO.} Each step takes 16 training prompts, samples eight option letters per prompt from the policy renormalized over the valid options, and rewards the target option (the correct answer for medical, the relabeled topic for policy), with advantages normalized within each group. The update is on-policy with one gradient step per batch, so the importance ratio is 1 and no clipping is needed. The KL term is the exact KL of the full next-token distribution at the decision position between policy and incumbent on the training prompts, with weight 0.04 or 1. LoRA, learning rate, and number of steps are those of the SFT updates; anchoring adds the out-of-scope decision-position KL of Eq.~(\ref{eq:anchor}) with $\beta=1$.

\paragraph{RLVR on GSM8K.} GRPO with LoRA (rank 16) on GSM8K training problems in the free-form template of \S\ref{sec:res-beyond}: 16 problems per step, eight samples each (temperature 1, at most 320 new tokens), reward 1 when the number after the last ``The answer is'' equals the gold answer, group-normalized advantages, a per-token k3 estimator of KL to the incumbent with weight 0.04, and loss averaged per sequence. On the 7B and 8B models the default run uses 150 steps at learning rate $2\cdot10^{-5}$ and the strong run 300 steps at $5\cdot10^{-5}$; all three Qwen2.5-1.5B runs use 300 steps at $3\cdot10^{-5}$. The declared scope is the MMLU subjects elementary, high-school, and college mathematics, abstract algebra, and high-school statistics (1{,}064 queries, 3.1\% of the pool). Anchoring adds the decision-position KL on 16 out-of-scope anchor prompts per step. The in-scope effect is GSM8K accuracy on the 800 problems of the free-form slice; out-of-scope free-form change is measured on the 2{,}000 NQ-open questions with the semantic judge.

\paragraph{Other models.} Qwen2.5-32B-Instruct and Mistral-7B-Instruct-v0.3 use the same prompts, decision rule, and LoRA settings, with batch size 16 at evaluation; 32B updates have two seeds and Mistral updates one. Qwen2.5-72B is split layer-wise over two H200 GPUs with the LoRA settings of the 7B updates. Base and self-disagreement evaluations use the full pool, and updates the evaluation subset.

\paragraph{Mistral diagnosis.} Anchored Mistral runs use $\beta=10$, the KL restricted to option letters, four times more anchor queries per step (64), or learning rate $3\times10^{-5}$ for both plain and anchored training (Table~\ref{tab:variants-other}); $\beta=10$ cuts the medical gain from $+5.3$ to $+2.1$. With $\beta=10$, letter-only KL, or more anchors, Mistral leaks 7.6--11.6\%. At learning rate $3\cdot10^{-5}$ anchored Mistral reaches 6.6\% (medical, effect $+3.8$ against $+5.3$ for default LoRA) and 4.2\% (policy, $+82.3$ against $+82.7$); the policy run certifies $\varepsilon=5\%$ on the fixed audit order but in only 1 of 20 random orders, and it is a diagnostic, not a pre-declared candidate.

\subsection{Alternatives to anchoring, effect estimates, and knowledge editors}

\paragraph{Alternatives to anchoring.} All use the medical and policy updates on Qwen and Llama (Table~\ref{tab:controls}, Figure~\ref{fig:pareto}). \emph{Learning-rate sweep}: plain LoRA with learning rates $\{10^{-5},3\times10^{-5},10^{-4}\}$ and one, two, or four epochs (Qwen: Table~\ref{tab:sweep}). \emph{Adapter scaling}: the default adapter with its LoRA scaling multiplied by 0.25, 0.5, or 0.75 before merging, a linear interpolation of the weight update toward the incumbent. \emph{L2 to the incumbent}: $\|\tfrac{\alpha}{r}BA\|_F^2$ summed over LoRA weight updates, with weight $\mu\in\{0.1,1\}$. \emph{Incumbent replay}: 1{,}000 anchor-set queries labeled with the incumbent's own decisions added to the training set. \emph{KL on generic text}: token-level KL to the incumbent averaged over the incumbent's greedy 64-token responses to 3{,}000 Alpaca instructions \citep{taori2023alpaca}, with the $\beta$ and batch size of envelope anchoring. \emph{KL on all traffic}: anchoring without restricting anchors to the complement of the envelope ($\beta\in\{1,10\}$; at $\beta=10$ the medical gains are 0.2--0.7 points lower than with out-of-scope anchoring, not significant). \emph{KL on generic multiple choice}: decision-position anchoring on 1{,}500 OpenBookQA \citep{mihaylov2018openbookqa} and 1{,}500 HellaSwag \citep{zellers2019hellaswag} questions in the pool's prompt format; neither dataset is in the pool. \emph{Judge envelope}: the NL-judge envelope with mass matched to the taxonomy on the anchor set; anchors are the anchor-set queries outside it, and leakage is measured outside it.

\paragraph{Effect with adequate power.} Plain and anchored medical adapters are evaluated on all 6{,}277 medical queries of the pool (both halves) and on the 1{,}273 MedQA-USMLE test questions \citep{jin2021medqa}, on which no model was trained or tuned; intervals are paired bootstrap intervals over questions (Table~\ref{tab:effect}).

\paragraph{Knowledge editors.} MEMIT \citep{meng2023memit} and AlphaEdit \citep{fang2025alphaedit} are re-implemented for Qwen2.5 and Llama-3.1 and insert the 400 CounterFact rewrites in one batch into the MLP down-projections of layers 4--8, with second-moment statistics of the down-projection inputs from one million WikiText-103 tokens \citep{merity2017wikitext}. Target residuals at layer 8 come from 25 Adam steps (learning rate 0.1, norm clamp 4, KL factor 0.0625, weight decay $10^{-3}$) on the rewrite prompt under six context prefixes; MEMIT spreads them with covariance weight $1.5\times10^4$ (for Llama also $6\times10^4$), and AlphaEdit projects onto the covariance eigenvectors with eigenvalue below $10^{-2}$ with ridge weight 10. On the standard CounterFact metrics (raw-text completion, 400 edits) the edited models reach efficacy 99.8--100\%, paraphrase 87--99\%, and neighborhood 70--84\% (85\% before editing), in line with published results; on the rewrite prompts in our chat multiple-choice format they choose the new object in 77--90\% of cases.

\subsection{Outputs beyond letter decisions}

\paragraph{Generation check.} 2{,}000 out-of-scope audit queries are answered with the plain chat template (no assistant prefix), greedy decoding, and 24 new tokens; the option letter is parsed with fixed patterns (``answer is X'', a leading letter, or a standalone letter), which succeed on 81--100\% of responses. Among queries parsed for both models, the fraction of letter flips that also change the written answer is 0.69 and 0.80 (Qwen medical and policy, plain), 0.75 and 0.80 (Llama, plain), and 0.31, 0.00, 0.44, and 0.14 for the anchored models (Qwen: the seed-0 medical and seed-1 policy models; Llama: seed-1 models); generation-measured leakage is, for example, 23.3\% (plain) and 1.3\% (anchored) for Qwen policy. Anchored Llama medical (seed 1) leaks 6.1\% under generation, above its 4.9\% letter leakage on the same queries. Re-generating the incumbent with batch size 16 changes 0.1--0.7\% of parsed answers, so generation-measured leakage of anchored models (1.3--6.1\%) is above the generation floor.

\paragraph{Free-form slice.} GSM8K: 800 test problems (fixed random subset), instructed to solve step by step and finish with ``The answer is N.'', greedy decoding with 512 new tokens; the decision is the number after the last ``answer is'', or \emph{incomplete} if there is none (under 3\% of responses for the 7B and 8B models; 13.9\% for the Qwen2.5-1.5B incumbent and 0.9--1.3\% after RLVR, so part of the 1.5B GSM8K gain comes from fewer incomplete answers). NQ-open: 2{,}000 validation questions answered with a short phrase, greedy decoding with 24 new tokens; answers are normalized (lowercase, punctuation and articles removed), and every pair whose normalized strings differ is judged by Qwen2.5-32B-Instruct (``Do Answer 1 and Answer 2 give the same answer to the question? Yes or No''). Anchored models are the seed-0 (Qwen medical) and seed-1 checkpoints (Table~\ref{tab:freeform}). \emph{Free-form anchoring} adds, to the decision-position KL, token-level KL to the incumbent over its greedy responses (up to 160 tokens) on 500 GSM8K training and 1{,}000 NQ-open training prompts in the templates of the free-form slice, which uses only GSM8K test and NQ-open validation questions; $\beta=1$ for both terms.

\paragraph{Refusal slice.} 720 unsafe prompts (the 200 unsafe XSTest prompts \citep{rottger2024xstest} and 520 AdvBench behaviors \citep{zou2023universal}) are answered greedily with 128 new tokens and no system prompt, and Qwen2.5-32B-Instruct classifies each response with the XSTest three-way rubric; a full or partial refusal counts as a refusal. The floor re-generates the incumbent with a different batch size. After the medical update, plain LoRA changes 1.4--2.1\% of refusal decisions for Qwen (floor 1.0\%) and 1.9\% for Llama (floor 0.3\%), in both directions, and anchored models 1.2--1.8\% and 1.5--1.9\%; no anchored--plain difference is significant (exact McNemar $p\ge0.14$; Table~\ref{tab:safety}).

\paragraph{Open-ended chat audit.} Prompts are first user turns of English OpenAssistant conversations (oasst2 train and validation, 20--1{,}500 characters), sampled with a fixed seed and answered greedily with at most 256 new tokens in batches of 32 ordered by prompt length; every adapter runs with exactly the batches of the incumbent, and the floor re-runs the incumbent with batches of 16. The tiny random adapter has LoRA $B$ matrices drawn with standard deviation $10^{-3}$. Qwen2.5-32B-Instruct answers two yes/no questions by comparing next-token scores: whether a prompt falls in an update's natural-language scope (\S\ref{sec:setup}), and, for every pair whose response text differs, whether the two responses ``give substantively the same content: the same answer, facts, recommendations, and conclusions, ignoring wording, formatting, length, and minor details''.

\section{Validity of monitored certificates: simulation}
\label{app:sim}
We simulate audit streams of i.i.d.\ Bernoulli$(\lambda)$ leak indicators of length 20{,}000, check every 5 samples, and stop at the first check where the procedure certifies $\varepsilon=0.02$ ($\alpha=0.05$, 2{,}000 replications per $\lambda$). Table~\ref{tab:sim} reports the certification rate and the median number of samples at certification. Rows with $\lambda\ge\varepsilon$ measure false certification. Both anytime-valid procedures stay below $\alpha$; the one-sided test needs 2.2 to 2.6 times fewer samples than the mixture sequence for $\lambda\le\varepsilon/2$, at the price of certifying only the declared $\varepsilon$.

\begin{table}[H]
\centering
\caption{Monitored certification at $\varepsilon=0.02$, $\alpha=0.05$ (2{,}000 replications, checked every 5 samples up to 20{,}000). Mixture: Beta-mixture confidence sequence (Theorem~\ref{thm:cs}). One-sided: mixture-of-SPRTs test at the declared $\varepsilon$ (Proposition~\ref{prop:onesided}). CP-monitored: one-sided Clopper--Pearson bound recomputed at every check. CP-fixed: Clopper--Pearson evaluated once at $n=20{,}000$. Median $n$ is over replications that certify. Rows with $\lambda\ge\varepsilon$ give false-certification rates.}
\label{tab:sim}
\small
\setlength{\tabcolsep}{4pt}
\begin{tabular}{lccccccc}
\toprule
 & \multicolumn{2}{c}{Mixture} & \multicolumn{2}{c}{One-sided} & \multicolumn{2}{c}{CP-monitored} & CP-fixed \\
\cmidrule(lr){2-3}\cmidrule(lr){4-5}\cmidrule(lr){6-7}
true $\lambda$ & cert.\ rate & median $n$ & cert.\ rate & median $n$ & cert.\ rate & median $n$ & cert.\ rate \\
\midrule
0.000 & 1.000 & 455 & 1.000 & 175 & 1.000 & 150 & 1.000 \\
0.005 & 1.000 & 910 & 1.000 & 385 & 1.000 & 240 & 1.000 \\
0.010 & 1.000 & 2{,}340 & 1.000 & 1{,}070 & 1.000 & 390 & 1.000 \\
0.015 & 0.908 & 10{,}350 & 0.991 & 5{,}310 & 1.000 & 1{,}030 & 1.000 \\
\midrule
0.020 (false) & \textbf{0.0005} & -- & \textbf{0.033} & -- & 0.332 & 1{,}275 & 0.044 \\
0.025 (false) & \textbf{0} & -- & \textbf{0.013} & -- & 0.059 & 240 & 0.000 \\
\bottomrule
\end{tabular}
\end{table}

\section{Additional results}
\label{app:extra}

\begin{figure}[tbp]
\centering
\includegraphics[width=\linewidth]{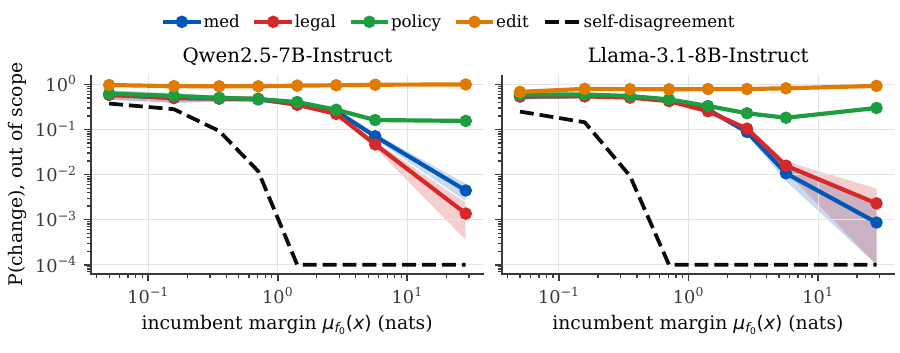}
\caption{Probability that an out-of-scope decision changes versus the incumbent's margin. Dashed: self-disagreement. Domain updates move mostly fragile decisions; rule-like updates also overwrite confident ones.}
\label{fig:margin}
\par\vspace{0.6\baselineskip}
\centering
\includegraphics[width=\linewidth]{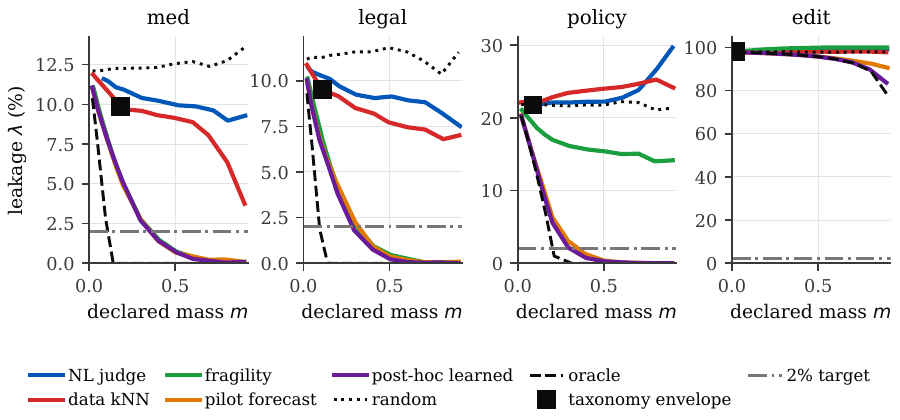}
\caption{Footprint curves on Qwen: leakage of the top-$m$ envelope under each declaration score. Square: the taxonomy envelope. Dash-dotted line: 2\% leakage. Pilot forecasts track the post-hoc reference; scope-based declarations (NL judge, data kNN) are close to random.}
\label{fig:curves}
\end{figure}

\begin{figure}[tbp]
\centering
\includegraphics[width=\linewidth]{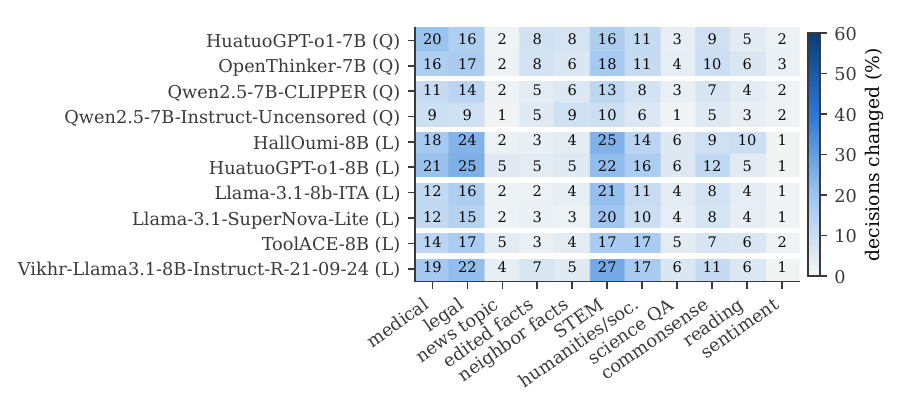}
\caption{Percentage of decisions changed per traffic domain by public fine-tunes of Qwen2.5-7B-Instruct (Q) and Llama-3.1-8B-Instruct (L).}
\label{fig:heatmap-public}
\end{figure}

This appendix collects the tables and figures behind the main-text results. For \S\ref{sec:res-leak} and \S\ref{sec:res-structure}: the per-domain footprints of public fine-tunes (Figure~\ref{fig:heatmap-public}, Table~\ref{tab:public}), the margin dependence of leakage (Figure~\ref{fig:margin}), and the declaration scores (Table~\ref{tab:decl}, Figure~\ref{fig:curves}). For \S\ref{sec:res-remedies}: every alternative to anchoring (Table~\ref{tab:controls}) and the learning-rate sweep (Table~\ref{tab:sweep}). For \S\ref{sec:res-anchor}: the medical effect with adequate power (Table~\ref{tab:effect}) and the confirmatory audit of the LoRA candidates (Table~\ref{tab:confirm}). For \S\ref{sec:res-types}--\S\ref{sec:res-edit}: all full fine-tuning and reinforcement-learning variants (Table~\ref{tab:updtypes-full}), scale and model family (Table~\ref{tab:scale}), and the edit update (Table~\ref{tab:edit}). For \S\ref{sec:res-beyond}: free-form answers (Table~\ref{tab:freeform}) and refusals (Table~\ref{tab:safety}). Tables~\ref{tab:variants}, \ref{tab:variants-llama}, and~\ref{tab:variants-other} list every trained update variant with its effect, leakage, decisive leakage, final upper confidence bound, and certification over 20 random audit orders.

\begin{table}[tbp]
\centering
\caption{Declaring the footprint before training: AUROC of each score for predicting which out-of-scope audit decisions change. The post-hoc row is a reference fit to the real update's anchor-set flips and is not available before training. Leakage-versus-mass curves: Figure~\ref{fig:curves}.}
\label{tab:decl}
\small
\setlength{\tabcolsep}{4pt}
\begin{tabular}{lrrrrrrrr}
\toprule
 & \multicolumn{4}{c}{Qwen2.5-7B} & \multicolumn{4}{c}{Llama-3.1-8B} \\
\cmidrule(lr){2-5}\cmidrule(lr){6-9}
Declaration & med & legal & policy & edit & med & legal & policy & edit \\
\midrule
NL judge & 0.50 & 0.52 & 0.47 & 0.47 & 0.54 & 0.54 & 0.47 & 0.43 \\
data kNN & 0.56 & 0.57 & 0.45 & 0.47 & 0.48 & 0.53 & 0.42 & 0.58 \\
fragility & 0.89 & 0.91 & 0.64 & 0.16 & 0.88 & 0.86 & 0.64 & 0.43 \\
fragility + judge & 0.80 & 0.82 & 0.59 & 0.24 & 0.80 & 0.79 & 0.59 & 0.39 \\
pilot forecast & 0.89 & 0.90 & 0.96 & 0.85 & 0.88 & 0.87 & 0.96 & 0.55 \\
post-hoc (reference) & 0.89 & 0.91 & 0.97 & 0.95 & 0.89 & 0.88 & 0.98 & 0.87 \\
\bottomrule
\end{tabular}

\par\vspace{0.6\baselineskip}
\centering
\caption{Public fine-tunes of Qwen2.5-7B-Instruct (Q) and Llama-3.1-8B-Instruct (L), evaluated on the audit traffic (percent). None of them declares general multiple-choice question answering as its purpose. Reasoning models (HuatuoGPT-o1, OpenThinker) are evaluated through the same forced-answer interface, without their chain of thought, so their numbers describe decision drift under a fixed interface rather than end-to-end quality.}
\label{tab:public}
\footnotesize
\setlength{\tabcolsep}{2pt}
\begin{tabular}{llrrrr}
\toprule
Fine-tune (base) & Declared purpose & changed & changed$_\tau$ & \hd{net\\acc.} & \hd{neg.\ / pos.\\flips} \\
\midrule
HuatuoGPT-o1-7B (Q) & medical reasoning & 11.0 & 9.6 & +1.3 & 3.6 / 5.0 \\
OpenThinker-7B (Q) & math/code/science reasoning & 10.6 & 9.3 & $-$1.2 & 4.6 / 3.4 \\
Qwen2.5-7B-CLIPPER (Q) & claim verification in books & 7.8 & 6.3 & $-$0.3 & 3.1 / 2.8 \\
Qwen2.5-7B-Instruct-Uncensored (Q) & removing refusals & 6.1 & 4.6 & +0.3 & 2.1 / 2.4 \\
HallOumi-8B (L) & hallucination detection & 13.3 & 10.0 & $-$1.1 & 5.6 / 4.5 \\
HuatuoGPT-o1-8B (L) & medical reasoning & 14.2 & 11.2 & +1.7 & 4.7 / 6.4 \\
Llama-3.1-8b-ITA (L) & Italian language & 9.6 & 6.4 & +1.1 & 3.0 / 4.2 \\
Llama-3.1-SuperNova-Lite (L) & general distillation & 8.8 & 5.7 & +1.4 & 2.6 / 4.0 \\
ToolACE-8B (L) & function calling & 11.5 & 8.5 & +0.4 & 4.1 / 4.5 \\
\hd[l]{Vikhr-Llama3.1-8B-Instruct-\\R-21-09-24 (L)} & Russian language & 14.1 & 11.0 & +0.3 & 5.1 / 5.4 \\
\bottomrule
\end{tabular}

\end{table}

\begin{table}[tbp]
\centering
\caption{Every alternative to anchoring on declared out-of-scope traffic (percent, development audit set). $^\ast$Leakage measured outside the judge envelope.}
\label{tab:controls}
\scriptsize
\setlength{\tabcolsep}{4pt}
\begin{tabular}{llrr}%
\toprule
Update & Method & effect & $\lambda$ \\
\midrule
\multicolumn{4}{l}{\emph{Qwen2.5-7B}} \\
med & plain LoRA (default) & +2.2 & 9.9 \\
& plain LoRA, lowest-leakage of 9 lr$\times$epoch configs (lr 1e-5, 1 ep.) & +0.6 & 4.2 \\
& adapter scaled by 0.25 & +2.5 & 4.8 \\
& adapter scaled by 0.5 & +3.2 & 7.0 \\
& L2 penalty to incumbent ($\mu{=}1$) & +3.5 & 6.7 \\
& incumbent replay (1{,}000) & +2.3 & 6.5 \\
& KL on generic chat text ($\beta{=}1$) & +1.9 & 8.4 \\
& KL on generic multiple choice ($\beta{=}1$) & +2.9 & 3.8 \\
& KL on all traffic incl.\ envelope ($\beta{=}1$) & +2.7 & 3.4 \\
& KL on all traffic incl.\ envelope ($\beta{=}10$) & +1.7 & 1.5 \\
& anchoring on out-of-scope traffic ($\beta{=}1$) & +2.2 & 3.3 \\
& anchoring on out-of-scope traffic ($\beta{=}10$) & +2.0 & 1.4 \\
& anchoring on the NL-judge envelope ($\beta{=}1$)$^\ast$ & +2.9 & 3.9 \\
\midrule
policy & plain LoRA (default) & +82.7 & 21.8 \\
& plain LoRA, lowest-leakage of 9 lr$\times$epoch configs (lr 1e-5, 2 ep.) & +81.5 & 12.1 \\
& adapter scaled by 0.25 & +35.8 & 4.4 \\
& adapter scaled by 0.5 & +79.9 & 8.1 \\
& L2 penalty to incumbent ($\mu{=}1$) & +81.5 & 10.8 \\
& incumbent replay (1{,}000) & +83.9 & 6.6 \\
& KL on generic chat text ($\beta{=}1$) & +83.9 & 19.8 \\
& KL on generic multiple choice ($\beta{=}1$) & +83.5 & 1.8 \\
& KL on all traffic incl.\ envelope ($\beta{=}1$) & +82.7 & 1.6 \\
& KL on all traffic incl.\ envelope ($\beta{=}10$) & +81.1 & 1.3 \\
& anchoring on out-of-scope traffic ($\beta{=}1$) & +84.3 & 1.6 \\
& anchoring on out-of-scope traffic ($\beta{=}10$) & +82.3 & 1.1 \\
& anchoring on the NL-judge envelope ($\beta{=}1$)$^\ast$ & +83.5 & 2.2 \\
\midrule
\multicolumn{4}{l}{\emph{Llama-3.1-8B}} \\
med & plain LoRA (default) & +1.3 & 15.9 \\
& plain LoRA, lowest-leakage of 9 lr$\times$epoch configs (lr 1e-5, 1 ep.) & +1.1 & 7.3 \\
& adapter scaled by 0.25 & +1.2 & 6.6 \\
& adapter scaled by 0.5 & +1.5 & 10.9 \\
& L2 penalty to incumbent ($\mu{=}1$) & +1.5 & 13.5 \\
& KL on generic chat text ($\beta{=}1$) & +2.3 & 13.4 \\
& KL on generic multiple choice ($\beta{=}1$) & +2.1 & 6.7 \\
& KL on all traffic incl.\ envelope ($\beta{=}1$) & +2.1 & 5.1 \\
& KL on all traffic incl.\ envelope ($\beta{=}10$) & +1.3 & 2.1 \\
& anchoring on out-of-scope traffic ($\beta{=}1$) & +2.5 & 5.2 \\
& anchoring on out-of-scope traffic ($\beta{=}10$) & +2.0 & 2.5 \\
& anchoring on the NL-judge envelope ($\beta{=}1$)$^\ast$ & +2.6 & 6.1 \\
\midrule
policy & plain LoRA (default) & +83.1 & 30.5 \\
& plain LoRA, lowest-leakage of 9 lr$\times$epoch configs (lr 3e-5, 2 ep.) & +83.1 & 23.5 \\
& adapter scaled by 0.25 & +45.3 & 7.5 \\
& adapter scaled by 0.5 & +78.0 & 14.9 \\
& L2 penalty to incumbent ($\mu{=}1$) & +82.3 & 20.4 \\
& KL on generic chat text ($\beta{=}1$) & +84.6 & 30.0 \\
& KL on generic multiple choice ($\beta{=}1$) & +84.6 & 2.7 \\
& KL on all traffic incl.\ envelope ($\beta{=}1$) & +82.3 & 2.5 \\
& KL on all traffic incl.\ envelope ($\beta{=}10$) & +82.7 & 2.0 \\
& anchoring on out-of-scope traffic ($\beta{=}1$) & +83.9 & 2.7 \\
& anchoring on out-of-scope traffic ($\beta{=}10$) & +83.5 & 2.0 \\
& anchoring on the NL-judge envelope ($\beta{=}1$)$^\ast$ & +83.1 & 3.1 \\
\bottomrule
\end{tabular}

\end{table}

\begin{table}[tbp]
\centering
\caption{Plain-LoRA learning-rate $\times$ epoch sweep, incumbent replay, and anchoring on Qwen2.5-7B (percent, development audit set).}
\label{tab:sweep}
\small
\setlength{\tabcolsep}{4pt}
\begin{tabular}{lrrrr}%
\toprule
 & \multicolumn{2}{c}{med} & \multicolumn{2}{c}{policy} \\
\cmidrule(lr){2-3}\cmidrule(lr){4-5}
Configuration & effect & $\lambda$ & effect & $\lambda$ \\
\midrule
LoRA default (lr 1e-4, 2 ep.) & +2.2 & 9.9 & +82.7 & 21.8 \\
LoRA lr 1e-4, 1 ep. & +3.5 & 8.3 & +83.9 & 19.3 \\
LoRA lr 1e-4, 4 ep. & +1.8 & 10.1 & +84.6 & 19.9 \\
LoRA lr 1e-5, 1 ep. & +0.6 & 4.2 & +81.5 & 13.0 \\
LoRA lr 1e-5, 2 ep. & +1.5 & 4.8 & +81.5 & 12.1 \\
LoRA lr 1e-5, 4 ep. & +2.0 & 5.1 & +82.7 & 14.1 \\
LoRA lr 3e-5, 1 ep. & +1.7 & 4.7 & +81.5 & 17.6 \\
LoRA lr 3e-5, 2 ep. & +3.2 & 6.3 & +82.3 & 13.6 \\
LoRA lr 3e-5, 4 ep. & +1.3 & 7.9 & +83.9 & 15.7 \\
anchored $\beta{=}1$ & +2.2 & 3.3 & +84.3 & 1.6 \\
anchored $\beta{=}10$ & +2.0 & 1.4 & +82.3 & 1.1 \\
incumbent replay & +2.3 & 6.5 & +83.9 & 6.6 \\
\bottomrule
\end{tabular}

\end{table}

\begin{table}[tbp]
\centering
\caption{Medical effect with adequate power (accuracy gain over the incumbent in points, with paired bootstrap 95\% intervals; last two columns: anchored minus the mean of the plain seeds).}
\label{tab:effect}
\scriptsize
\setlength{\tabcolsep}{4pt}
\begin{tabular}{lrrrr}%
\toprule
Variant & gain, pool medical & gain, MedQA & $-$ plain, pool medical & $-$ plain, MedQA \\
\midrule
\multicolumn{5}{l}{\emph{Qwen2.5-7B}} \\
plain s0 & +2.8 [+1.8, +3.7] & +3.8 [+1.6, +6.0] & -- & -- \\
plain s1 & +2.8 [+1.9, +3.8] & +3.1 [+1.2, +5.1] & -- & -- \\
kl10 s1 & +1.5 [+0.9, +2.2] & +1.4 [+0.3, +2.6] & $-$1.3 [$-$2.1, $-$0.5] & $-$2.0 [$-$3.7, $-$0.1] \\
kl10 s2 & +1.7 [+0.9, +2.4] & +2.3 [+1.0, +3.5] & $-$1.1 [$-$2.0, $-$0.3] & $-$1.2 [$-$2.9, +0.5] \\
kl1 s0 & +1.6 [+0.8, +2.4] & +2.4 [+0.6, +4.2] & $-$1.2 [$-$1.8, $-$0.5] & $-$1.0 [$-$2.5, +0.5] \\
kl1 s1 & +2.2 [+1.4, +3.0] & +1.2 [$-$0.5, +2.8] & $-$0.6 [$-$1.3, +0.1] & $-$2.3 [$-$3.8, $-$0.9] \\
kl1 s2 & +1.9 [+1.1, +2.8] & +2.6 [+0.9, +4.3] & $-$0.9 [$-$1.6, $-$0.2] & $-$0.9 [$-$2.3, +0.5] \\
\midrule
\multicolumn{5}{l}{\emph{Llama-3.1-8B}} \\
plain s0 & +0.9 [$-$0.1, +2.0] & +1.5 [$-$0.7, +3.7] & -- & -- \\
plain s1 & +1.0 [+0.0, +2.0] & +2.4 [+0.3, +4.6] & -- & -- \\
kl10 s1 & +1.2 [+0.4, +1.9] & +1.3 [$-$0.4, +3.0] & +0.2 [$-$0.6, +1.0] & $-$0.6 [$-$2.2, +1.0] \\
kl10 s2 & +1.2 [+0.5, +1.9] & +1.4 [$-$0.2, +3.1] & +0.3 [$-$0.5, +1.1] & $-$0.5 [$-$2.1, +1.1] \\
kl1 s1 & +1.4 [+0.5, +2.2] & +2.7 [+0.7, +4.7] & +0.4 [$-$0.2, +1.1] & +0.8 [$-$0.6, +2.2] \\
kl1 s2 & +1.5 [+0.6, +2.3] & +1.7 [$-$0.3, +3.6] & +0.5 [$-$0.2, +1.2] & $-$0.2 [$-$1.7, +1.3] \\
\bottomrule
\end{tabular}

\par\vspace{0.6\baselineskip}
\centering
\caption{Confirmatory audit and seeds (percent). Development leakage is the mean and standard deviation over seeds 0--2; held-out columns give seeds 1 and 2 (anchored) or seeds 0 and 1 (plain) on the 11{,}951 audit queries never evaluated during development, with the smallest certified $\varepsilon$ at level $\alpha/4$ in one pre-registered order.}
\label{tab:confirm}
\scriptsize
\setlength{\tabcolsep}{4pt}
\begin{tabular}{lrrrrl}%
\toprule
Variant & seeds & $\lambda$ dev (mean $\pm$ sd) & $\lambda$ held-out & effect held-out & certified held-out \\
\midrule
\multicolumn{6}{l}{\emph{Qwen2.5-7B}} \\
edit kl1 & 3 & 5.6 $\pm$ 0.24 & 5.4/5.5 & +69.2/+69.9 & no/no \\
legal kl1 & 3 & 2.2 $\pm$ 0.01 & 2.0/2.4 & +6.0/+5.4 & 5\%/5\% \\
med kl1 & 3 & 3.3 $\pm$ 0.35 & 3.0/3.6 & +1.3/+1.6 & 5\%/5\% \\
med kl10 & 3 & 1.5 $\pm$ 0.07 & 1.6/1.8 & +1.1/+0.8 & 5\%/5\% \\
policy kl1 & 3 & 1.6 $\pm$ 0.04 & 1.9/1.6 & +76.4/+76.4 & 5\%/5\% \\
policy kl10 & 3 & 1.2 $\pm$ 0.17 & 1.5/1.3 & +76.4/+76.7 & 5\%/2\% \\
edit s0 (plain) & 1 & -- & 98.2 & +71.6 & -- \\
edit s1 (plain) & 1 & -- & 97.0 & +74.2 & -- \\
legal s0 (plain) & 1 & -- & 9.4 & +5.0 & -- \\
legal s1 (plain) & 1 & -- & 9.0 & +6.2 & -- \\
med s0 (plain) & 1 & -- & 10.6 & +2.8 & -- \\
med s1 (plain) & 1 & -- & 8.3 & +2.2 & -- \\
policy s0 (plain) & 1 & -- & 20.9 & +76.7 & -- \\
policy s1 (plain) & 1 & -- & 15.9 & +76.4 & -- \\
\addlinespace[2pt]
\multicolumn{6}{l}{\emph{Llama-3.1-8B}} \\
legal kl1 & 3 & 3.1 $\pm$ 0.13 & 3.2/3.0 & +11.7/+10.2 & 5\%/5\% \\
med kl1 & 3 & 5.2 $\pm$ 0.08 & 4.9/4.9 & +1.4/+0.9 & no/no \\
med kl10 & 3 & 2.6 $\pm$ 0.10 & 2.6/2.5 & +0.4/+0.9 & 5\%/5\% \\
policy kl1 & 3 & 2.5 $\pm$ 0.17 & 2.3/2.1 & +80.7/+80.4 & 5\%/5\% \\
edit s0 (plain) & 1 & -- & 81.1 & +82.6 & -- \\
edit s1 (plain) & 1 & -- & 86.7 & +84.6 & -- \\
legal s0 (plain) & 1 & -- & 15.6 & +9.4 & -- \\
legal s1 (plain) & 1 & -- & 14.8 & +9.4 & -- \\
med s0 (plain) & 1 & -- & 16.0 & +0.2 & -- \\
med s1 (plain) & 1 & -- & 14.1 & +0.6 & -- \\
policy s0 (plain) & 1 & -- & 31.0 & +80.4 & -- \\
policy s1 (plain) & 1 & -- & 32.4 & +80.0 & -- \\
\bottomrule
\end{tabular}

\end{table}

\begin{table}[tbp]
\centering
\caption{Full fine-tuning and reinforcement learning, all variants (percent, development audit set; columns as in Table~\ref{tab:updtypes}).}
\label{tab:updtypes-full}
\scriptsize
\renewcommand{\arraystretch}{0.92}
\setlength{\tabcolsep}{2pt}
\begin{tabular}{llrrlrrl}%
\toprule
& & \multicolumn{3}{c}{Qwen2.5-7B} & \multicolumn{3}{c}{Llama-3.1-8B} \\
\cmidrule(lr){3-5}\cmidrule(lr){6-8}
Type & Variant & effect & $\lambda$ & certified & effect & $\lambda$ & certified \\
\midrule
\multicolumn{8}{l}{\emph{med update}} \\
LoRA & plain & +2.2 & 9.9 & -- & +1.3 & 15.9 & -- \\
LoRA & anchored $\beta{=}1$ & +2.2 & 3.3 & 5\% (2{,}950) & +2.5 & 5.2 & -- \\
LoRA & anchored $\beta{=}10$ & +2.0 & 1.4 & 5\% (650) & +2.0 & 2.5 & 5\% (1{,}375) \\
full FT & plain, lr $10^{-6}$ & +3.0 & 6.1 & -- & +1.3 & 12.8 & -- \\
full FT & plain, lr $3{\cdot}10^{-6}$ & +2.6 & 9.6 & -- & $-$1.2 & 16.8 & -- \\
full FT & plain, lr $10^{-5}$ & $-$2.2 & 12.7 & -- & $-$5.3 & 24.1 & -- \\
full FT & anchored $\beta{=}10$, lr $10^{-5}$ & +1.4 & 12.2 & -- & $-$2.4 & 16.7 & -- \\
full FT & anchored $\beta{=}10$, 64 anchors, lr $10^{-6}$ & +1.0 & 1.7 & 5\% (850) & +2.1 & 2.8 & 5\% (2{,}225) \\
full FT & anchored $\beta{=}10$, 64 anchors, lr $3{\cdot}10^{-6}$ & +3.4 & 4.4 & -- & +2.4 & 8.3 & -- \\
full FT & anchored $\beta{=}10$, 128 anchors, lr $3{\cdot}10^{-6}$ & +3.7 & 4.3 & -- & +2.9 & 7.5 & -- \\
full FT & plain, lr $3{\cdot}10^{-6}$, seed 1 & +2.1 & 8.4 & -- & +0.6 & 16.8 & -- \\
full FT & anchored $\beta{=}10$, 64 anchors, lr $3{\cdot}10^{-6}$, seed 1 & +3.1 & 4.5 & -- & +1.9 & 7.9 & -- \\
GRPO & reference KL 0.04 & +1.9 & 5.0 & -- & +1.6 & 13.8 & -- \\
GRPO & reference KL 1 & +1.2 & 2.8 & 5\% (1{,}850) & +1.0 & 5.0 & -- \\
GRPO & ref.\ KL 0.04 + anchored $\beta{=}1$ & +2.4 & 2.2 & 5\% (800) & +1.2 & 4.3 & -- \\
GRPO & reference KL 0.04, seed 1 & +1.3 & 7.4 & -- & +2.4 & 10.9 & -- \\
GRPO & reference KL 1, seed 1 & +0.9 & 3.2 & 5\% (2{,}625) & +1.0 & 5.3 & -- \\
GRPO & anchored, seed 1 & +1.6 & 2.3 & 5\% (1{,}050) & +2.0 & 3.8 & 5\% (5{,}375) \\
\midrule
\multicolumn{8}{l}{\emph{policy update}} \\
LoRA & plain & +82.7 & 21.8 & -- & +83.1 & 30.5 & -- \\
LoRA & anchored $\beta{=}1$ & +84.3 & 1.6 & 5\% (775) & +83.9 & 2.7 & 5\% (2{,}825) \\
LoRA & anchored $\beta{=}10$ & +82.3 & 1.1 & 2\% (2{,}325) & +83.5 & 2.0 & 5\% (1{,}675) \\
full FT & plain, lr $10^{-6}$ & +82.7 & 15.4 & -- & +81.9 & 22.7 & -- \\
full FT & plain, lr $3{\cdot}10^{-6}$ & +83.9 & 24.8 & -- & +83.9 & 39.2 & -- \\
full FT & plain, lr $10^{-5}$ & +83.9 & 38.3 & -- & +83.9 & 71.0 & -- \\
full FT & anchored $\beta{=}10$, lr $10^{-5}$ & +82.7 & 14.4 & -- & +83.9 & 20.2 & -- \\
full FT & anchored $\beta{=}10$, 64 anchors, lr $10^{-6}$ & +81.9 & 1.6 & 5\% (650) & +83.5 & 2.9 & 5\% (2{,}150) \\
full FT & anchored $\beta{=}10$, 64 anchors, lr $3{\cdot}10^{-6}$ & +82.3 & 4.5 & -- & +83.5 & 9.3 & -- \\
full FT & anchored $\beta{=}10$, 128 anchors, lr $3{\cdot}10^{-6}$ & +83.1 & 4.1 & -- & +83.9 & 9.7 & -- \\
full FT & plain, lr $3{\cdot}10^{-6}$, seed 1 & +82.7 & 24.3 & -- & +82.7 & 38.9 & -- \\
full FT & anchored $\beta{=}10$, 64 anchors, lr $3{\cdot}10^{-6}$, seed 1 & +82.3 & 5.1 & -- & +83.5 & 9.8 & -- \\
GRPO & reference KL 0.04 & +80.3 & 19.7 & -- & +82.7 & 27.3 & -- \\
GRPO & reference KL 1 & +41.7 & 2.1 & 5\% (1{,}100) & +49.6 & 8.3 & -- \\
GRPO & ref.\ KL 0.04 + anchored $\beta{=}1$ & +84.6 & 1.5 & 5\% (450) & +83.5 & 2.0 & 5\% (1{,}025) \\
GRPO & reference KL 0.04, seed 1 & +78.0 & 8.6 & -- & +82.7 & 19.2 & -- \\
GRPO & reference KL 1, seed 1 & +80.3 & 6.8 & -- & +83.1 & 24.4 & -- \\
GRPO & anchored, seed 1 & +82.7 & 1.2 & 2\% (1{,}625) & +81.9 & 2.0 & 5\% (425) \\
\bottomrule
\end{tabular}

\par\vspace{0.6\baselineskip}
\caption{All trained update variants, Mistral-7B and Qwen2.5-32B (columns as in Table~\ref{tab:variants}).}
\label{tab:variants-other}
\scriptsize
\setlength{\tabcolsep}{1.5pt}
\renewcommand{\arraystretch}{0.92}
\begin{tabular}[t]{lrrrrrrrr}
\toprule
Variant & effect & $\lambda$ & $\lambda_\tau$ & $U$ & $\varepsilon$ & $n$ & IQR & \% \\
\midrule
\multicolumn{9}{l}{\emph{Mistral-7B, med}} \\
base$^\ast$ & +5.3 & 17.0 & 14.9 & 18.6 & no & & & \\
kl1$^\ast$ & +4.2 & 10.9 & 9.0 & 12.3 & no & & & \\
kl10$^\ast$ & +2.1 & 9.6 & 7.7 & 10.9 & no & & & \\
kl1a64 & +4.4 & 9.1 & 7.1 & 10.4 & no & & & \\
kl1let & +4.4 & 11.6 & 9.5 & 13.0 & no & & & \\
kl1lr3e-5 & +3.8 & 6.6 & 4.7 & 7.7 & no & & & \\
lr3e-5 & +5.9 & 14.9 & 12.7 & 16.5 & no & & & \\
\multicolumn{9}{l}{\emph{Mistral-7B, policy}} \\
base$^\ast$ & +82.7 & 33.0 & 31.3 & 34.9 & no & & & \\
kl1$^\ast$ & +79.9 & 9.5 & 7.3 & 10.8 & no & & & \\
kl10$^\ast$ & +79.5 & 10.7 & 8.4 & 12.0 & no & & & \\
\bottomrule
\end{tabular}\hspace{1.2em}%
\begin{tabular}[t]{lrrrrrrrr}
\toprule
Variant & effect & $\lambda$ & $\lambda_\tau$ & $U$ & $\varepsilon$ & $n$ & IQR & \% \\
\midrule
\multicolumn{9}{l}{\emph{Mistral-7B, policy (cont.)}} \\
kl1a64 & +83.5 & 7.6 & 5.6 & 8.8 & no & & & \\
kl1let & +81.1 & 10.7 & 8.4 & 12.0 & no & & & \\
kl1lr3e-5 & +82.3 & 4.2 & 2.4 & 5.2 & no & & & \\
lr3e-5 & +82.7 & 23.1 & 21.4 & 24.9 & no & & & \\
\multicolumn{9}{l}{\emph{Qwen2.5-32B, med}} \\
base$^\ast$ & +3.6 & 8.0 & 7.4 & 9.2 & no & & & \\
kl1$^\ast$ & +2.5 & 2.3 & 1.9 & 3.1 & 5 & 825 & 725--1{,}125 & 100 \\
kl1s1 & +2.0 & 2.2 & 1.6 & 2.9 & 5 & 762 & 618--968 & 100 \\
\multicolumn{9}{l}{\emph{Qwen2.5-32B, policy}} \\
base$^\ast$ & +72.4 & 9.9 & 9.0 & 11.1 & no & & & \\
kl1$^\ast$ & +72.4 & 1.0 & 0.4 & 1.5 & 2 & 2{,}887 & 1{,}950--3{,}875 & 100 \\
kl1s1 & +71.7 & 1.1 & 0.5 & 1.7 & 2 & 3{,}262 & 2{,}700--4{,}600 & 100 \\
\bottomrule
\end{tabular}

\end{table}

\begin{table}[tbp]
\centering
\caption{Scale and model family (percent, audit set): medical and policy updates, plain LoRA versus anchoring at $\beta=1$. Floor: self-disagreement under re-batching.}
\label{tab:scale}
\small
\setlength{\tabcolsep}{4pt}
\begin{tabular}{lrrrrr}%
\toprule
& & \multicolumn{2}{c}{LoRA} & \multicolumn{2}{c}{anchored ($\beta{=}1$)} \\
\cmidrule(lr){3-4}\cmidrule(lr){5-6}
Update & floor & effect & $\lambda$ & effect & $\lambda$ \\
\midrule
\multicolumn{6}{l}{\emph{Qwen2.5-7B}} \\
med & 0.7 & +2.2 & 9.9 & +2.2 & 3.3 \\
policy & 0.7 & +82.7 & 21.8 & +84.3 & 1.6 \\
\addlinespace[2pt]
\multicolumn{6}{l}{\emph{Llama-3.1-8B}} \\
med & 0.9 & +1.3 & 15.9 & +2.5 & 5.2 \\
policy & 0.9 & +83.1 & 30.5 & +83.9 & 2.7 \\
\addlinespace[2pt]
\multicolumn{6}{l}{\emph{Mistral-7B}} \\
med & 0.8 & +5.3 & 17.0 & +4.2 & 10.9 \\
policy & 0.8 & +82.7 & 33.0 & +79.9 & 9.5 \\
\addlinespace[2pt]
\multicolumn{6}{l}{\emph{Qwen2.5-32B}} \\
med & 0.4 & +3.6 & 8.0 & +2.5 & 2.3 \\
policy & 0.4 & +72.4 & 9.9 & +72.4 & 1.0 \\
\addlinespace[2pt]
\multicolumn{6}{l}{\emph{Qwen2.5-72B}} \\
med & 0.4 & +3.5 & 6.9 & +3.1 & 2.8 \\
policy & 0.4 & +70.9 & 15.1 & +70.5 & 1.7 \\
\bottomrule
\end{tabular}

\par\vspace{0.6\baselineskip}
\centering
\caption{Edit update (400 CounterFact rewrites; percent). Efficacy gain: new-object decisions on paraphrases of edited facts. Neighbors: prompts about other subjects that share the edited relation and options (3.6\% of audit traffic).}
\label{tab:edit}
\footnotesize
\setlength{\tabcolsep}{4pt}
\begin{tabular}{lrrrrrl}%
\toprule
Method & efficacy & $\lambda$ & \hd{$\lambda$ excl.\\neighbors} & \hd{neighbors\\changed} & \hd{out-of-scope\\acc.} & ES / PS / NS \\
\midrule
\multicolumn{7}{l}{\emph{Qwen2.5-7B}} \\
LoRA, counterfactual only & +71.4 & 98.1 & 98.3 & 91.7 & $-$68.2 & -- \\
\quad + anchoring ($\beta{=}1$) & +71.0 & 5.3 & 2.1 & 89.4 & $-$2.6 & -- \\
LoRA + 400 true facts & +29.3 & 12.1 & 11.4 & 30.3 & $-$2.5 & -- \\
\quad + anchoring ($\beta{=}1$) & +25.1 & 2.8 & 1.8 & 29.8 & $-$1.0 & -- \\
MEMIT & +56.0 & 2.4 & 2.4 & 3.9 & $-$0.1 & 100 / 87 / 84 \\
AlphaEdit & +56.8 & 3.6 & 3.5 & 4.4 & $-$0.3 & 100 / 88 / 84 \\
\midrule
\multicolumn{7}{l}{\emph{Llama-3.1-8B}} \\
LoRA, counterfactual only & +85.7 & 81.3 & 81.1 & 86.4 & $-$52.0 & -- \\
\quad + anchoring ($\beta{=}1$) & +86.1 & 7.3 & 4.1 & 90.3 & $-$2.6 & -- \\
LoRA + 400 true facts & +45.9 & 32.1 & 31.5 & 48.3 & $-$11.7 & -- \\
\quad + anchoring ($\beta{=}1$) & +50.2 & 5.7 & 4.3 & 42.3 & $-$1.5 & -- \\
MEMIT & +69.1 & 14.2 & 14.1 & 16.4 & $-$2.5 & 100 / 99 / 73 \\
MEMIT, cov.\ weight $6\cdot10^4$ & +73.4 & 7.7 & 7.5 & 12.7 & $-$0.7 & 100 / 97 / 76 \\
AlphaEdit & +62.5 & 18.2 & 18.0 & 23.3 & $-$4.5 & 100 / 99 / 70 \\
\bottomrule
\end{tabular}

\end{table}

\begin{table}[tbp]
\centering
\caption{Free-form traffic outside every scope (percent). Changed: decision differs from the incumbent's under the same batching; semantic: NQ changes judged non-equivalent by Qwen2.5-32B-Instruct.}
\label{tab:freeform}
\scriptsize
\setlength{\tabcolsep}{4pt}
\begin{tabular}{lrrrrr}%
\toprule
Variant & \hd{GSM8K\\changed} & \hd{GSM8K\\acc.} & \hd{NQ changed\\(string)} & \hd{NQ changed\\(semantic)} & \hd{NQ\\acc.} \\
\midrule
\multicolumn{6}{l}{\emph{Qwen2.5-7B}} \\
incumbent re-batched (floor) & 4.0 & +0.1 & 7.0 & 5.2 & $-$0.1 \\
zero adapter, same batching & 0.0 & +0.0 & 0.0 & 0.0 & +0.0 \\
random adapter, $\sigma_B{=}10^{-4}$ & 4.2 & +0.0 & 9.5 & -- & +0.0 \\
random adapter, $\sigma_B{=}10^{-3}$ & 2.9 & +0.5 & 10.6 & -- & +0.1 \\
medical, LoRA & 8.0 & +0.8 & 60.5 & 46.5 & +2.3 \\
medical, anchored ($\beta{=}1$) & 6.0 & $-$0.9 & 26.9 & 20.1 & +1.0 \\
policy, LoRA & 10.5 & $-$2.1 & 97.6 & 61.3 & +5.0 \\
policy, anchored ($\beta{=}1$) & 4.1 & $-$0.1 & 13.1 & 9.8 & +0.2 \\
medical, anchored + free-form KL & 5.2 & $-$0.4 & 22.7 & 16.9 & +0.6 \\
policy, anchored + free-form KL & 4.1 & +0.0 & 12.0 & 8.9 & +0.1 \\
\midrule
\multicolumn{6}{l}{\emph{Llama-3.1-8B}} \\
incumbent re-batched (floor) & 6.8 & +0.5 & 7.9 & 4.6 & $-$0.2 \\
zero adapter, same batching & 0.0 & +0.0 & 0.0 & 0.0 & +0.0 \\
random adapter, $\sigma_B{=}10^{-4}$ & 6.9 & +0.5 & 9.8 & -- & +0.2 \\
random adapter, $\sigma_B{=}10^{-3}$ & 7.4 & $-$0.7 & 11.7 & -- & +0.2 \\
medical, LoRA & 16.1 & $-$1.7 & 68.3 & 48.5 & $-$2.7 \\
medical, anchored ($\beta{=}1$) & 12.5 & $-$0.9 & 38.2 & 26.4 & $-$0.3 \\
policy, LoRA & 14.5 & $-$0.6 & 48.2 & 31.3 & $-$0.9 \\
policy, anchored ($\beta{=}1$) & 8.5 & $-$1.1 & 19.6 & 13.0 & +0.2 \\
medical, anchored + free-form KL & 9.0 & +0.1 & 30.6 & 21.1 & +0.6 \\
policy, anchored + free-form KL & 8.0 & +0.1 & 14.1 & 8.7 & +0.4 \\
\bottomrule
\end{tabular}

\par\vspace{0.6\baselineskip}
\centering
\caption{Refusal decisions on 720 unsafe prompts after the medical update, judged by Qwen2.5-32B-Instruct with the XSTest rubric (percent).}
\label{tab:safety}
\scriptsize
\setlength{\tabcolsep}{4pt}
\begin{tabular}{lrrrrr}%
\toprule
Variant & \hd{refusal\\rate} & changed & \hd{newly\\complied} & \hd{newly\\refused} & \hd{changed\\XSTest / AdvBench} \\
\midrule
\multicolumn{6}{l}{\emph{Qwen2.5-7B}} \\
incumbent & 95.7 & -- & -- & -- & -- \\
incumbent re-batched (floor) & 95.6 & 1.0 & 0.6 & 0.4 & 3.0 / 0.2 \\
LoRA seed 0 & 95.3 & 2.1 & 1.2 & 0.8 & 5.5 / 0.8 \\
LoRA seed 1 & 96.5 & 1.4 & 0.3 & 1.1 & 4.5 / 0.2 \\
anchored seed 0 & 96.7 & 1.8 & 0.4 & 1.4 & 6.5 / 0.0 \\
anchored seed 1 & 95.8 & 1.5 & 0.7 & 0.8 & 5.0 / 0.2 \\
anchored seed 2 & 95.8 & 1.2 & 0.6 & 0.7 & 4.5 / 0.0 \\
\midrule
\multicolumn{6}{l}{\emph{Llama-3.1-8B}} \\
incumbent & 97.1 & -- & -- & -- & -- \\
incumbent re-batched (floor) & 97.1 & 0.3 & 0.1 & 0.1 & 0.5 / 0.2 \\
LoRA seed 0 & 98.8 & 1.9 & 0.1 & 1.8 & 1.5 / 2.1 \\
LoRA seed 1 & 97.9 & 1.9 & 0.6 & 1.4 & 1.5 / 2.1 \\
anchored seed 0 & 97.5 & 1.5 & 0.6 & 1.0 & 1.5 / 1.5 \\
anchored seed 1 & 97.1 & 1.9 & 1.0 & 1.0 & 2.5 / 1.7 \\
anchored seed 2 & 97.6 & 1.9 & 0.7 & 1.2 & 2.5 / 1.7 \\
\bottomrule
\end{tabular}

\end{table}

\begin{table}[p]
\centering
\caption{All trained update variants, Qwen2.5-7B (percent, development audit set; sweep configurations are in Tables~\ref{tab:controls} and~\ref{tab:sweep}). Rows are grouped by update; tags follow the code release without the update prefix. $^\ast$: the four pre-declared promotion candidates (plain LoRA and anchoring with $\beta\in\{0.1,1,10\}$); all other rows are diagnostics. $U$: final upper confidence bound on $\lambda$ at level $\alpha/4$ (98.75\%). $\varepsilon$: smallest certified $\varepsilon$ (percent) at level $\alpha/4$; $n$ and IQR of $n$: median and interquartile range of the number of out-of-scope samples needed over 20 random audit orders; \%: fraction of orders that certify. \emph{kl}$\beta$: envelope-anchored; \emph{perm}: options shuffled per training example; \emph{ho}: anchoring on half of the domains, audited on all; \emph{replay}: 1{,}000 anchor items labeled with the incumbent's own decisions; \emph{mild}: learning rate $3\times10^{-5}$ and three epochs.}
\label{tab:variants}
\scriptsize
\setlength{\tabcolsep}{1.5pt}
\renewcommand{\arraystretch}{0.92}
\resizebox{\linewidth}{!}{%
\begin{tabular}[t]{lrrrrrrrr}
\toprule
Variant & effect & $\lambda$ & $\lambda_\tau$ & $U$ & $\varepsilon$ & $n$ & IQR & \% \\
\midrule
\multicolumn{9}{l}{\emph{edit}} \\
alphaedit & +56.8 & 3.6 & 2.1 & 4.4 & 5 & 3{,}825 & 2{,}831--4{,}656 & 100 \\
base & +71.4 & 98.1 & 98.3 & 98.6 & no & & & \\
kl1 & +71.0 & 5.3 & 4.3 & 6.3 & no & & & \\
memit & +56.0 & 2.4 & 1.1 & 3.1 & 5 & 1{,}087 & 650--1{,}487 & 100 \\
memitw2k & +57.1 & 3.5 & 2.1 & 4.3 & 5 & 3{,}337 & 3{,}012--4{,}531 & 100 \\
mild & +64.5 & 96.2 & 96.7 & 97.0 & no & & & \\
mix & +29.3 & 12.1 & 10.7 & 13.4 & no & & & \\
mixkl1 & +25.1 & 2.8 & 1.7 & 3.6 & 5 & 1{,}575 & 1{,}106--1{,}987 & 100 \\
perm & +67.6 & 98.2 & 98.5 & 98.7 & no & & & \\
\multicolumn{9}{l}{\emph{legal}} \\
anchsv & +7.0 & 2.3 & 1.0 & 3.0 & 5 & 650 & 587--875 & 100 \\
base$^\ast$ & +8.1 & 9.5 & 8.1 & 10.8 & no & & & \\
kl1$^\ast$ & +6.9 & 2.2 & 1.0 & 2.9 & 5 & 575 & 518--762 & 100 \\
plainsv & +8.2 & 9.2 & 7.8 & 10.5 & no & & & \\
\multicolumn{9}{l}{\emph{med}} \\
anchsv & +2.1 & 1.6 & 0.6 & 2.3 & 5 & 500 & 412--650 & 100 \\
base$^\ast$ & +2.2 & 9.9 & 8.6 & 11.2 & no & & & \\
dpo & +2.7 & 9.0 & 7.7 & 10.4 & no & & & \\
dpokl1 & +2.3 & 2.7 & 1.5 & 3.4 & 5 & 1{,}125 & 812--1{,}487 & 100 \\
fd1 & +2.1 & 5.3 & 4.1 & 6.3 & no & & & \\
fd10 & +1.1 & 2.2 & 1.1 & 2.9 & 5 & 762 & 600--868 & 100 \\
fft & $-$2.2 & 12.7 & 11.4 & 14.2 & no & & & \\
fftkl1 & $-$1.8 & 11.7 & 10.5 & 13.1 & no & & & \\
fftkl10 & +1.4 & 12.2 & 11.0 & 13.6 & no & & & \\
fftkl10ab128lr3e-6 & +3.7 & 4.3 & 3.1 & 5.2 & no & & & \\
fftkl10ab64lr1e-6 & +1.0 & 1.7 & 0.7 & 2.4 & 5 & 575 & 518--650 & 100 \\
fftkl10ab64lr3e-6 & +3.4 & 4.4 & 3.3 & 5.4 & no & & & \\
fftkl10ab64lr3e-6s1 & +3.1 & 4.5 & 3.4 & 5.5 & no & & & \\
fftkl10lr3e-6 & +2.5 & 5.4 & 4.1 & 6.4 & no & & & \\
fftkl1lr3e-6 & +2.1 & 6.5 & 5.2 & 7.7 & no & & & \\
fftlr1e-6 & +3.0 & 6.1 & 5.0 & 7.2 & no & & & \\
fftlr3e-6 & +2.6 & 9.6 & 8.4 & 11.0 & no & & & \\
fftlr3e-6s1 & +2.1 & 8.4 & 7.2 & 9.7 & no & & & \\
genkl1 & +1.9 & 8.4 & 7.2 & 9.7 & no & & & \\
grpo & +1.9 & 5.0 & 3.8 & 6.1 & no & & & \\
grpokl1 & +2.4 & 2.2 & 1.2 & 2.9 & 5 & 775 & 637--1{,}200 & 100 \\
grpokl1s1 & +1.6 & 2.3 & 1.1 & 3.0 & 5 & 825 & 606--1{,}025 & 100 \\
grporef1 & +1.2 & 2.8 & 1.7 & 3.6 & 5 & 1{,}587 & 1{,}187--1{,}962 & 100 \\
grporef1s1 & +0.9 & 3.2 & 2.0 & 4.1 & 5 & 2{,}437 & 1{,}631--2{,}868 & 100 \\
grpos1 & +1.3 & 7.4 & 6.1 & 8.6 & no & & & \\
kl0.1$^\ast$ & +1.8 & 6.8 & 5.6 & 8.0 & no & & & \\
kl1$^\ast$ & +2.2 & 3.3 & 2.1 & 4.2 & 5 & 2{,}375 & 1{,}487--2{,}875 & 100 \\
kl10$^\ast$ & +2.0 & 1.4 & 0.3 & 2.0 & 5 & 475 & 337--600 & 100 \\
kl10ho & +2.7 & 1.7 & 0.7 & 2.3 & 5 & 525 & 450--568 & 100 \\
kl10hoB & +2.1 & 1.4 & 0.5 & 2.0 & 5 & 450 & 368--562 & 100 \\
kl1ho & +2.8 & 3.7 & 2.5 & 4.6 & 5 & 3{,}687 & 2{,}762--4{,}712 & 100 \\
kl1hoB & +2.0 & 3.5 & 2.3 & 4.4 & 5 & 3{,}112 & 2{,}212--3{,}893 & 100 \\
kl1let & +2.8 & 5.1 & 3.9 & 6.1 & no & & & \\
klall & +2.7 & 3.4 & 2.1 & 4.2 & 5 & 2{,}325 & 1{,}862--3{,}037 & 100 \\
klall10 & +1.7 & 1.5 & 0.5 & 2.1 & 5 & 525 & 443--650 & 100 \\
l2d0.1 & +2.7 & 8.5 & 7.2 & 9.7 & no & & & \\
l2d1 & +3.5 & 6.7 & 5.4 & 7.8 & no & & & \\
lr3e-5 & +3.2 & 6.3 & 5.1 & 7.4 & no & & & \\
mcffkl1 & +2.3 & 2.9 & 1.8 & 3.7 & 5 & 1{,}550 & 875--2{,}106 & 100 \\
\bottomrule
\end{tabular}\hspace{1.2em}%
\begin{tabular}[t]{lrrrrrrrr}
\toprule
Variant & effect & $\lambda$ & $\lambda_\tau$ & $U$ & $\varepsilon$ & $n$ & IQR & \% \\
\midrule
\multicolumn{9}{l}{\emph{med (cont.)}} \\
mcgenkl1 & +2.9 & 3.8 & 2.5 & 4.7 & 5 & 5{,}125 & 4{,}225--5{,}975 & 100 \\
plainsv & +2.7 & 9.8 & 8.5 & 11.1 & no & & & \\
r4 & +4.1 & 8.1 & 6.9 & 9.4 & no & & & \\
r64 & +0.6 & 10.8 & 9.6 & 12.2 & no & & & \\
replay & +2.3 & 6.5 & 5.3 & 7.6 & no & & & \\
scale0.25 & +2.5 & 4.8 & 3.6 & 5.8 & no & & & \\
scale0.5 & +3.2 & 7.0 & 5.8 & 8.2 & no & & & \\
scale0.75 & +2.7 & 8.4 & 7.2 & 9.7 & no & & & \\
\multicolumn{9}{l}{\emph{policy}} \\
anchsv & +82.7 & 1.2 & 0.2 & 1.8 & 2 & 4{,}837 & 4{,}137--6{,}306 & 100 \\
base$^\ast$ & +82.7 & 21.8 & 20.5 & 23.5 & no & & & \\
dpo & +86.2 & 34.1 & 33.0 & 36.0 & no & & & \\
dpokl1 & +81.5 & 1.4 & 0.4 & 1.9 & 2 & 7{,}575 & 6{,}837--8{,}412 & 100 \\
fd1 & +80.3 & 4.5 & 3.0 & 5.4 & no & & & \\
fd10 & +2.4 & 1.4 & 0.4 & 2.0 & 2 & 9{,}175 & 7{,}425--10{,}275 & 55 \\
fft & +83.9 & 38.3 & 37.3 & 40.2 & no & & & \\
fftkl1 & +84.3 & 14.1 & 12.6 & 15.5 & no & & & \\
fftkl10 & +82.7 & 14.4 & 13.0 & 15.8 & no & & & \\
fftkl10ab128lr3e-6 & +83.1 & 4.1 & 2.7 & 5.0 & 5 & 9{,}362 & 5{,}375--9{,}775 & 60 \\
fftkl10ab64lr1e-6 & +81.9 & 1.6 & 0.5 & 2.2 & 5 & 450 & 375--600 & 100 \\
fftkl10ab64lr3e-6 & +82.3 & 4.5 & 3.1 & 5.4 & no & & & \\
fftkl10ab64lr3e-6s1 & +82.3 & 5.1 & 3.7 & 6.1 & no & & & \\
fftkl10lr3e-6 & +81.9 & 6.2 & 4.8 & 7.3 & no & & & \\
fftkl1lr3e-6 & +83.1 & 6.4 & 5.0 & 7.5 & no & & & \\
fftlr1e-6 & +82.7 & 15.4 & 14.1 & 16.9 & no & & & \\
fftlr3e-6 & +83.9 & 24.8 & 23.7 & 26.6 & no & & & \\
fftlr3e-6s1 & +82.7 & 24.3 & 23.1 & 26.1 & no & & & \\
genkl1 & +83.9 & 19.8 & 18.6 & 21.5 & no & & & \\
grpo & +80.3 & 19.7 & 18.6 & 21.4 & no & & & \\
grpokl1 & +84.6 & 1.5 & 0.3 & 2.1 & 5 & 375 & 287--487 & 100 \\
grpokl1s1 & +82.7 & 1.2 & 0.2 & 1.7 & 2 & 4{,}550 & 2{,}831--5{,}087 & 100 \\
grporef1 & +41.7 & 2.1 & 0.8 & 2.7 & 5 & 687 & 481--931 & 100 \\
grporef1s1 & +80.3 & 6.8 & 5.3 & 7.9 & no & & & \\
grpos1 & +78.0 & 8.6 & 7.1 & 9.8 & no & & & \\
kl0.1$^\ast$ & +84.6 & 2.6 & 1.3 & 3.4 & 5 & 1{,}200 & 700--1{,}462 & 100 \\
kl1$^\ast$ & +84.3 & 1.6 & 0.4 & 2.2 & 5 & 400 & 287--531 & 100 \\
kl10$^\ast$ & +82.3 & 1.1 & 0.3 & 1.7 & 2 & 4{,}437 & 3{,}237--5{,}312 & 100 \\
kl10ho & +81.9 & 1.3 & 0.3 & 1.8 & 2 & 5{,}575 & 3{,}700--7{,}725 & 100 \\
kl10hoB & +81.5 & 1.2 & 0.2 & 1.8 & 2 & 4{,}900 & 3{,}131--6{,}025 & 100 \\
kl1ho & +83.9 & 1.7 & 0.5 & 2.3 & 5 & 500 & 375--650 & 100 \\
kl1hoB & +83.1 & 1.5 & 0.4 & 2.1 & 5 & 425 & 250--462 & 100 \\
klall & +82.7 & 1.6 & 0.4 & 2.2 & 5 & 500 & 412--550 & 100 \\
klall10 & +81.1 & 1.3 & 0.3 & 1.8 & 2 & 6{,}025 & 4{,}531--7{,}525 & 100 \\
l2d0.1 & +83.1 & 11.2 & 9.8 & 12.6 & no & & & \\
l2d1 & +81.5 & 10.8 & 9.4 & 12.1 & no & & & \\
lr3e-5 & +82.3 & 13.6 & 12.2 & 15.0 & no & & & \\
mcffkl1 & +83.9 & 1.7 & 0.5 & 2.3 & 5 & 525 & 450--712 & 100 \\
mcgenkl1 & +83.5 & 1.8 & 0.6 & 2.4 & 5 & 587 & 406--737 & 100 \\
perm & +82.7 & 7.9 & 6.5 & 9.0 & no & & & \\
permkl1 & +83.1 & 1.8 & 0.6 & 2.4 & 5 & 512 & 425--631 & 100 \\
plainsv & +82.7 & 21.7 & 20.5 & 23.4 & no & & & \\
replay & +83.9 & 6.6 & 5.1 & 7.7 & no & & & \\
scale0.25 & +35.8 & 4.4 & 2.9 & 5.4 & no & & & \\
scale0.5 & +79.9 & 8.1 & 6.7 & 9.3 & no & & & \\
scale0.75 & +81.9 & 13.7 & 12.3 & 15.2 & no & & & \\
\bottomrule
\end{tabular}}

\end{table}

\begin{table}[p]
\centering
\caption{All trained update variants, Llama-3.1-8B (columns as in Table~\ref{tab:variants}).}
\label{tab:variants-llama}
\scriptsize
\setlength{\tabcolsep}{1.5pt}
\renewcommand{\arraystretch}{0.92}
\resizebox{\linewidth}{!}{%
\begin{tabular}[t]{lrrrrrrrr}
\toprule
Variant & effect & $\lambda$ & $\lambda_\tau$ & $U$ & $\varepsilon$ & $n$ & IQR & \% \\
\midrule
\multicolumn{9}{l}{\emph{edit}} \\
alphaedit & +62.5 & 18.2 & 15.1 & 19.8 & no & & & \\
base & +85.7 & 81.3 & 81.7 & 82.8 & no & & & \\
kl1 & +86.1 & 7.3 & 5.2 & 8.4 & no & & & \\
memit & +69.1 & 14.2 & 11.2 & 15.6 & no & & & \\
memitw60k & +73.4 & 7.7 & 4.9 & 8.8 & no & & & \\
mild & +76.4 & 92.6 & 93.3 & 93.6 & no & & & \\
mix & +45.9 & 32.1 & 29.6 & 33.9 & no & & & \\
mixkl1 & +50.2 & 5.7 & 3.3 & 6.7 & no & & & \\
\multicolumn{9}{l}{\emph{legal}} \\
anchsv & +12.3 & 3.5 & 1.2 & 4.3 & 5 & 3{,}012 & 1{,}525--4{,}162 & 100 \\
base$^\ast$ & +8.8 & 15.6 & 12.7 & 17.1 & no & & & \\
kl1$^\ast$ & +11.7 & 3.1 & 1.0 & 3.9 & 5 & 1{,}900 & 1{,}331--2{,}662 & 100 \\
plainsv & +8.8 & 16.3 & 13.4 & 17.8 & no & & & \\
\multicolumn{9}{l}{\emph{med}} \\
anchsv & +2.4 & 2.4 & 0.7 & 3.2 & 5 & 950 & 700--1{,}062 & 100 \\
base$^\ast$ & +1.3 & 15.9 & 13.0 & 17.5 & no & & & \\
dpo & +1.5 & 17.4 & 14.8 & 19.1 & no & & & \\
dpokl1 & +1.8 & 4.4 & 2.1 & 5.4 & no & & & \\
fd1 & +1.5 & 4.2 & 1.8 & 5.1 & no & & & \\
fd10 & +0.3 & 2.5 & 0.6 & 3.3 & 5 & 1{,}100 & 762--1{,}212 & 100 \\
fft & $-$5.3 & 24.1 & 21.5 & 25.9 & no & & & \\
fftkl1 & $-$2.4 & 15.9 & 13.0 & 17.5 & no & & & \\
fftkl10 & $-$2.4 & 16.7 & 13.7 & 18.3 & no & & & \\
fftkl10ab128lr3e-6 & +2.9 & 7.5 & 4.5 & 8.7 & no & & & \\
fftkl10ab64lr1e-6 & +2.1 & 2.8 & 0.8 & 3.7 & 5 & 1{,}462 & 1{,}300--1{,}762 & 100 \\
fftkl10ab64lr3e-6 & +2.4 & 8.3 & 5.3 & 9.5 & no & & & \\
fftkl10ab64lr3e-6s1 & +1.9 & 7.9 & 5.0 & 9.1 & no & & & \\
fftkl10lr3e-6 & +2.6 & 9.6 & 6.8 & 10.9 & no & & & \\
fftkl1lr3e-6 & +2.1 & 9.8 & 6.9 & 11.2 & no & & & \\
fftlr1e-6 & +1.3 & 12.8 & 10.0 & 14.3 & no & & & \\
fftlr3e-6 & $-$1.2 & 16.8 & 14.3 & 18.5 & no & & & \\
fftlr3e-6s1 & +0.6 & 16.8 & 13.9 & 18.4 & no & & & \\
genkl1 & +2.3 & 13.4 & 10.5 & 14.9 & no & & & \\
grpo & +1.6 & 13.8 & 10.9 & 15.4 & no & & & \\
grpokl1 & +1.2 & 4.3 & 2.0 & 5.2 & no & & & \\
grpokl1s1 & +2.0 & 3.8 & 1.7 & 4.7 & 5 & 4{,}812 & 3{,}587--5{,}812 & 100 \\
grporef1 & +1.0 & 5.0 & 2.4 & 6.0 & no & & & \\
grporef1s1 & +1.0 & 5.3 & 2.7 & 6.3 & no & & & \\
grpos1 & +2.4 & 10.9 & 8.1 & 12.3 & no & & & \\
kl1$^\ast$ & +2.5 & 5.2 & 2.8 & 6.3 & no & & & \\
kl10$^\ast$ & +2.0 & 2.5 & 0.7 & 3.3 & 5 & 950 & 706--1{,}343 & 100 \\
kl10ho & +2.7 & 4.2 & 1.9 & 5.2 & no & & & \\
kl10hoB & +2.8 & 2.4 & 0.6 & 3.1 & 5 & 937 & 618--1{,}093 & 100 \\
kl1ho & +1.7 & 7.4 & 4.6 & 8.6 & no & & & \\
kl1hoB & +2.5 & 5.1 & 2.8 & 6.2 & no & & & \\
klall & +2.1 & 5.1 & 2.7 & 6.2 & no & & & \\
klall10 & +1.3 & 2.1 & 0.4 & 2.9 & 5 & 762 & 625--812 & 100 \\
l2d0.1 & +1.5 & 14.3 & 11.5 & 15.8 & no & & & \\
\bottomrule
\end{tabular}\hspace{1.2em}%
\begin{tabular}[t]{lrrrrrrrr}
\toprule
Variant & effect & $\lambda$ & $\lambda_\tau$ & $U$ & $\varepsilon$ & $n$ & IQR & \% \\
\midrule
\multicolumn{9}{l}{\emph{med (cont.)}} \\
l2d1 & +1.5 & 13.5 & 10.8 & 15.0 & no & & & \\
mcffkl1 & +1.9 & 4.5 & 2.1 & 5.5 & no & & & \\
mcgenkl1 & +2.1 & 6.7 & 3.9 & 7.8 & no & & & \\
plainsv & +1.5 & 15.9 & 13.0 & 17.5 & no & & & \\
scale0.25 & +1.2 & 6.6 & 4.0 & 7.8 & no & & & \\
scale0.5 & +1.5 & 10.9 & 8.2 & 12.3 & no & & & \\
scale0.75 & +1.9 & 13.8 & 11.0 & 15.3 & no & & & \\
\multicolumn{9}{l}{\emph{policy}} \\
anchsv & +83.5 & 1.9 & 0.2 & 2.6 & 5 & 625 & 525--900 & 100 \\
base$^\ast$ & +83.1 & 30.5 & 28.1 & 32.4 & no & & & \\
fd1 & +81.9 & 7.5 & 4.5 & 8.7 & no & & & \\
fd10 & $-$0.4 & 2.0 & 0.1 & 2.6 & 5 & 775 & 543--931 & 100 \\
fft & +83.9 & 71.0 & 70.6 & 72.8 & no & & & \\
fftkl1 & +84.3 & 19.6 & 16.1 & 21.2 & no & & & \\
fftkl10 & +83.9 & 20.2 & 16.9 & 21.8 & no & & & \\
fftkl10ab128lr3e-6 & +83.9 & 9.7 & 6.1 & 11.0 & no & & & \\
fftkl10ab64lr1e-6 & +83.5 & 2.9 & 0.6 & 3.7 & 5 & 1{,}762 & 1{,}093--1{,}993 & 100 \\
fftkl10ab64lr3e-6 & +83.5 & 9.3 & 5.9 & 10.6 & no & & & \\
fftkl10ab64lr3e-6s1 & +83.5 & 9.8 & 6.4 & 11.1 & no & & & \\
fftkl10lr3e-6 & +83.5 & 11.1 & 7.6 & 12.4 & no & & & \\
fftkl1lr3e-6 & +83.5 & 10.2 & 6.8 & 11.5 & no & & & \\
fftlr1e-6 & +81.9 & 22.7 & 19.7 & 24.4 & no & & & \\
fftlr3e-6 & +83.9 & 39.2 & 37.0 & 41.2 & no & & & \\
fftlr3e-6s1 & +82.7 & 38.9 & 36.6 & 40.9 & no & & & \\
genkl1 & +84.6 & 30.0 & 27.4 & 31.9 & no & & & \\
grpo & +82.7 & 27.3 & 24.9 & 29.1 & no & & & \\
grpokl1 & +83.5 & 2.0 & 0.2 & 2.7 & 5 & 575 & 412--937 & 100 \\
grpokl1s1 & +81.9 & 2.0 & 0.3 & 2.7 & 5 & 725 & 487--881 & 100 \\
grporef1 & +49.6 & 8.3 & 5.4 & 9.5 & no & & & \\
grporef1s1 & +83.1 & 24.4 & 22.2 & 26.2 & no & & & \\
grpos1 & +82.7 & 19.2 & 16.5 & 20.9 & no & & & \\
kl1$^\ast$ & +83.9 & 2.7 & 0.4 & 3.4 & 5 & 1{,}487 & 1{,}181--1{,}650 & 100 \\
kl10$^\ast$ & +83.5 & 2.0 & 0.3 & 2.7 & 5 & 775 & 537--893 & 100 \\
kl10ho & +83.1 & 2.3 & 0.3 & 3.0 & 5 & 1{,}037 & 606--1{,}125 & 100 \\
kl10hoB & +83.1 & 1.9 & 0.2 & 2.5 & 5 & 575 & 450--875 & 100 \\
kl1ho & +82.7 & 2.7 & 0.4 & 3.4 & 5 & 1{,}475 & 1{,}068--1{,}750 & 100 \\
kl1hoB & +83.5 & 2.7 & 0.4 & 3.4 & 5 & 1{,}550 & 1{,}168--1{,}837 & 100 \\
klall & +82.3 & 2.5 & 0.4 & 3.2 & 5 & 1{,}175 & 875--1{,}462 & 100 \\
klall10 & +82.7 & 2.0 & 0.2 & 2.6 & 5 & 850 & 587--1{,}056 & 100 \\
l2d0.1 & +83.1 & 18.0 & 14.8 & 19.6 & no & & & \\
l2d1 & +82.3 & 20.4 & 17.2 & 22.1 & no & & & \\
mcffkl1 & +83.5 & 2.5 & 0.4 & 3.2 & 5 & 1{,}112 & 881--1{,}481 & 100 \\
mcgenkl1 & +84.6 & 2.7 & 0.5 & 3.5 & 5 & 1{,}500 & 1{,}043--1{,}918 & 100 \\
perm & +83.9 & 11.4 & 8.2 & 12.7 & no & & & \\
permkl1 & +83.5 & 2.8 & 0.6 & 3.5 & 5 & 1{,}562 & 1{,}075--1{,}931 & 100 \\
plainsv & +83.1 & 30.6 & 28.1 & 32.5 & no & & & \\
scale0.25 & +45.3 & 7.5 & 4.2 & 8.6 & no & & & \\
scale0.5 & +78.0 & 14.9 & 11.7 & 16.4 & no & & & \\
scale0.75 & +83.1 & 24.1 & 21.3 & 25.9 & no & & & \\
\bottomrule
\end{tabular}}

\end{table}

\clearpage
\vskip 0.2in
\setlength{\bibsep}{2pt plus 1pt}
\bibliography{references}

\begin{thebibliography}{81}
\providecommand{\natexlab}[1]{#1}
\providecommand{\url}[1]{\texttt{#1}}
\expandafter\ifx\csname urlstyle\endcsname\relax
  \providecommand{\doi}[1]{doi: #1}\else
  \providecommand{\doi}{doi: \begingroup \urlstyle{rm}\Url}\fi

\bibitem[Angelopoulos et~al.(2023)Angelopoulos, Bates, Fannjiang, Jordan, and
  Zrnic]{angelopoulos2023ppi}
A.~N. Angelopoulos, S.~Bates, C.~Fannjiang, M.~I. Jordan, and T.~Zrnic.
\newblock {Prediction-powered inference}.
\newblock \emph{Science}, 382\penalty0 (6671):\penalty0 669--674, 2023.
\newblock \doi{10.1126/science.adi6000}.

\bibitem[Angelopoulos et~al.(2025)Angelopoulos, Bates, Cand{\`e}s, Jordan, and
  Lei]{angelopoulos2021learn}
A.~N. Angelopoulos, S.~Bates, E.~J. Cand{\`e}s, M.~I. Jordan, and L.~Lei.
\newblock {Learn then Test: Calibrating Predictive Algorithms to Achieve Risk
  Control}.
\newblock \emph{The Annals of Applied Statistics}, 19\penalty0 (2):\penalty0
  1641--1662, 2025.
\newblock \doi{10.1214/24-AOAS1998}.
\newblock Preprint arXiv:2110.01052 (2021).

\bibitem[Balachandran(2026)]{balachandran2026discern}
V.~Balachandran.
\newblock {Pay Only for Disagreement: Certified No-Regression Verdicts for
  Model Updates with Matching Label-Complexity Bounds}.
\newblock \emph{arXiv preprint arXiv:2609.17560}, 2026.

\bibitem[Bansal et~al.(2019)Bansal, Nushi, Kamar, Weld, Lasecki, and
  Horvitz]{bansal2019updates}
G.~Bansal, B.~Nushi, E.~Kamar, D.~S. Weld, W.~S. Lasecki, and E.~Horvitz.
\newblock {Updates in Human-AI Teams: Understanding and Addressing the
  Performance/Compatibility Tradeoff}.
\newblock In \emph{Proceedings of the AAAI Conference on Artificial
  Intelligence}, volume~33, pages 2429--2437, 2019.
\newblock \doi{10.1609/aaai.v33i01.33012429}.

\bibitem[Betley et~al.(2025)Betley, Tan, Warncke, Sztyber-Betley, Bao, Soto,
  Labenz, and Evans]{betley2025emergent}
J.~Betley, D.~Tan, N.~Warncke, A.~Sztyber-Betley, X.~Bao, M.~Soto, N.~Labenz,
  and O.~Evans.
\newblock {Emergent Misalignment: Narrow finetuning can produce broadly
  misaligned LLMs}.
\newblock In \emph{Proceedings of the 42nd International Conference on Machine
  Learning}, volume 267 of \emph{Proceedings of Machine Learning Research},
  pages 4043--4068. PMLR, 2025.

\bibitem[Biderman et~al.(2024)Biderman, Portes, Gonzalez~Ortiz, Paul,
  Greengard, Jennings, King, Havens, Chiley, Frankle, Blakeney, and
  Cunningham]{biderman2024lora}
D.~Biderman, J.~Portes, J.~J. Gonzalez~Ortiz, M.~Paul, P.~Greengard,
  C.~Jennings, D.~King, S.~Havens, V.~Chiley, J.~Frankle, C.~Blakeney, and
  J.~P. Cunningham.
\newblock {LoRA Learns Less and Forgets Less}.
\newblock \emph{Transactions on Machine Learning Research}, 2024.

\bibitem[Cai et~al.(2022)Cai, Mansimov, Lai, Su, Shu, and
  Zhang]{cai2022measuring}
D.~Cai, E.~Mansimov, Y.-A. Lai, Y.~Su, L.~Shu, and Y.~Zhang.
\newblock Measuring and reducing model update regression in structured
  prediction for {NLP}.
\newblock In \emph{Advances in Neural Information Processing Systems}, 2022.

\bibitem[Chalkidis et~al.(2022)Chalkidis, Jana, Hartung, Bommarito,
  Androutsopoulos, Katz, and Aletras]{chalkidis2022lexglue}
I.~Chalkidis, A.~Jana, D.~Hartung, M.~Bommarito, I.~Androutsopoulos, D.~M.
  Katz, and N.~Aletras.
\newblock {LexGLUE: A Benchmark Dataset for Legal Language Understanding in
  English}.
\newblock In \emph{Proceedings of the 60th Annual Meeting of the Association
  for Computational Linguistics (Volume 1: Long Papers)}, pages 4310--4330.
  Association for Computational Linguistics, 2022.

\bibitem[Chen et~al.(2024)Chen, Zaharia, and Zou]{chen2023chatgpt}
L.~Chen, M.~Zaharia, and J.~Zou.
\newblock {How Is ChatGPT's Behavior Changing over Time?}
\newblock \emph{Harvard Data Science Review}, 6\penalty0 (2), 2024.

\bibitem[Chugg et~al.(2023)Chugg, Cortes-Gomez, Wilder, and
  Ramdas]{chugg2023auditing}
B.~Chugg, S.~Cortes-Gomez, B.~Wilder, and A.~Ramdas.
\newblock Auditing fairness by betting.
\newblock In \emph{Advances in Neural Information Processing Systems}, 2023.

\bibitem[Clark et~al.(2019)Clark, Lee, Chang, Kwiatkowski, Collins, and
  Toutanova]{clark2019boolq}
C.~Clark, K.~Lee, M.-W. Chang, T.~Kwiatkowski, M.~Collins, and K.~Toutanova.
\newblock {BoolQ: Exploring the Surprising Difficulty of Natural Yes/No
  Questions}.
\newblock In \emph{Proceedings of the 2019 Conference of the North American
  Chapter of the Association for Computational Linguistics: Human Language
  Technologies, Volume 1 (Long and Short Papers)}, pages 2924--2936.
  Association for Computational Linguistics, 2019.
\newblock \doi{10.18653/v1/N19-1300}.

\bibitem[Clark et~al.(2018)Clark, Cowhey, Etzioni, Khot, Sabharwal, Schoenick,
  and Tafjord]{clark2018arc}
P.~Clark, I.~Cowhey, O.~Etzioni, T.~Khot, A.~Sabharwal, C.~Schoenick, and
  O.~Tafjord.
\newblock {Think you have Solved Question Answering? Try ARC, the AI2 Reasoning
  Challenge}.
\newblock \emph{arXiv preprint arXiv:1803.05457}, 2018.

\bibitem[Cobbe et~al.(2021)Cobbe, Kosaraju, Bavarian, Chen, Jun, Kaiser,
  Plappert, Tworek, Hilton, Nakano, Hesse, and Schulman]{cobbe2021gsm8k}
K.~Cobbe, V.~Kosaraju, M.~Bavarian, M.~Chen, H.~Jun, L.~Kaiser, M.~Plappert,
  J.~Tworek, J.~Hilton, R.~Nakano, C.~Hesse, and J.~Schulman.
\newblock Training verifiers to solve math word problems.
\newblock \emph{arXiv preprint arXiv:2110.14168}, 2021.

\bibitem[Cohen et~al.(2024)Cohen, Biran, Yoran, Globerson, and
  Geva]{cohen2024evaluating}
R.~Cohen, E.~Biran, O.~Yoran, A.~Globerson, and M.~Geva.
\newblock {Evaluating the Ripple Effects of Knowledge Editing in Language
  Models}.
\newblock \emph{Transactions of the Association for Computational Linguistics},
  12:\penalty0 283--298, 2024.
\newblock \doi{10.1162/tacl_a_00644}.

\bibitem[Dettmers et~al.(2022)Dettmers, Lewis, Shleifer, and
  Zettlemoyer]{dettmers2022optimizers}
T.~Dettmers, M.~Lewis, S.~Shleifer, and L.~Zettlemoyer.
\newblock 8-bit optimizers via block-wise quantization.
\newblock In \emph{International Conference on Learning Representations}, 2022.

\bibitem[Dunlap et~al.(2025)Dunlap, Mandal, Darrell, Steinhardt, and
  Gonzalez]{dunlap2025vibecheck}
L.~Dunlap, K.~Mandal, T.~Darrell, J.~Steinhardt, and J.~E. Gonzalez.
\newblock {VibeCheck: Discover and Quantify Qualitative Differences in Large
  Language Models}.
\newblock In \emph{International Conference on Learning Representations
  (ICLR)}, 2025.

\bibitem[Echterhoff et~al.(2024)Echterhoff, Faghri, Vemulapalli, Hu, Li, Tuzel,
  and Pouransari]{echterhoff2024muscle}
J.~M. Echterhoff, F.~Faghri, R.~Vemulapalli, T.-Y. Hu, C.-L. Li, O.~Tuzel, and
  H.~Pouransari.
\newblock {MUSCLE: A Model Update Strategy for Compatible LLM Evolution}.
\newblock In \emph{Findings of the Association for Computational Linguistics:
  EMNLP 2024}, pages 7320--7332, Miami, Florida, USA, 2024. Association for
  Computational Linguistics.
\newblock URL \url{https://aclanthology.org/2024.findings-emnlp.430/}.

\bibitem[{European Parliament and Council of the European
  Union}(2024)]{euaiact2024}
{European Parliament and Council of the European Union}.
\newblock {Regulation (EU) 2024/1689 of the European Parliament and of the
  Council of 13 June 2024 laying down harmonised rules on artificial
  intelligence (Artificial Intelligence Act)}.
\newblock Official Journal of the European Union, L series, 12 July 2024, 2024.
\newblock URL \url{https://eur-lex.europa.eu/eli/reg/2024/1689/oj}.

\bibitem[Fang et~al.(2025)Fang, Jiang, Wang, Ma, Wang, He, and
  Chua]{fang2025alphaedit}
J.~Fang, H.~Jiang, K.~Wang, Y.~Ma, X.~Wang, X.~He, and T.-S. Chua.
\newblock {AlphaEdit: Null-Space Constrained Knowledge Editing for Language
  Models}.
\newblock In \emph{International Conference on Learning Representations
  (ICLR)}, 2025.

\bibitem[Gao et~al.(2025)Gao, Liang, and Guestrin]{gao2025met}
I.~Gao, P.~Liang, and C.~Guestrin.
\newblock Model equality testing: Which model is this {API} serving?
\newblock In \emph{International Conference on Learning Representations}, 2025.

\bibitem[Grattafiori et~al.(2024)Grattafiori, Dubey, Jauhri, Pandey, Kadian,
  et~al.]{grattafiori2024llama3}
A.~Grattafiori, A.~Dubey, A.~Jauhri, A.~Pandey, A.~Kadian, et~al.
\newblock {The Llama 3 Herd of Models}.
\newblock \emph{arXiv preprint arXiv:2407.21783}, 2024.

\bibitem[Gu et~al.(2024)Gu, Xu, Ma, Lu, Ling, Chang, and Peng]{gu2024editing}
J.-C. Gu, H.-X. Xu, J.-Y. Ma, P.~Lu, Z.-H. Ling, K.-W. Chang, and N.~Peng.
\newblock {Model Editing Harms General Abilities of Large Language Models:
  Regularization to the Rescue}.
\newblock In \emph{Proceedings of the Conference on Empirical Methods in
  Natural Language Processing}, pages 16801--16819, 2024.

\bibitem[Gupta et~al.(2024)Gupta, Rao, and Anumanchipalli]{gupta2024model}
A.~Gupta, A.~Rao, and G.~Anumanchipalli.
\newblock {Model Editing at Scale leads to Gradual and Catastrophic
  Forgetting}.
\newblock In \emph{Findings of the Association for Computational Linguistics:
  ACL 2024}, pages 15202--15232. Association for Computational Linguistics,
  2024.

\bibitem[He and {Thinking Machines Lab}(2025)]{he2025nondeterminism}
H.~He and {Thinking Machines Lab}.
\newblock {Defeating Nondeterminism in LLM Inference}.
\newblock Thinking Machines Lab: Connectionism, Sept. 2025.
\newblock URL
  \url{https://thinkingmachines.ai/blog/defeating-nondeterminism-in-llm-inference/}.

\bibitem[Hendrycks et~al.(2021)Hendrycks, Burns, Basart, Zou, Mazeika, Song,
  and Steinhardt]{hendrycks2021mmlu}
D.~Hendrycks, C.~Burns, S.~Basart, A.~Zou, M.~Mazeika, D.~Song, and
  J.~Steinhardt.
\newblock {Measuring Massive Multitask Language Understanding}.
\newblock In \emph{International Conference on Learning Representations
  (ICLR)}, 2021.

\bibitem[Howard et~al.(2021)Howard, Ramdas, McAuliffe, and
  Sekhon]{howard2021time}
S.~R. Howard, A.~Ramdas, J.~McAuliffe, and J.~Sekhon.
\newblock {Time-uniform, nonparametric, nonasymptotic confidence sequences}.
\newblock \emph{The Annals of Statistics}, 49\penalty0 (2):\penalty0
  1055--1080, 2021.
\newblock \doi{10.1214/20-AOS1991}.

\bibitem[Hu et~al.(2022)Hu, Shen, Wallis, Allen-Zhu, Li, Wang, Wang, and
  Chen]{hu2022lora}
E.~J. Hu, Y.~Shen, P.~Wallis, Z.~Allen-Zhu, Y.~Li, S.~Wang, L.~Wang, and
  W.~Chen.
\newblock {LoRA: Low-Rank Adaptation of Large Language Models}.
\newblock In \emph{International Conference on Learning Representations
  (ICLR)}, 2022.

\bibitem[Jackson et~al.(2025)Jackson, Kravtsov, and Jain]{cursor2025tab}
J.~Jackson, P.~Kravtsov, and S.~Jain.
\newblock {Improving Cursor Tab with online RL}.
\newblock Cursor Blog, Sept. 2025.
\newblock URL \url{https://cursor.com/blog/tab-rl}.

\bibitem[Jiang et~al.(2023)Jiang, Sablayrolles, Mensch, Bamford, Chaplot,
  et~al.]{jiang2023mistral}
A.~Q. Jiang, A.~Sablayrolles, A.~Mensch, C.~Bamford, D.~S. Chaplot, et~al.
\newblock {Mistral 7B}.
\newblock \emph{arXiv preprint arXiv:2310.06825}, 2023.

\bibitem[Jiang et~al.(2022)Jiang, Narasimhan, Bahri, Cotter, and
  Rostamizadeh]{jiang2022churn}
H.~Jiang, H.~Narasimhan, D.~Bahri, A.~Cotter, and A.~Rostamizadeh.
\newblock {Churn Reduction via Distillation}.
\newblock In \emph{International Conference on Learning Representations}, 2022.

\bibitem[Jin et~al.(2021)Jin, Pan, Oufattole, Weng, Fang, and
  Szolovits]{jin2021medqa}
D.~Jin, E.~Pan, N.~Oufattole, W.-H. Weng, H.~Fang, and P.~Szolovits.
\newblock What disease does this patient have? a large-scale open domain
  question answering dataset from medical exams.
\newblock \emph{Applied Sciences}, 11\penalty0 (14):\penalty0 6421, 2021.

\bibitem[Johari et~al.(2017)Johari, Koomen, Pekelis, and
  Walsh]{johari2017peeking}
R.~Johari, P.~Koomen, L.~Pekelis, and D.~Walsh.
\newblock {Peeking at A/B Tests: Why it matters, and what to do about it}.
\newblock In \emph{Proceedings of the 23rd ACM SIGKDD International Conference
  on Knowledge Discovery and Data Mining}, 2017.

\bibitem[Kassem et~al.(2025)Kassem, Shi, Rostamzadeh, and
  Farnadi]{kassem2025mneme}
A.~M. Kassem, Z.~Shi, N.~Rostamzadeh, and G.~Farnadi.
\newblock {Reviving Your MNEME: Predicting The Side Effects of LLM Unlearning
  and Fine-Tuning via Sparse Model Diffing}.
\newblock In \emph{Proceedings of the Conference on Empirical Methods in
  Natural Language Processing}, 2025.

\bibitem[Kaufmann et~al.(2016)Kaufmann, Capp{\'e}, and
  Garivier]{kaufmann2016complexity}
E.~Kaufmann, O.~Capp{\'e}, and A.~Garivier.
\newblock {On the Complexity of Best-Arm Identification in Multi-Armed Bandit
  Models}.
\newblock \emph{Journal of Machine Learning Research}, 17\penalty0
  (1):\penalty0 1--42, 2016.

\bibitem[Kirkpatrick et~al.(2017)Kirkpatrick, Pascanu, Rabinowitz, Veness,
  Desjardins, Rusu, Milan, Quan, Ramalho, Grabska-Barwinska, Hassabis, Clopath,
  Kumaran, and Hadsell]{kirkpatrick2017overcoming}
J.~Kirkpatrick, R.~Pascanu, N.~Rabinowitz, J.~Veness, G.~Desjardins, A.~A.
  Rusu, K.~Milan, J.~Quan, T.~Ramalho, A.~Grabska-Barwinska, D.~Hassabis,
  C.~Clopath, D.~Kumaran, and R.~Hadsell.
\newblock {Overcoming catastrophic forgetting in neural networks}.
\newblock \emph{Proceedings of the National Academy of Sciences}, 114\penalty0
  (13):\penalty0 3521--3526, 2017.
\newblock \doi{10.1073/pnas.1611835114}.

\bibitem[K{\"o}pf et~al.(2023)K{\"o}pf, Kilcher, von R{\"u}tte, Anagnostidis,
  Tam, Stevens, Barhoum, Duc, Stanley, Nagyfi, ES, Suri, Glushkov, Dantuluri,
  Maguire, Schuhmann, Nguyen, and Mattick]{kopf2023openassistant}
A.~K{\"o}pf, Y.~Kilcher, D.~von R{\"u}tte, S.~Anagnostidis, Z.-R. Tam,
  K.~Stevens, A.~Barhoum, N.~M. Duc, O.~Stanley, R.~Nagyfi, S.~ES, S.~Suri,
  D.~Glushkov, A.~Dantuluri, A.~Maguire, C.~Schuhmann, H.~Nguyen, and
  A.~Mattick.
\newblock {OpenAssistant} conversations: Democratizing large language model
  alignment.
\newblock In \emph{Advances in Neural Information Processing Systems Datasets
  and Benchmarks Track}, 2023.

\bibitem[Laroche et~al.(2019)Laroche, Trichelair, and Tachet~des
  Combes]{laroche2019safe}
R.~Laroche, P.~Trichelair, and R.~Tachet~des Combes.
\newblock {Safe Policy Improvement with Baseline Bootstrapping}.
\newblock In \emph{Proceedings of the 36th International Conference on Machine
  Learning}, volume~97 of \emph{Proceedings of Machine Learning Research},
  pages 3652--3661. PMLR, 2019.

\bibitem[Lee et~al.(2019)Lee, Chang, and Toutanova]{lee2019latent}
K.~Lee, M.-W. Chang, and K.~Toutanova.
\newblock Latent retrieval for weakly supervised open domain question
  answering.
\newblock In \emph{Proceedings of the 57th Annual Meeting of the Association
  for Computational Linguistics}, pages 6086--6096, 2019.

\bibitem[Li and Hoiem(2018)]{li2017lwf}
Z.~Li and D.~Hoiem.
\newblock {Learning without Forgetting}.
\newblock \emph{IEEE Transactions on Pattern Analysis and Machine
  Intelligence}, 40\penalty0 (12):\penalty0 2935--2947, 2018.
\newblock \doi{10.1109/TPAMI.2017.2773081}.

\bibitem[Luo et~al.(2025)Luo, Yang, Meng, Li, Zhou, and
  Zhang]{luo2023empirical}
Y.~Luo, Z.~Yang, F.~Meng, Y.~Li, J.~Zhou, and Y.~Zhang.
\newblock {An Empirical Study of Catastrophic Forgetting in Large Language
  Models During Continual Fine-Tuning}.
\newblock \emph{IEEE Transactions on Audio, Speech and Language Processing},
  33:\penalty0 3776--3786, 2025.
\newblock \doi{10.1109/TASLPRO.2025.3606231}.
\newblock Preprint arXiv:2308.08747 (2023).

\bibitem[Maini et~al.(2024)Maini, Feng, Schwarzschild, Lipton, and
  Kolter]{maini2024tofu}
P.~Maini, Z.~Feng, A.~Schwarzschild, Z.~C. Lipton, and J.~Z. Kolter.
\newblock {TOFU}: A task of fictitious unlearning for {LLMs}.
\newblock In \emph{Conference on Language Modeling}, 2024.

\bibitem[Meng et~al.(2022)Meng, Bau, Andonian, and Belinkov]{meng2022locating}
K.~Meng, D.~Bau, A.~Andonian, and Y.~Belinkov.
\newblock {Locating and Editing Factual Associations in GPT}.
\newblock In \emph{Advances in Neural Information Processing Systems},
  volume~35, pages 17359--17372, 2022.

\bibitem[Meng et~al.(2023)Meng, Sen~Sharma, Andonian, Belinkov, and
  Bau]{meng2023memit}
K.~Meng, A.~Sen~Sharma, A.~Andonian, Y.~Belinkov, and D.~Bau.
\newblock {Mass-Editing Memory in a Transformer}.
\newblock In \emph{International Conference on Learning Representations
  (ICLR)}, 2023.

\bibitem[Merity et~al.(2017)Merity, Xiong, Bradbury, and
  Socher]{merity2017wikitext}
S.~Merity, C.~Xiong, J.~Bradbury, and R.~Socher.
\newblock Pointer sentinel mixture models.
\newblock In \emph{International Conference on Learning Representations}, 2017.

\bibitem[Mihaylov et~al.(2018)Mihaylov, Clark, Khot, and
  Sabharwal]{mihaylov2018openbookqa}
T.~Mihaylov, P.~Clark, T.~Khot, and A.~Sabharwal.
\newblock Can a suit of armor conduct electricity? a new dataset for open book
  question answering.
\newblock In \emph{Proceedings of the 2018 Conference on Empirical Methods in
  Natural Language Processing}, pages 2381--2391, 2018.

\bibitem[Milani~Fard et~al.(2016)Milani~Fard, Cormier, Canini, and
  Gupta]{milanifard2016launch}
M.~Milani~Fard, Q.~Cormier, K.~Canini, and M.~Gupta.
\newblock {Launch and Iterate: Reducing Prediction Churn}.
\newblock In \emph{Advances in Neural Information Processing Systems}, pages
  3171--3179, 2016.

\bibitem[Mitchell et~al.(2022)Mitchell, Lin, Bosselut, Finn, and
  Manning]{mitchell2022mend}
E.~Mitchell, C.~Lin, A.~Bosselut, C.~Finn, and C.~D. Manning.
\newblock Fast model editing at scale.
\newblock In \emph{International Conference on Learning Representations}, 2022.

\bibitem[Pal et~al.(2022)Pal, Umapathi, and Sankarasubbu]{pal2022medmcqa}
A.~Pal, L.~K. Umapathi, and M.~Sankarasubbu.
\newblock {MedMCQA: A Large-scale Multi-Subject Multi-Choice Dataset for
  Medical domain Question Answering}.
\newblock In \emph{Proceedings of the Conference on Health, Inference, and
  Learning}, volume 174 of \emph{Proceedings of Machine Learning Research},
  pages 248--260. PMLR, 2022.

\bibitem[Podkopaev and Ramdas(2022)]{podkopaev2022tracking}
A.~Podkopaev and A.~Ramdas.
\newblock {Tracking the Risk of a Deployed Model and Detecting Harmful
  Distribution Shifts}.
\newblock In \emph{International Conference on Learning Representations}, 2022.

\bibitem[Qi et~al.(2024)Qi, Zeng, Xie, Chen, Jia, Mittal, and
  Henderson]{qi2024finetuning}
X.~Qi, Y.~Zeng, T.~Xie, P.-Y. Chen, R.~Jia, P.~Mittal, and P.~Henderson.
\newblock {Fine-tuning Aligned Language Models Compromises Safety, Even When
  Users Do Not Intend To!}
\newblock In \emph{International Conference on Learning Representations
  (ICLR)}, 2024.

\bibitem[{Qwen Team}(2024)]{qwen2025qwen25}
{Qwen Team}.
\newblock {Qwen2.5 Technical Report}.
\newblock \emph{arXiv preprint arXiv:2412.15115}, 2024.

\bibitem[Rafailov et~al.(2023)Rafailov, Sharma, Mitchell, Ermon, Manning, and
  Finn]{rafailov2023dpo}
R.~Rafailov, A.~Sharma, E.~Mitchell, S.~Ermon, C.~D. Manning, and C.~Finn.
\newblock {Direct Preference Optimization: Your Language Model is Secretly a
  Reward Model}.
\newblock In \emph{Advances in Neural Information Processing Systems}, 2023.

\bibitem[Robbins(1970)]{robbins1970statistical}
H.~Robbins.
\newblock {Statistical Methods Related to the Law of the Iterated Logarithm}.
\newblock \emph{The Annals of Mathematical Statistics}, 41\penalty0
  (5):\penalty0 1397--1409, 1970.
\newblock \doi{10.1214/aoms/1177696786}.

\bibitem[R{\"o}ttger et~al.(2024)R{\"o}ttger, Kirk, Vidgen, Attanasio, Bianchi,
  and Hovy]{rottger2024xstest}
P.~R{\"o}ttger, H.~R. Kirk, B.~Vidgen, G.~Attanasio, F.~Bianchi, and D.~Hovy.
\newblock {XSTest: A Test Suite for Identifying Exaggerated Safety Behaviours
  in Large Language Models}.
\newblock In \emph{Proceedings of the Conference of the North American Chapter
  of the Association for Computational Linguistics}, 2024.

\bibitem[Schumann et~al.(2023)Schumann, Mansimov, Lai, Pappas, Gao, and
  Zhang]{schumann2023backward}
R.~Schumann, E.~Mansimov, Y.-A. Lai, N.~Pappas, X.~Gao, and Y.~Zhang.
\newblock Backward compatibility during data updates by weight interpolation.
\newblock \emph{arXiv preprint arXiv:2301.10546}, 2023.

\bibitem[Shao et~al.(2024)Shao, Wang, Zhu, Xu, Song, Bi, Zhang, Zhang, Li, Wu,
  and Guo]{shao2024deepseekmath}
Z.~Shao, P.~Wang, Q.~Zhu, R.~Xu, J.~Song, X.~Bi, H.~Zhang, M.~Zhang, Y.~K. Li,
  Y.~Wu, and D.~Guo.
\newblock {DeepSeekMath}: Pushing the limits of mathematical reasoning in open
  language models.
\newblock \emph{arXiv preprint arXiv:2402.03300}, 2024.

\bibitem[Shenfeld et~al.(2026{\natexlab{a}})Shenfeld, Damani, H{\"u}botter, and
  Agrawal]{shenfeld2026sdft}
I.~Shenfeld, M.~Damani, J.~H{\"u}botter, and P.~Agrawal.
\newblock {Self-Distillation Enables Continual Learning}.
\newblock In \emph{International Conference on Machine Learning (ICML)},
  2026{\natexlab{a}}.
\newblock arXiv:2601.19897.

\bibitem[Shenfeld et~al.(2026{\natexlab{b}})Shenfeld, Pari, and
  Agrawal]{shenfeld2025rl}
I.~Shenfeld, J.~Pari, and P.~Agrawal.
\newblock {RL's Razor: Why Online Reinforcement Learning Forgets Less}.
\newblock In \emph{International Conference on Learning Representations
  (ICLR)}, 2026{\natexlab{b}}.
\newblock arXiv:2509.04259 (2025).

\bibitem[Socher et~al.(2013)Socher, Perelygin, Wu, Chuang, Manning, Ng, and
  Potts]{socher2013sst}
R.~Socher, A.~Perelygin, J.~Wu, J.~Chuang, C.~D. Manning, A.~Ng, and C.~Potts.
\newblock {Recursive Deep Models for Semantic Compositionality Over a Sentiment
  Treebank}.
\newblock In \emph{Proceedings of the 2013 Conference on Empirical Methods in
  Natural Language Processing}, pages 1631--1642. Association for Computational
  Linguistics, 2013.

\bibitem[Srivastava et~al.(2020)Srivastava, Nushi, Kamar, Shah, and
  Horvitz]{srivastava2020empirical}
M.~Srivastava, B.~Nushi, E.~Kamar, S.~Shah, and E.~Horvitz.
\newblock {An Empirical Analysis of Backward Compatibility in Machine Learning
  Systems}.
\newblock In \emph{Proceedings of the 26th ACM SIGKDD International Conference
  on Knowledge Discovery \& Data Mining}, pages 3272--3280, 2020.
\newblock \doi{10.1145/3394486.3403379}.

\bibitem[Talmor et~al.(2019)Talmor, Herzig, Lourie, and
  Berant]{talmor2019commonsenseqa}
A.~Talmor, J.~Herzig, N.~Lourie, and J.~Berant.
\newblock {CommonsenseQA: A Question Answering Challenge Targeting Commonsense
  Knowledge}.
\newblock In \emph{Proceedings of the 2019 Conference of the North American
  Chapter of the Association for Computational Linguistics: Human Language
  Technologies, Volume 1 (Long and Short Papers)}, pages 4149--4158.
  Association for Computational Linguistics, 2019.
\newblock \doi{10.18653/v1/N19-1421}.

\bibitem[Taori et~al.(2023)Taori, Gulrajani, Zhang, Dubois, Li, Guestrin,
  Liang, and Hashimoto]{taori2023alpaca}
R.~Taori, I.~Gulrajani, T.~Zhang, Y.~Dubois, X.~Li, C.~Guestrin, P.~Liang, and
  T.~B. Hashimoto.
\newblock Stanford alpaca: An instruction-following {LLaMA} model.
\newblock \url{https://github.com/tatsu-lab/stanford_alpaca}, 2023.

\bibitem[Thomas et~al.(2015)Thomas, Theocharous, and
  Ghavamzadeh]{thomas2015high}
P.~S. Thomas, G.~Theocharous, and M.~Ghavamzadeh.
\newblock {High Confidence Policy Improvement}.
\newblock In \emph{Proceedings of the 32nd International Conference on Machine
  Learning}, volume~37 of \emph{Proceedings of Machine Learning Research},
  pages 2380--2388. PMLR, 2015.

\bibitem[Thomas et~al.(2019)Thomas, Castro~da Silva, Barto, Giguere, Brun, and
  Brunskill]{thomas2019preventing}
P.~S. Thomas, B.~Castro~da Silva, A.~G. Barto, S.~Giguere, Y.~Brun, and
  E.~Brunskill.
\newblock {Preventing undesirable behavior of intelligent machines}.
\newblock \emph{Science}, 366\penalty0 (6468):\penalty0 999--1004, 2019.
\newblock \doi{10.1126/science.aag3311}.

\bibitem[Titsias et~al.(2020)Titsias, Schwarz, de~G.~Matthews, Pascanu, and
  Teh]{titsias2020functional}
M.~K. Titsias, J.~Schwarz, A.~G. de~G.~Matthews, R.~Pascanu, and Y.~W. Teh.
\newblock Functional regularisation for continual learning with {G}aussian
  processes.
\newblock In \emph{International Conference on Learning Representations}, 2020.

\bibitem[{U.S. Food and Drug Administration}(2024)]{fda2024pccp}
{U.S. Food and Drug Administration}.
\newblock {Marketing Submission Recommendations for a Predetermined Change
  Control Plan for Artificial Intelligence-Enabled Device Software Functions:
  Guidance for Industry and Food and Drug Administration Staff}.
\newblock Final guidance, Dec. 2024.
\newblock URL
  \url{https://www.fda.gov/regulatory-information/search-fda-guidance-documents/marketing-submission-recommendations-predetermined-change-control-plan-artificial-intelligence}.

\bibitem[Ville(1939)]{ville1939etude}
J.~Ville.
\newblock \emph{{{\'E}tude critique de la notion de collectif}}.
\newblock Number~3 in Monographies des Probabilit{\'e}s. Gauthier-Villars,
  Paris, 1939.

\bibitem[Wald(1945)]{wald1945sequential}
A.~Wald.
\newblock {Sequential Tests of Statistical Hypotheses}.
\newblock \emph{The Annals of Mathematical Statistics}, 16\penalty0
  (2):\penalty0 117--186, 1945.

\bibitem[Wang et~al.(2024)Wang, Ma, Hu, Weber-Genzel, R{\"o}ttger, Kreuter,
  Hovy, and Plank]{wang2024myanswer}
X.~Wang, B.~Ma, C.~Hu, L.~Weber-Genzel, P.~R{\"o}ttger, F.~Kreuter, D.~Hovy,
  and B.~Plank.
\newblock {``My Answer is C'': First-Token Probabilities Do Not Match Text
  Answers in Instruction-Tuned Language Models}.
\newblock In \emph{Findings of the Association for Computational Linguistics:
  ACL 2024}, pages 7407--7416, 2024.

\bibitem[Waudby-Smith and Ramdas(2020)]{waudbysmith2020wor}
I.~Waudby-Smith and A.~Ramdas.
\newblock Confidence sequences for sampling without replacement.
\newblock In \emph{Advances in Neural Information Processing Systems}, 2020.

\bibitem[Waudby-Smith and Ramdas(2024)]{waudbysmith2024estimating}
I.~Waudby-Smith and A.~Ramdas.
\newblock {Estimating means of bounded random variables by betting}.
\newblock \emph{Journal of the Royal Statistical Society Series B: Statistical
  Methodology}, 86\penalty0 (1):\penalty0 1--27, 2024.
\newblock \doi{10.1093/jrsssb/qkad009}.

\bibitem[Xie et~al.(2021)Xie, Lai, Xiong, Zhang, and Soatto]{xie2021regression}
Y.~Xie, Y.-A. Lai, Y.~Xiong, Y.~Zhang, and S.~Soatto.
\newblock Regression bugs are in your model! measuring, reducing and analyzing
  regressions in {NLP} model updates.
\newblock In \emph{Proceedings of the 59th Annual Meeting of the Association
  for Computational Linguistics}, 2021.

\bibitem[Yan et~al.(2021)Yan, Xiong, Kundu, Yang, Deng, Wang, Xia, and
  Soatto]{yan2021positive}
S.~Yan, Y.~Xiong, K.~Kundu, S.~Yang, S.~Deng, M.~Wang, W.~Xia, and S.~Soatto.
\newblock {Positive-Congruent Training: Towards Regression-Free Model Updates}.
\newblock In \emph{Proceedings of the IEEE/CVF Conference on Computer Vision
  and Pattern Recognition (CVPR)}, pages 14294--14303, 2021.
\newblock \doi{10.1109/CVPR46437.2021.01407}.

\bibitem[Yang et~al.(2024)Yang, Liu, Pang, Wang, Feng, Zhu, and
  Chen]{yang2024sdft}
Z.~Yang, Q.~Liu, T.~Pang, H.~Wang, H.~Feng, M.~Zhu, and W.~Chen.
\newblock {Self-Distillation Bridges Distribution Gap in Language Model
  Fine-Tuning}.
\newblock In \emph{Proceedings of the Annual Meeting of the Association for
  Computational Linguistics}, 2024.

\bibitem[Zellers et~al.(2019)Zellers, Holtzman, Bisk, Farhadi, and
  Choi]{zellers2019hellaswag}
R.~Zellers, A.~Holtzman, Y.~Bisk, A.~Farhadi, and Y.~Choi.
\newblock {HellaSwag}: Can a machine really finish your sentence?
\newblock In \emph{Proceedings of the 57th Annual Meeting of the Association
  for Computational Linguistics}, pages 4791--4800, 2019.

\bibitem[Zhang et~al.(2026)Zhang, Wu, and Qian]{zhang2026slicewise}
S.-W. Zhang, T.~Wu, and F.-F. Qian.
\newblock {Certifying Model Upgrades with Slice-Wise Non-Regression and
  Incumbent Fallback}.
\newblock \emph{arXiv preprint arXiv:2609.13714}, 2026.

\bibitem[Zhang et~al.(2015)Zhang, Zhao, and LeCun]{zhang2015agnews}
X.~Zhang, J.~Zhao, and Y.~LeCun.
\newblock {Character-level Convolutional Networks for Text Classification}.
\newblock In \emph{Advances in Neural Information Processing Systems},
  volume~28, pages 649--657, 2015.

\bibitem[Zheng et~al.(2021)Zheng, Guha, Anderson, Henderson, and
  Ho]{zheng2021casehold}
L.~Zheng, N.~Guha, B.~R. Anderson, P.~Henderson, and D.~E. Ho.
\newblock {When Does Pretraining Help? Assessing Self-Supervised Learning for
  Law and the CaseHOLD Dataset of 53,000+ Legal Holdings}.
\newblock In \emph{Proceedings of the Eighteenth International Conference on
  Artificial Intelligence and Law (ICAIL)}, pages 159--168, 2021.
\newblock \doi{10.1145/3462757.3466088}.

\bibitem[Zhong et~al.(2022)Zhong, Snell, Klein, and
  Steinhardt]{zhong2022describing}
R.~Zhong, C.~Snell, D.~Klein, and J.~Steinhardt.
\newblock {Describing Differences between Text Distributions with Natural
  Language}.
\newblock In \emph{Proceedings of the 39th International Conference on Machine
  Learning}, volume 162 of \emph{Proceedings of Machine Learning Research},
  pages 27099--27116. PMLR, 2022.

\bibitem[Zou et~al.(2023)Zou, Wang, Kolter, and Fredrikson]{zou2023universal}
A.~Zou, Z.~Wang, J.~Z. Kolter, and M.~Fredrikson.
\newblock Universal and transferable adversarial attacks on aligned language
  models.
\newblock \emph{arXiv preprint arXiv:2307.15043}, 2023.

\bibitem[Zweiger et~al.(2025)Zweiger, Pari, Guo, Aky{\"u}rek, Kim, and
  Agrawal]{zweiger2025seal}
A.~Zweiger, J.~Pari, H.~Guo, E.~Aky{\"u}rek, Y.~Kim, and P.~Agrawal.
\newblock {Self-Adapting Language Models}.
\newblock In \emph{Advances in Neural Information Processing Systems},
  volume~38, 2025.

\end{thebibliography}

\end{document}